\pdfoutput=1
\documentclass[10pt]{article}
\usepackage[preprint]{style/tmlr}
\usepackage{amsmath,amsfonts,bm}

\def\eqref#1{equation~\ref{#1}}

\def\1{\bm{1}}

\DeclareMathAlphabet{\mathsfit}{\encodingdefault}{\sfdefault}{m}{sl}
\SetMathAlphabet{\mathsfit}{bold}{\encodingdefault}{\sfdefault}{bx}{n}

\usepackage{amsthm}
\usepackage{hyperref}
\usepackage{url}
\usepackage{graphicx}
\usepackage{booktabs}
\usepackage{multirow}
\usepackage{makecell}
\usepackage{subcaption}
\usepackage[dvipsnames,table]{xcolor}
\usepackage{colortbl}
\usepackage{pifont}
\usepackage{xspace}
\usepackage[breakable,skins]{tcolorbox}
\usepackage{titlesec}
\usepackage{wrapfig}
\newcommand{\wraptail}{\vspace{-9pt}}
\usepackage{needspace}
\usepackage{dsfont}
\usepackage{amssymb}
\usepackage{titletoc}
\usepackage{algorithm}
 
\usepackage{algorithmic}
\usepackage{enumitem}
\usepackage{tabularx}
\usepackage{alltt}
\usepackage{CJKutf8}

\theoremstyle{remark}

\colorlet{phaseI}{rgb:red!2,65;green!30,60;blue!20,125}
\colorlet{phaseII}{rgb:red!2,65;green!30,90;blue!20,125}
\colorlet{phaseIII}{rgb:red!60,100;green!20,90;blue!30,125}

\newcommand{\Comment}[1]{\hfill$\triangleright$ \textit{#1}}

\newcommand{\up}[1]{{\color{RedOrange}$\uparrow$#1}}
\newcommand{\down}[1]{{\color{BlueGreen}$\downarrow$#1}}

\newcommand{\std}[1]{$_{\pm\text{\scriptsize #1}}$}
\definecolor{deepred}{rgb}{0.6,0,0}
\definecolor{crystalblue}{rgb}{0.20,0.40,0.80}

\definecolor{darkgreen}{RGB}{0, 181, 18}
\definecolor{darkred}{RGB}{252, 90, 90}
\newcommand{\gain}[1]{$\uparrow$ #1}
\newcommand{\mygreen}[1]{\cellcolor{darkgreen!#1}}

\definecolor{DeltaBg}{HTML}{D4F2D7}
\definecolor{SearchBg}{HTML}{C2E6F5}
\definecolor{AgenticBg}{HTML}{F5C2CC}
\definecolor{MathBg}{HTML}{E6D4F2}
\definecolor{ScienceBg}{HTML}{FBE0BC}

\definecolor{insightcolor}{HTML}{B79AD0}
\definecolor{risecolor}{HTML}{159A9C}
\definecolor{oursrowcolor}{HTML}{F4EEF8}
\definecolor{oursrowcolorsecond}{HTML}{F9F2FB}
\definecolor{questionblue}{rgb}{0.21,0.49,0.74}

\newcommand{\tmlrband}[3]{%
  \refstepcounter{#1}%
  \vspace{0.30em}\noindent\begingroup\setlength{\fboxsep}{0pt}%
  \colorbox{insightcolor!7}{%
    \parbox{0.97\linewidth}{%
      \vspace{0.35em}\hspace{0.65em}%
      \textcolor{insightcolor}{\rule{1.4pt}{1.05em}}\hspace{0.55em}%
      \parbox[t]{0.89\linewidth}{\textbf{\textsc{#2}~\csname the#1\endcsname.}~\emph{#3}}%
      \vspace{0.35em}}}%
  \endgroup\vspace{0.30em}}
\newcommand{\obs}[1]{\tmlrband{observation}{Observation}{#1}}

\newcommand{\llmname}[1]{{\fontfamily{pcr}\selectfont #1}}

\providecommand{\std}[1]{$_{\pm\text{\scriptsize #1}}$}

\colorlet{AccentW}{insightcolor}
\definecolor{AccentCgb}{HTML}{9EB5C8}
\definecolor{AccentC}{HTML}{F9BB79}
\newcommand{\corequestion}[1]{%
  \begin{tcolorbox}[
    enhanced,
    colback=white,
    colframe=AccentW!60!black,
    coltitle=black,
    boxrule=0.9pt,
    left=6pt,right=6pt,top=7pt,bottom=6pt,
    title=\textbf{Core question},
    attach boxed title to top left={yshift=-3.2mm, xshift=8mm},
    boxed title style={
      colback=white,
      coltext=black,
      boxrule=0pt,
      frame hidden,
      sharp corners,
      interior style={fill=white},
      left=4pt,right=4pt,top=1pt,bottom=1pt,
    },
  ]
  \centering
  \textit{#1}
  \end{tcolorbox}}
\newcommand{\hlw}[1]{\begingroup\setlength{\fboxsep}{2pt}\colorbox{AccentW!45}{\ensuremath{\displaystyle #1}}\endgroup}
\newcommand{\hlc}[1]{\begingroup\setlength{\fboxsep}{2pt}\colorbox{AccentC!40}{\ensuremath{\displaystyle #1}}\endgroup}

\newcommand{\eqcmt}[1]{\textcolor{gray!70!black}{\text{\scriptsize\ttfamily $\triangleright$\,#1}}}
\newcommand{\wcol}[1]{\textcolor{AccentW!80!black}{#1}}
\newcommand{\ccol}[1]{\textcolor{AccentC!55!black}{#1}}
\newcommand{\ind}{\mathds{1}}

\definecolor{mylinkcolor}{RGB}{11, 20, 110}

\definecolor{DeltaBg}{HTML}{D4F2D7}
\definecolor{SearchBg}{HTML}{C2E6F5}
\definecolor{AgenticBg}{HTML}{F5C2CC}
\definecolor{MathBg}{HTML}{E6D4F2}
\definecolor{ScienceBg}{HTML}{FBE0BC}

\definecolor{darkgreen}{RGB}{0, 181, 18}
\definecolor{darkred}{RGB}{252, 90, 90}
\providecommand{\mygreen}[1]{\cellcolor{darkgreen!#1}}

\definecolor{toolcardbox}{RGB}{240, 248, 255}
\definecolor{toolcardborder}{RGB}{52, 52, 173}
\definecolor{mygray}{RGB}{242, 242, 242}
\definecolor{codegray}{gray}{0.95}
\definecolor{failbg}{RGB}{248, 230, 234}
\definecolor{failframe}{RGB}{176, 36, 24}
\definecolor{failbadge}{RGB}{225, 151, 168}
\definecolor{successbg}{RGB}{239, 255, 229}
\definecolor{successframe}{RGB}{34, 139, 34}
\definecolor{successbadge}{RGB}{182, 200, 108}

\newtcolorbox{summarybox}{%
    colback=toolcardbox,
    colframe=toolcardborder,
    arc=3pt,
    boxrule=0.8pt,
    left=2pt, right=2pt, top=2pt, bottom=2pt,
    fonttitle=\bfseries,
    breakable,
}

\newtcolorbox{limitationbox}{%
    colback=failbg,
    colframe=failframe,
    arc=1pt,
    boxrule=0.8pt,
    left=4pt, right=4pt, top=4pt, bottom=4pt,
    fonttitle=\bfseries\small,
    title={Limitation},
    coltitle=white,
    fontupper=\small,
    breakable,
    before skip=6pt, after skip=6pt
}

\newtcolorbox{bestpracticebox}{%
    colback=successbg,
    colframe=successframe,
    arc=1pt,
    boxrule=0.8pt,
    left=4pt, right=4pt, top=4pt, bottom=4pt,
    fonttitle=\bfseries\small,
    title={Best Practice},
    coltitle=white,
    fontupper=\small,
    breakable,
    before skip=6pt, after skip=6pt
}

\newtcolorbox{custombox}[2][]{
    colback=toolcardbox,
    colframe=toolcardborder,
    coltitle=white,
    arc=1pt,
    boxrule=1pt,
    fonttitle=\bfseries,
    title={#2},
    left=5pt, right=5pt, top=5pt, bottom=5pt,
    before skip=1em, after skip=1em,
    fontupper=\small,
    breakable,
    width=1.\linewidth,
    #1
}

\newtcolorbox{casebox}[3]{
    colback=#1,
    colframe=#2,
    arc=1pt,
    coltitle=white,
    fonttitle=\bfseries,
    title={#3},
    boxrule=1pt,
    rounded corners,
    breakable
}
\newtcolorbox{casecode}[1]{%
    colback=#1,
    colframe=toolcardbox,
    arc=1pt,
    boxrule=0pt,
    top=-3pt, bottom=0pt,
    left=3pt, right=3pt,
    boxsep=0pt,
    fontupper=\ttfamily\fontsize{8.0}{10}\selectfont,
    breakable,
}
\newtcolorbox{simplecode}[1][]{%
    colback=toolcardbox,
    colframe=toolcardbox,
    arc=1pt,
    boxrule=0pt,
    top=-3pt, bottom=0pt,
    left=3pt, right=3pt,
    boxsep=0pt,
    fontupper=\ttfamily\fontsize{8.0}{10}\selectfont,
    breakable,
    #1
}

\definecolor{zebra}{gray}{0.93}
\definecolor{herobg}{RGB}{230,239,246}
\definecolor{grpblue}{RGB}{228,238,246}
\definecolor{grpteal}{RGB}{228,242,240}
\definecolor{grppurple}{RGB}{236,232,243}%
\definecolor{grporange}{RGB}{252,240,229}%

\newtcolorbox{envcardbox}[1]{breakable, enhanced, colback=toolcardbox,
  colframe=toolcardborder, coltitle=white, fonttitle=\bfseries\small,
  title={Environment Card: #1}, rounded corners, boxrule=1pt,
  left=5pt, right=5pt, top=4pt, bottom=4pt, fontupper=\small}

\definecolor{bandblue}{RGB}{232,239,249}
\definecolor{bandgreen}{RGB}{233,243,232}
\definecolor{bandorange}{RGB}{251,239,225}
\definecolor{bandpurple}{RGB}{242,235,244}
\definecolor{bandred}{RGB}{248,234,232}
\providecommand{\Comment}[1]{\hfill$\triangleright$ \textit{#1}}

\newcommand{\nApps}{20}
\newcommand{\nTemplates}{416}
\newcommand{\nTrainTemplates}{160}
\newcommand{\nTestTemplates}{256}
\newcommand{\nSuites}{24}
\newcommand{\nUnseenApps}{5}
\newcommand{\nExpTemplates}{122}
\newcommand{\nSeenTemplates}{165}
\newcommand{\nUnseenTemplates}{59}
\newcommand{\nMixedTemplates}{32}
\newcommand{\kInstances}{8}
\newcommand{\nPairsPerSeed}{1\,106}
\newcommand{\nEpPerTemplate}{20}

\newcommand{\nAWTasks}{116}
\newcommand{\nAWApps}{20}
\newcommand{\nAWSeenApps}{12}
\newcommand{\nAWUnseenApps}{8}
\newcommand{\nAWHeldoutTasks}{58}
\newcommand{\kAWInstances}{four}
\newcommand{\nAWPairsPerSeed}{232}

\newcommand{\nSeeds}{three}
\newcommand{\nBoot}{10\,000}
\newcommand{\kNotes}{five}
\newcommand{\jExact}{0.9}
\newcommand{\jRelaxed}{0.5}
\newcommand{\agreeFrac}{0.5}

\newcommand{\devAcc}{0.90}
\newcommand{\nBackbones}{three}
\newcommand{\nRetry}{three}
\newcommand{\kAtomic}{ten}

\newcommand{\nAWNodes}{500}
\newcommand{\nFPred}{five}
\newcommand{\nAWEpPerTask}{ten}

\newcommand{\nOATasks}{15}
\newcommand{\nOAApps}{six}
\newcommand{\nOAEp}{20}
\newcommand{\nOAInstances}{150}
\newcommand{\nAllCPer}{two}
\newcommand{\nAllCTotal}{eight}
\newcommand{\nAllCTotalFlat}{five}
\newcommand{\nDistillEp}{ten}
\newcommand{\jSensHi}{0.8}
\newcommand{\jSensLo}{0.6}
\newcommand{\kSensLo}{three}
\newcommand{\kSensHi}{ten}
\newcommand{\rankSensLo}{64}
\newcommand{\rankSensHi}{128}
\newcommand{\poolSensLo}{five}
\newcommand{\poolSensHi}{ten}
\newcommand{\capSensLo}{100}
\newcommand{\capSensHi}{300}
\newcommand{\sameDiag}{2 percent}
\newcommand{\loraRank}{16}
\newcommand{\loraAlpha}{32}
\newcommand{\loraEpochs}{three}
\newcommand{\loraLR}{$10^{-5}$}

\newcommand{\AllWGain}{7.1}
\newcommand{\AllCGain}{8.0}
\newcommand{\DeltaG}{+4.9}
\newcommand{\DeltaP}{$-$4.2}
\newcommand{\DeltaF}{$-$4.1}
\newcommand{\DeltaL}{+1.5}
\newcommand{\SignAgree}{24 of 24}
\newcommand{\DupShift}{+3.5}
\newcommand{\DestateDrop}{3.4}
\newcommand{\KappaOnline}{0.56}
\newcommand{\RoutedGain}{+3.5}
\newcommand{\AntiGap}{+8.9}
\newcommand{\UseDrop}{16 percent}
\newcommand{\CtxInfo}{0.65}
\newcommand{\PolSlope}{$-$0.52}
\newcommand{\ConflictNeither}{36 percent}

\newcommand{\GainWG}{+6.7}
\newcommand{\GainCG}{+1.8}
\newcommand{\GainWP}{+1.0}
\newcommand{\GainCP}{+5.2}

\newcommand{\CoexG}{+0.1}
\newcommand{\CoexF}{+3.5}
\newcommand{\CrossW}{$-$7.5}
\newcommand{\CrossC}{+3.6}
\newcommand{\SelfW}{+0.9}
\newcommand{\SelfC}{+3.6}
\newcommand{\PracticeGain}{+0.6}
\newcommand{\UpC}{+3.2}
\newcommand{\UpW}{$-$1.4}
\newcommand{\ReadBefore}{42 percent}
\newcommand{\ReadAfterG}{16 percent}
\newcommand{\ReadAfterP}{22 percent}
\newcommand{\DownC}{+4.8}
\newcommand{\DownW}{$-$1.4}
\newcommand{\SmallC}{+2.2}
\newcommand{\SmallW}{$-$2.3}
\newcommand{\CrossWb}{$-$2.0}
\newcommand{\rG}{0.39}
\newcommand{\kG}{0.10}
\newcommand{\rP}{0.05}
\newcommand{\kP}{0.66}
\newcommand{\rF}{0.32}
\newcommand{\kF}{0.56}
\newcommand{\rL}{0.45}
\newcommand{\kL}{0.53}
\newcommand{\ReadRho}{$-$0.92}
\newcommand{\SignClear}{19}
\newcommand{\SignScore}{21.5/24}
\newcommand{\GainWF}{+0.8}

\newcommand{\PracticeDiff}{1.1}
\newcommand{\ReadAfterL}{15 percent}
\newcommand{\ReadBeforeP}{33 percent}
\newcommand{\ReadBeforeL}{48 percent}
\newcommand{\CtxCap}{0.20}
\newcommand{\PolSlopeInfo}{0.49}
\newcommand{\ConflictNeitherLow}{14 percent}

\newcommand{\RuleAlpha}{11.29}
\newcommand{\RuleBeta}{5.73}
\newcommand{\RuleGamma}{1.28}
\newcommand{\RuleAlphaLo}{9.01}
\newcommand{\RuleAlphaHi}{9.32}
\newcommand{\RuleBetaLo}{4.86}
\newcommand{\RuleBetaHi}{5.28}
\newcommand{\RuleGammaLo}{0.70}
\newcommand{\RuleGammaHi}{0.94}
\newcommand{\ScoreG}{2.52}
\newcommand{\ScoreP}{$-$4.45}
\newcommand{\ScoreF}{$-$0.89}
\newcommand{\ScoreL}{0.74}
\newcommand{\DupEight}{$-$0.8}
\newcommand{\DupEightQwen}{+1.0}
\newcommand{\DupShiftF}{+3.6}
\newcommand{\DestateDropP}{3.9}
\newcommand{\ShuffleMax}{1.1}
\newcommand{\RandomNotes}{$-$1.6}
\newcommand{\OAStaleW}{$-$8.4}
\newcommand{\OAFreshC}{+3.3}
\newcommand{\OAConflictLo}{2.7}
\newcommand{\OAConflictHi}{12.7}
\newcommand{\ConflictCtxLow}{52 percent}
\newcommand{\ConflictCtxHigh}{43 percent}
\newcommand{\ConflictWLow}{34 percent}
\newcommand{\ConflictWHigh}{21 percent}
\newcommand{\RoutedGainMG}{+4.4}
\newcommand{\RoutedGainAWQwen}{+5.0}
\newcommand{\RoutedVsTraj}{+9.3}
\newcommand{\RoutedSelfGap}{$-$0.2}
\newcommand{\nFamilyCells}{720}
\newcommand{\nProducerPairs}{15}
\newcommand{\nFamilies}{16}
\newcommand{\CtxCapPartial}{0.05}
\newcommand{\PolPartial}{$-$0.52}

\newcommand{\GapCorr}{$-$0.13}
\newcommand{\CtxBinLo}{+0.2}
\newcommand{\CtxBinHi}{+11.3}
\newcommand{\WBinLo}{+4.0}
\newcommand{\WBinHi}{$-$5.7}
\newcommand{\InfoSmall}{0.04}
\newcommand{\InfoSelf}{0.06}
\newcommand{\InfoLarge}{0.08}
\newcommand{\PolSame}{0.16}
\newcommand{\InfoForeign}{0.07}
\newcommand{\PolForeign}{0.46}
\newcommand{\InfoRetrained}{0.07}
\newcommand{\PolRetrained}{0.17}

\newcommand{\RoutedGainAWScale}{+3.2}

\newcommand{\AntiGapAW}{+6.7}
\newcommand{\AntiGapBoth}{+7.8}
\newcommand{\DupR}{0.42}
\newcommand{\DupBoundaryR}{0.44}
\newcommand{\DestateGainP}{+1.3}
\newcommand{\DestateGainF}{+1.6}
\newcommand{\AtomicGain}{+1.4}
\newcommand{\SFTGain}{+0.7}
\newcommand{\ReadItemGLo}{0.73}
\newcommand{\ReadItemGHi}{0.11}

\newcommand{\ReadItemLLo}{0.63}
\newcommand{\ReadItemLHi}{0.11}
\newcommand{\SplitLeadMGLo}{+3.7}
\newcommand{\SplitLeadMGHi}{+5.1}
\newcommand{\SplitLeadAWLo}{+2.4}
\newcommand{\SplitLeadAWHi}{+2.6}
\newcommand{\SplitSame}{five of the six}
\newcommand{\CtxInfoHead}{0.65}

\newcommand{\nTemplatesEight}{126}
\newcommand{\nTemplatesOne}{98}
\newcommand{\nOASteps}{520}
\newcommand{\nOAStepsOwl}{312}
\newcommand{\nOAStepsOther}{104}
\newcommand{\nBudgetB}{5\,600}
\newcommand{\nBudgetBL}{1\,900}
\newcommand{\nFamSingle}{eight}
\newcommand{\nFamCross}{three}
\newcommand{\nFamShared}{five}

\newcommand{\LrTenDeltaP}{$-$3.3}
\newcommand{\LrTenGainP}{+1.0}
\newcommand{\LrThreeDeltaP}{$-$3.4}
\newcommand{\ElemDeltaP}{$-$3.5}
\newcommand{\ElemCostP}{0.7}
\newcommand{\StateStepLeadP}{3.6}
\newcommand{\LrTenDeltaF}{$-$3.2}
\newcommand{\LrTenGainF}{+1.0}
\newcommand{\LrThreeDeltaF}{$-$3.4}
\newcommand{\ElemDeltaF}{$-$3.5}
\newcommand{\ElemCostF}{0.7}
\newcommand{\StateStepLeadF}{3.4}
\newcommand{\StateStepShift}{+0.2}
\newcommand{\StateStepGain}{+1.6}
\newcommand{\HalfHiL}{+2.5}
\newcommand{\HalfLoL}{$-$1.8}
\newcommand{\HalfDiffL}{+4.3}
\newcommand{\HalfHiF}{+2.9}
\newcommand{\HalfLoF}{$-$2.1}
\newcommand{\HalfDiffF}{+4.9}
\newcommand{\HalfAgree}{twelve}
\newcommand{\HalfThrL}{0.377}
\newcommand{\HalfThrF}{0.394}
\newcommand{\ElemKappaRho}{0.99}

\titleformat*{\paragraph}{\bfseries\itshape}

\title{Not All Experience Belongs in the Weights: \\ Component Routing for Self-Improving GUI Agents}

\author{\name Beining Wu \email Wu.Beining@jacks.sdstate.edu \\
      \name Zihao Ding \email Zihao.Ding@jacks.sdstate.edu \\
      \name Jun Huang \email Jun.Huang@sdstate.edu \\
      \addr Department of Electrical Engineering and Computer Science, South Dakota State University}

\begin{document}

\maketitle

\begin{abstract}
Self-improving GUI agents keep the trajectories they produce and return them to the agent, by fine-tuning or by
retrieval into the prompt, and studies that compare the two destinations disagree. We attribute this
to the unit of experience: a trajectory bundles items with different properties, so a conclusion about the bundle
depends on its mix. To address this, (i)~we introduce \emph{component routing}, which splits the experience into
locators, procedures, state facts and lessons and sends each component to the context or to the weights,
compared on the same items across three backbone families, two environments and three seeds. One pool has two
destinations: locators and lessons win in the weights, procedures and state facts in the context.
(ii)~We fit a rule in two properties measured before any training, recurrence and state-conditionality; it
recovers the destination of a held-out backbone family in \SignAgree{} cells, two interventions move a component
toward the boundary, and routing by the rule beats every whole-trajectory baseline and, by \RoutedGain{} points on
average, the better single destination of each backbone.
(iii)~We identify how training and producer--consumer differences change the value of the two destinations:
note readout decreases after the same component is written into the weights, most for the items that recur most,
context gains increase with the information gap, and weights gains decrease with the policy gap. Code and data
will be released.
\end{abstract}

\begin{figure}[h]
  \centering
  \includegraphics[width=0.95\textwidth]{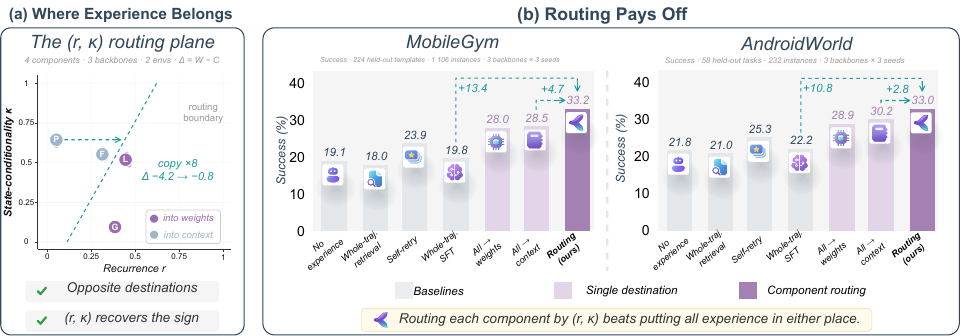}
  \caption{\textbf{Component routing.} \textbf{(a)}~The four components on the plane of recurrence $r$ and
  state-conditionality $\kappa$, measured before any training; fill marks the winning destination on held-out
  tasks, the dashed line is the fitted boundary, the arrow one intervention. \textbf{(b)}~Held-out success, mean
  over three backbones and three seeds; arrows: gain over whole-trajectory fine-tuning and the higher
  single-destination bar.}
  \label{fig:one}
\end{figure}

\section{Introduction}
\label{sec:intro}
GUI agents that operate phones and desktops are increasingly improved with their own traces: an agent runs a
batch of tasks, the trajectories are kept, and the experience is returned to the agent.
\begin{wrapfigure}{r}{0.40\textwidth}
  \vspace{-0.5\baselineskip}
  \centering
  \includegraphics[width=0.38\textwidth]{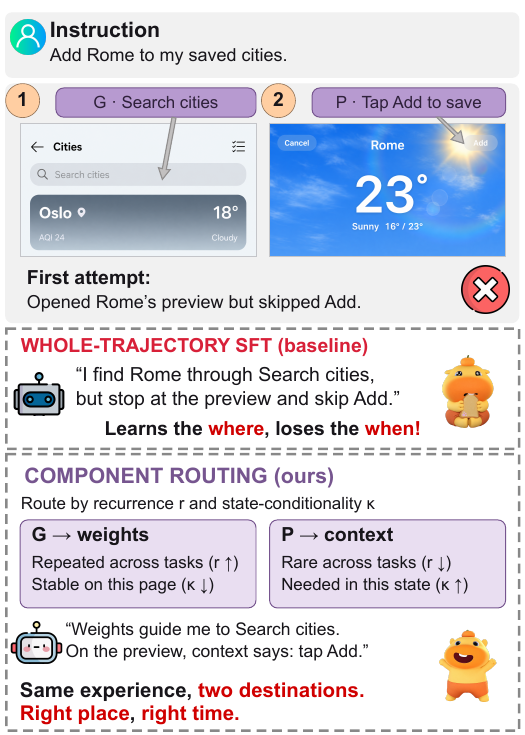}
  \caption{\textbf{A motivating example.} G recurs across tasks and goes to the weights; P matters only on the
  preview screen and stays in context. Screens are illustrative.}
  \label{fig:motivating}
  \wraptail\vspace{-12pt}
\end{wrapfigure}
Where that experience
should go is an open decision. One line of work fine-tunes the agent on its successful trajectories; another keeps
the trajectories, or notes distilled from them, in a library that is retrieved into the prompt.
On MobileGym, an 8B agent whose whole pool goes to the weights gains \AllWGain{} points of held-out success and one whose whole
pool goes to the context gains \AllCGain{}, yet the two destinations disagree on which parts of the pool are
responsible: the components that gain most in the weights are the ones that gain least in context
(Figure~\ref{fig:one}a). Figure~\ref{fig:motivating} shows the smallest version of this. One trajectory on a
weather application teaches where the search field is and that a city is saved by tapping Add on its preview;
fine-tuning on the whole trajectory learns the first and loses the second.

We attribute this to the unit of experience. Existing methods treat a trajectory as the unit: success-filtered
fine-tuning distills whole trajectories into the weights \citep{Zhang2026ICML,Men2026ACL,Fan2026ARXIVa}, and
retrieval keeps whole trajectories or trajectory-level skills in context
\citep{Xiong2026ACL,Hu2026ARXIV,Sui2026ARXIV}. Comparisons of the two destinations reach opposite
conclusions: internalized hints beat the prompt on tool tasks \citep{Alakuijala2026TMLR}, prompt-side updates beat
gradient updates on web tasks \citep{Li2026ICML,Agrawal2026ICLR}, and the same skill representation helps one task
and hurts another \citep{Liu2026ARXIV}. A trajectory bundles items with different properties, so any conclusion
about the bundle depends on its mix.

Deciding where a piece of experience belongs requires two things: a property of the piece that can be measured
before any training, and a comparison in which both destinations consume the same items and are read on the same
held-out tasks. Existing comparisons pit one whole pipeline against another, so the items differ before the
destinations do \citep{Alakuijala2026TMLR,Gou2026ARXIV}, and none asks it per component.

\corequestion{Which components of an agent's own experience belong in its context and which in its weights, and
does the answer survive training and a change of backbone?}

We answer this question with \emph{component routing}. Experience is split into four components, locators,
procedures, state facts and lessons, and each component is characterized by two properties measured on the pool
before any training: \emph{recurrence}, how often its key comes up in unrelated tasks, and
\emph{state-conditionality}, how much its usefulness depends on the current state matching its source state. Each
component is then sent to the context, to the weights, or to both, on the same items and the same held-out
instances, across three backbone families and two environments.

Three observations follow. The four components split by destination: the route gain, weights minus context,
is \DeltaG{} points for locators and \DeltaL{} for lessons, \DeltaP{} for procedures and \DeltaF{} for state
facts (Figure~\ref{fig:one}a). The two properties account for the split: a rule fitted with the tested backbone
held out recovers the sign in \SignAgree{} cells, an eightfold training dose moves procedures toward the weights
by \DupShift{} points, and stripping the state condition from state facts costs \DestateDrop{}. Routing each
component by the rule exceeds the better single destination of each backbone by \RoutedGain{} points on average
over three backbones and two environments, and the reverse assignment by \AntiGapBoth{}
(Figure~\ref{fig:one}b).

\textbf{Contributions.} (i)~A measurement of where an agent's own experience belongs: four components, two
pool-level properties, and a paired comparison of the context and the weights on the same items across three
backbones, two environments and three seeds. (ii)~A routing rule in recurrence and state-conditionality that
recovers the winning destination of every component on a held-out backbone family, with two interventions that
move a component toward the boundary. (iii)~An account of how training, information gaps and policy gaps shape
experience reuse: the readout of high-recurrence notes falls to \UseDrop{} once the same items are in the weights,
most for the items that recur most, and the context route gains from an information gap while the weights route
loses to a policy gap.

\section{Related Work}
\label{sec:related}

\paragraph{Two destinations for experience.}
Self-improving agents return their own experience along two routes. \emph{Fine-tuning} writes it into the
weights \citep{Ding2026MNET,Ding2026ARXIVEASE,Ding2026ICDCS,Wu2026ARXIVLifecycle,Wu2026TMM,Wu2026TNSE,Wu2026ICDCS}, on the agent's rollouts \citep{Zhang2026ICML,Men2026ACL,Ding2026ICNC,Huang2025RACS,Huang2025TMC,Wu2023MPE,Wu2023ACCESS,Wu2025WASA}, on skills extracted from them
\citep{Fan2026ARXIVa,Wu2026ARXIVb}, or by distilling a helpful context into the parameters
\citep{Alakuijala2026TMLR,Gou2026ARXIV,He2026ARXIV}. \emph{Retrieval} keeps it in a store that is read into the
prompt \citep{Wu2026ARXIVCrystalMem,Wu2026ARXIVERRAND,Wu2026ARXIVForget}: trajectory and skill libraries \citep{Xiong2026ACL,Hu2026ARXIV,Wei2026ARXIV}, evolved prompts
\citep{Agrawal2026ICLR}, or logit-level updates without gradients \citep{Li2026ICML}. Comparisons of the two
disagree, fine-tuning leading on tool tasks \citep{Alakuijala2026TMLR} and the prompt on generalization and on
knowledge written in sequence \citep{Lampinen2025ARXIV,Wang2026ARXIVa}; theory ties the answer to the data
\citep{Chan2025ICLR}, and recent systems route by volatility \citep{Ji2026ARXIV} or run both routes under one
budget \citep{Kim2026ARXIVc}. All of them take a whole trajectory as the unit and change the pipeline with the
destination; component routing holds the items fixed and compares the two destinations per component.

\paragraph{Experience forms in GUI agents.}
GUI agents are trained on trajectories, human demonstrations at first \citep{Wang2025NEURIPS,Liu2026ICLR} and now
their own rollouts in verifiable environments \citep{Wu2026ARXIVa,Wang2026ARXIVb}. What is kept from a trajectory
varies: skills distilled into the weights \citep{Fan2026ARXIVa}, skill cards retrieved or distilled
\citep{Liu2026ARXIV}, hindsight-relabelled experience \citep{Men2026ACL}, and memories built to survive interface
change \citep{Sun2026ACL}. Diagnoses show that trajectory fine-tuning learns interface shortcuts as readily as
semantics \citep{Gu2026ICML} and that a form which helps grounding can hurt action prediction \citep{Liu2026ARXIV}.
Each of these works fixes one form and one destination. We split one pool into four components and let two
measured properties decide where each goes.

\paragraph{Producer, consumer and fragility.}
Whether self-generated experience helps depends on who produced it and who consumes it \citep{Ding2025IPCCC,Ding2026JSAC,Duan2023COMST,Fang2025ARXIV,Fang2025TON,Fang2025JSAC,Pan2023SCIS,Tchalla2026ARXIV,Wu2026IoTJ,Wu2026MNET,Wu2026TON,Wu2026COMST,Xing2025ACR}: a stronger agent can
specialize a smaller one through its weights \citep{Kim2026ARXIVa}, shared trajectory stores help consumers on
either side of the ability gap \citep{Kim2026ARXIVb}, synthesized tools help only models weaker than their
builder \citep{Lou2026COLM}, and the gain from a harness changes with base ability in opposite directions across
studies \citep{Lin2026ARXIV,Wei2026ARXIV}. Self-improvement loops are also fragile: memory gains move with seeds,
task order and backbone \citep{Ye2026ARXIVa}, much of what a module stores comes from failed or falsely successful
runs \citep{Hajimiri2026ARXIV,Advani2026ARXIV}, and repeated internalization collapses rather than compounds
\citep{Chen2026ARXIV}. These studies report the net effect of a producer or a consumer. We separate it into an
information gap and a policy gap, measured on one pool across three backbone families and a size ladder.

\section{Experimental Setup}
\label{sec:setup}

\begin{figure}[t]
  \centering
  \includegraphics[width=0.95\textwidth]{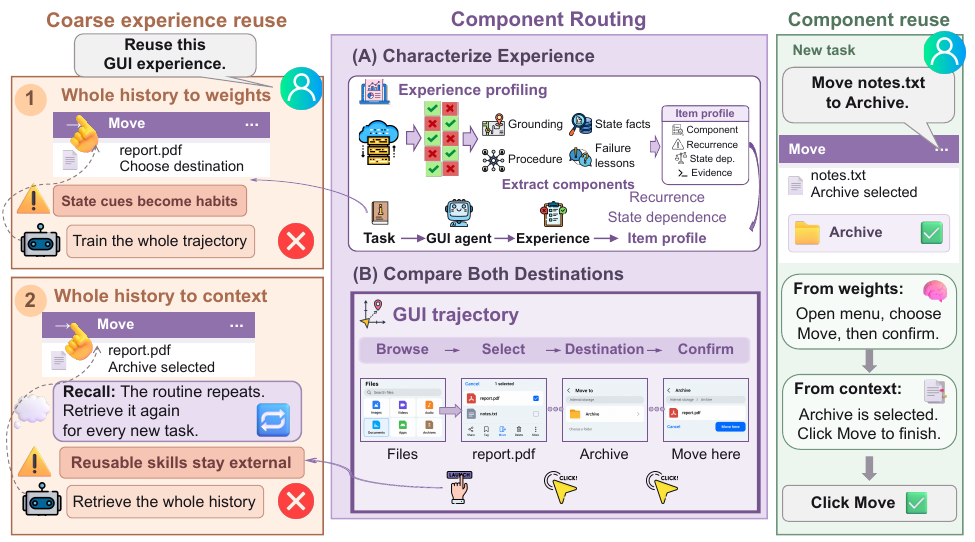}
  \caption{\textbf{Overview.} Whole-history reuse (left) against component routing: (A)~profile every item of the
  pool, (B)~compare both destinations per component (middle), and reuse each component from its destination on a
  new task (right).}
  \label{fig:framework}
\end{figure}

Component routing splits an agent's own experience into four components and sends each to the context or to the
weights of the same agent (Figure~\ref{fig:framework}). It treats a trajectory not as one unit but as a bundle of
items whose properties can be measured before any route is trained; Appendix~A gives the details.

\paragraph{Problem formulation.}
An agent with policy $\pi$ runs a batch of tasks on a fixed set of applications and keeps every trajectory,
successful or not; each step records the screenshot, the structured application state, the on-screen elements,
and the action with its target element. This batch is the \emph{experience pool} $E$. An extractor maps $E$ to
four item sets $D_c$, $c \in \{G, P, F, L\}$, where $G$, $P$, $F$ and $L$ denote locators, procedures, state facts
and lessons, and every item $d \in D_c$ carries a key $k(d)$ that identifies it across episodes. A destination
turns an item set into a policy: the \emph{context route} $\ccol{C_c}$ retrieves items of $D_c$ into the prompt
at each step, the \emph{weights route} $\wcol{W_c}$ fine-tunes $\pi$ on them, and the \emph{coexistence} arm
$\ccol{C_c}{+}\wcol{W_c}$ does both. Let $T$ be the set of held-out task instances and $u(x;\rho) \in \{0, 1\}$
the checker's verdict on instance $x$ under policy $\rho$, so that $U(\rho) = |T|^{-1}\sum_{x \in T} u(x;\rho)$
is held-out success. The quantity this paper measures is the \emph{route gain}
\begin{equation}
\begin{aligned}
\Delta(c) \;&\triangleq\; \hlw{U(\wcol{W_c})} \;-\; \hlc{U(\ccol{C_c})} \\
&=\; \frac{1}{|T|}\sum_{x \in T}\Bigl[\;\hlw{u\bigl(x;\wcol{W_c}\bigr)} \;-\; \hlc{u\bigl(x;\ccol{C_c}\bigr)}\;\Bigr],
\qquad \eqcmt{one paired comparison per instance}
\end{aligned}
\label{eq:delta}
\end{equation}
where $\Delta(c) > 0$ means that component $c$ is better off in the weights.

\paragraph{Components.}
Every item carries a key, a note that the context route serves and a sample that the weights route trains on. A
\emph{locator} says where a named element sits on a screen: its key is application, role and element name, and its
sample pairs the screenshot with a click on the element. A \emph{procedure} is a short run of action targets that
recurs inside a task family: its key is three consecutive targets, its note gives the subgoal sequence, and its
sample predicts the next action with the subgoals as the thought. A \emph{state fact} is a precondition that must
hold before an action succeeds, extracted from a failed and a successful attempt at the same target: its key is
application, target and predicate, and its sample predicts the action with the predicate as the thought. A
\emph{lesson} pairs an action that failed in a state with the one that later succeeded there: its key is
application, state and failed target, and its sample is the preference pair. A rule-based extractor produces
$D_c$.

\paragraph{State signature.}
The state at a step $x$ is summarized by a signature $S(x)$, the application route together with the set of
path--value pairs of the environment state without volatile fields such as timestamps; on AndroidWorld,
whose application state is not exposed, the signature is the set of visible accessibility nodes, the view a
deployed agent has (Appendix~\ref{app:splits}). Steps of one application are compared by the Jaccard
similarity $J(x, y) = |S(x) \cap S(y)| \,/\, |S(x) \cup S(y)|$;
a match is \emph{exact}, written $\mathrm{E}$, when $J(x, y) \geq \theta$ with $\theta = \jExact$, and
\emph{relaxed}, written $\mathrm{R}$, when $\theta' \leq J(x, y) < \theta$ with $\theta' = \jRelaxed$. Retrieval
and both properties use these tiers.

\paragraph{Recurrence and state-conditionality.}
For an item $d$, let $P_d$ be the episodes in which $d$ is produced, $N_p$ the episodes of the pool that run on
the same application as episode $p$ but on a different task template, and $K_q$ the set of keys of all items
extracted from episode $q$. \emph{Recurrence} is
\begin{equation}
\begin{aligned}
r(d) \;&\triangleq\; \frac{1}{|P_d|}\sum_{p \in P_d}\;\frac{1}{|N_p|}\sum_{q \in N_p} \ind\bigl[\,k(d) \in K_q\,\bigr],
\qquad \eqcmt{share of unrelated tasks where the key returns} \\
r(c) \;&\triangleq\; \frac{1}{|D_c|}\sum_{d \in D_c} r(d),
\end{aligned}
\label{eq:recurrence}
\end{equation}
where the component summary is a mean because the item distribution is right-skewed. \emph{State-conditionality}
is how much the usefulness of an item depends on the current state matching its source state. Let $\Omega_\tau(d)$ be the pairs of occurrences of $k(d)$ whose states match
at tier $\tau \in \{\mathrm{E}, \mathrm{R}\}$, and $\phi_d(x)$ the payload of $d$ observed at $x$, the position of
a locator or the next target of a procedure. Before any route is trained we use the proxy
\begin{equation}
\begin{aligned}
a_\tau(d) \;&\triangleq\; \frac{1}{|\Omega_\tau(d)|}\sum_{(x, y) \in \Omega_\tau(d)} \ind\bigl[\,\phi_d(x) \simeq \phi_d(y)\,\bigr],
\qquad \tau \in \{\mathrm{E}, \mathrm{R}\}, \\
\kappa_0(d) \;&\triangleq\; 1 - \frac{a_{\mathrm{R}}(d)}{a_{\mathrm{E}}(d)},
\qquad
\kappa_0(c) \;\triangleq\; \operatorname*{med}_{d \in \tilde{D}_c}\, \kappa_0(d),
\qquad \eqcmt{$\tilde{D}_c$: items with pairs at both tiers}
\end{aligned}
\label{eq:kappa0}
\end{equation}
where $\simeq$ is payload agreement, within \agreeFrac{} percent of the screen diagonal for a position and
identity for a target. Once the context route has run with exact-only retrieval $\ccol{C_c^{\mathrm{E}}}$ and
relaxed-only retrieval $\ccol{C_c^{\mathrm{R}}}$, the online quantity
\begin{equation}
\kappa(c) \;\triangleq\; \frac{\hlc{U(\ccol{C_c^{\mathrm{E}}})} \;-\; \hlc{U(\ccol{C_c^{\mathrm{R}}})}}
{\hlc{U(\ccol{C_c^{\mathrm{E}}})} \;-\; U_0},
\qquad \eqcmt{share of the context gain that needs an exact state match}
\label{eq:kappa}
\end{equation}
confirms the proxy, where $U_0 = U(\pi)$ is the success of the agent without experience. Both $r$ and $\kappa_0$
come from the pool alone, so a practitioner has them before any training decision.

\paragraph{Two destinations.}
The context route keeps $D_c$ in a library indexed by application and key. At a step $x$, the items of the same
application are ranked by $J(x, \cdot)$ against their source states, and the \kNotes{} best are appended to the
prompt as notes, exact matches first; the rest of the prompt is the base agent's, character for character. The
weights route fine-tunes the backbone on the samples of the same items with low-rank adaptation
\citep[LoRA;][]{Hu2022ICLR}, one configuration for all backbones and arms (Appendix~\ref{app:training}), by direct
preference optimization \citep{Rafailov2023NEURIPS} for lessons and supervised fine-tuning for the other
components; every trained arm reaches a token accuracy of at least \devAcc{} on a held-in development split. The
tuned model is merged and served with the base agent's prompt, and $\ccol{C_c}{+}\wcol{W_c}$ serves it with the
same notes.

\paragraph{Reference arms.}
Every backbone also runs without experience and with undivided whole trajectories: success-filtered
fine-tuning on entire trajectories, retrieval of one similar trajectory, retrieval of \kAtomic{} step-level records
by task similarity, and context-then-distill. Three controls probe the instrument: random notes from other
applications, notes with shuffled keys and contents, and training samples with shuffled labels. A retry arm keeps
the first of \nRetry{} attempts that the agent declares complete, and three aggregate arms send all components to
context, all to the weights, or each to the destination the fitted rule assigns (Appendix~\ref{app:arms}).

\paragraph{Evaluation setup.}
We evaluate on MobileGym \citep{Wu2026ARXIVa} and AndroidWorld \citep{Rawles2025ICLR}. MobileGym is a
simulated phone with \nApps{} applications and
\nTemplates{} task templates (\nTrainTemplates{} training, \nTestTemplates{} test) whose checker compares the
application state to the task target; the same state is readable at every step, which makes the keys
above well defined. A fixed rank rule holds out \nUnseenApps{} applications; the remaining training templates form
the pool (\nExpTemplates{} templates), and the test templates form two held-out tiers, new templates on practiced
applications (\nSeenTemplates{}) and templates on unseen applications (\nUnseenTemplates{}), with up to
\kInstances{} instances per template and seed (\nPairsPerSeed{} paired instances per seed). AndroidWorld has
\nAWTasks{} tasks on \nAWApps{} real applications and an accessibility tree; \nAWSeenApps{} applications provide
experience, \nAWUnseenApps{} are held out, and \nAWHeldoutTasks{} held-out tasks are run with \kAWInstances{}
instances each (\nAWPairsPerSeed{} paired instances per seed). Held-out success is decided by each environment's deterministic
checker, with no model judge. Every arm of a backbone is evaluated on the same instances, and differences are
read with a paired bootstrap over instances (\nBoot{} resamples, 95 percent intervals), pooled over the two tiers
and per tier. Producer and consumer are the same model from \nBackbones{} families in the 7--8B
range, \llmname{GUI-Owl-1.5-8B} \citep[\llmname{GUI-Owl-8B} below;][]{Xu2026ARXIV}, \llmname{ScaleCUA-7B} \citep{Liu2026ICLR} and
\llmname{Qwen3-VL-8B} \citep{Bai2025ARXIV}; a same-family ladder, \llmname{Qwen3-VL} 4B, 8B and 32B, serves the
producer--consumer measurements. Pool sampling, training and
instance sampling are controlled by \nSeeds{} seeds, and the analysis of every result was fixed before the runs.
Splits, instances and cost are in Appendices B and H.

\section{Observations}
\label{sec:obs}

This section reports what the pool and the two destinations show before any rule is fitted. Every gain is a
paired difference on the same held-out instances, with 95 percent intervals from the paired bootstrap.

\begin{figure}[t]
  \centering
  \includegraphics[width=\textwidth]{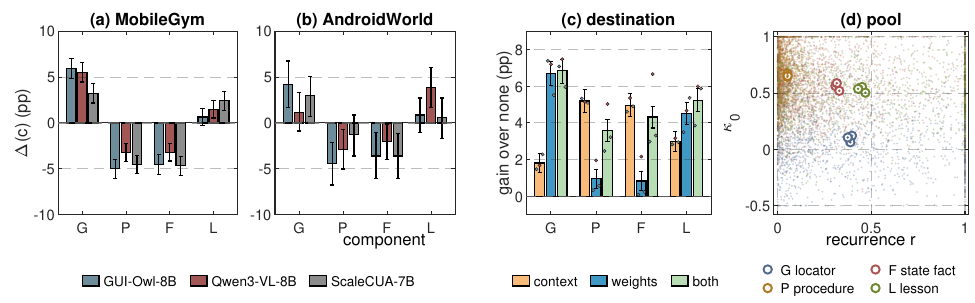}
  \caption{\textbf{Components split by destination.} (a,~b)~Route gain $\Delta(c)$ per backbone on MobileGym and
  AndroidWorld, with 95 percent intervals; (c)~gain of each destination over the base agent, pooled over
  backbones, one dot per backbone; (d)~the pool on the plane of recurrence and state-conditionality, with the
  component summary of each backbone.}
  \label{fig:obs1}
\end{figure}

\subsection{Components Split by Destination}
\label{sec:obs1}

\paragraph{Route gains by component.}
Figure~\ref{fig:obs1}(a,~b) shows the route gain of each component, backbone and environment. Locators are better
off in the weights on every backbone and in both environments (\DeltaG{} points pooled over backbones on
MobileGym), procedures and state facts are better off in the context (\DeltaP{} and \DeltaF{}), and lessons lean
toward the weights by a margin whose interval includes zero in half of the cells (\DeltaL{}). The sign pattern
does not depend on the backbone: the twelve MobileGym cells agree across the three families, and the
AndroidWorld cells, whose state signature is read from the screen, repeat the pattern with the wider intervals of
a smaller task set. \emph{Both destinations help every component.} Against the base agent and
pooled over backbones (Figure~\ref{fig:obs1}c), the context gains between \GainCG{} and \GainCP{} points and the
weights between \GainWF{} and \GainWG{}; what differs is which route helps more, the weights for locators
(\GainWG{} against \GainCG{}) and the context for procedures (\GainCP{} against \GainWP{}). \emph{Coexistence adds little where the weights already win.}
Serving the notes on top of the tuned model adds \CoexG{} points for locators and \CoexF{} for state facts, so
the context remains useful exactly for the components that the context wins. \emph{The components differ before
any training.} Figure~\ref{fig:obs1}(d) places the items on the plane of the two properties. Procedures almost
never recur outside their task family (mean recurrence \rP{}) and are state-conditioned (\kP{}); locators recur
(\rG{}) and are nearly state-free (\kG{}); state facts and lessons recur (\rF{} and \rL{}) and are
state-conditioned (\kF{} and \kL{}). The three backbones place every component in the same region of the plane.

\obs{Split by component, one pool has opposite winning destinations: locators, and weakly lessons, in the
weights; procedures and state facts in the context; on three backbone families and two environments.}

\Needspace{23\baselineskip}
\subsection{Producer and Consumer}
\label{sec:obs2}

\begin{wrapfigure}{r}{0.60\textwidth}
  \vspace{-1.4\baselineskip}
  \centering
  \includegraphics[width=0.58\textwidth]{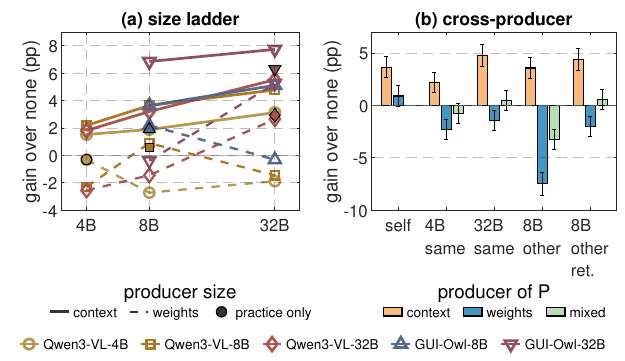}
  \caption{\textbf{Producer and consumer.} (a)~Gain from the procedures of a producer of each size, read from the
  context (solid) or written into the weights (dashed); filled markers train on the consumer's own procedures
  only. (b)~One consumer, five producers, three destinations.}
  \label{fig:producer}
  \wraptail
\end{wrapfigure}
\paragraph{Producers and consumers.}
Figure~\ref{fig:producer}(a) follows the Qwen3-VL family across sizes, in a batch where every procedures adapter
trains on the same number of samples (Appendix~\ref{app:splits}). Procedures read from the context help every
consumer whoever produced them, and larger producers help more: the 8B consumer gains \SelfC{} points from its
own procedures, \DownC{} from those of the 32B model and \SmallC{} from those of the 4B model, and the 32B
consumer still gains \UpC{} from the procedures of the 8B model. The same items written into the weights follow a
different rule. Fine-tuning the 8B consumer on the procedures of the 4B model changes it by \SmallW{}
points, on those of the 32B model by \DownW{}, and on those of another family by \CrossW{}
(Figure~\ref{fig:producer}b), while its own procedures give \SelfW{}. \emph{The damage is policy mismatch, not information.} The same other-family pool read
from the context gives \CrossC{}, and a producer of the other family retrained on the consumer's own successes
brings the weights route back to \CrossWb{}. \emph{Most of the weights gain on own procedures is practice.} Fine-tuning on the
consumer's own procedures alone, without any transferred item, gives \PracticeGain{} against \SelfW{} with all
samples, and the two agree within \PracticeDiff{} points on all five consumers (Figure~\ref{fig:producer}a,
filled markers), so the weights route pays for what the consumer already does more than for what it learns.

\obs{Notes help regardless of who produced them, and more so from a stronger producer; fine-tuning helps only when
producer and consumer are the same model, and most of its gain is practice on the consumer's own trajectories.}

\Needspace{21\baselineskip}
\subsection{Readout after Training}
\label{sec:obs3}

\begin{wrapfigure}{r}{0.60\textwidth}
  \vspace{-\baselineskip}
  \centering
  \includegraphics[width=0.58\textwidth]{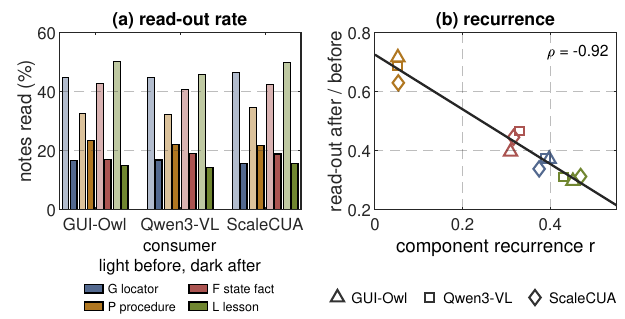}
  \caption{\textbf{Readout after training.} (a)~Share of probed steps whose action changes after a note is
  read, before and after the same items are in the weights; (b)~the after-to-before ratio against the
  component's recurrence.}
  \label{fig:readout}
  \wraptail
\end{wrapfigure}
\paragraph{Readout probe.}
The readout probe replays a decision step with and without a retrieved note and records whether the action
changes. Before training, a note changes the action in \ReadBefore{} of the probed steps, from \ReadBeforeP{} for
procedures to \ReadBeforeL{} for lessons, at about the same rate on every consumer (Figure~\ref{fig:readout}a).
After the same items are in the weights, the rate falls to \ReadAfterG{} for locators and \ReadAfterL{} for
lessons but only to \ReadAfterP{} for procedures, and the after-to-before ratio falls with the recurrence of the
component (Figure~\ref{fig:readout}b, Spearman \ReadRho{}). The same holds among the items of one component: split
into thirds by recurrence, the ratio falls from \ReadItemGLo{} to \ReadItemGHi{} for locators and from
\ReadItemLLo{} to \ReadItemLHi{} for lessons (Table~\ref{tab:app_readout}). \emph{Redundancy rather than
conflict.} The drop is
largest for lessons and locators, the two components that recur most and for which coexistence added the least
above, and smallest for procedures, the component for which the context keeps its value after training. A note is read when
it carries something the weights do not yet hold, and a component that recurs is the one the weights absorb.

\obs{Training on a component makes its notes redundant, and the more the component recurs, the less its notes are
read afterwards.}

\section{The Routing Rule}
\label{sec:rule}

This section fits a rule in the two pool measurements, tests it on held-out backbone families, and examines
interventions, training and version changes.

\subsection{A Rule in Two Properties}
\label{sec:rule1}

\paragraph{Held-out fit.}
Each component has one destination across backbone families, and a linear rule in the two properties recovers
it on a family it has not seen. We score a component by its recurrence against its state-conditionality and fit the score to the sign of
the route gain, holding out one family at a time. With $\mathcal{C}$ the twenty-four cells of both
environments, $\mathcal{C}_f$ the eight cells of family $f$ and $I(c)$ the interval of $\Delta(c)$,
\begin{equation}
\begin{aligned}
s_\theta(c) \;&\triangleq\; \hlw{\alpha\, r(c)} \;-\; \hlc{\beta\, \kappa_0(c)} \;-\; \gamma,
\qquad \theta = (\alpha, \beta, \gamma), \qquad
w(c) \;\triangleq\; 1 - \tfrac{1}{2}\,\ind\bigl[\,0 \in I(c)\,\bigr], \\
\theta_f \;&\triangleq\; \operatorname*{arg\,min}_{\theta}\;
\sum_{c \,\in\, \mathcal{C} \setminus \mathcal{C}_f} w(c)\,
\log\Bigl(1 + \exp\bigl[-\operatorname{sgn}\Delta(c)\; s_\theta(c)\bigr]\Bigr)
\quad \eqcmt{sixteen held-in cells per fold} \\
S \;&\triangleq\; \sum_{f=1}^{3}\;\sum_{c \,\in\, \mathcal{C}_f} w(c)\;
\ind\bigl[\operatorname{sgn} s_f(c) = \operatorname{sgn}\Delta(c)\bigr],
\quad \eqcmt{held-out score, at most 24}
\end{aligned}
\label{eq:rule}
\end{equation}
where $s_f$ is the score under $\theta_f$, the purple term pulls a component toward the weights and the orange
term toward the context, a cell whose interval crosses zero counts one half in fit and score, and the
fit carries a ridge penalty on the standardised inputs (Appendix~\ref{app:rule}). A held-out cell goes to $\wcol{W}$ if
$s_f(c) > 0$, else to $\ccol{C}$.

\Needspace{22\baselineskip}
\begin{wrapfigure}{r}{0.60\textwidth}
  \vspace{-0.4\baselineskip}
  \centering
  \includegraphics[width=0.58\textwidth]{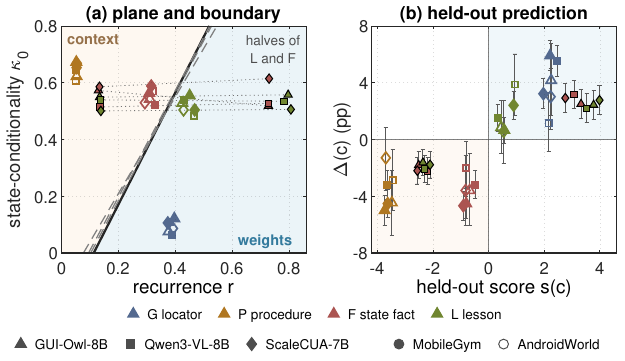}
  \caption{\textbf{The rule.} (a)~Component summaries of both environments on the plane, with the full-fit boundary
  (solid) and the three held-out boundaries (dashed); (b)~held-out score against the observed route gain, 24 cells
  with 95 percent intervals. Small black-edged markers: halves of lessons and state facts split by item
  recurrence, used in no fit.}
  \label{fig:rule}
  \wraptail
\end{wrapfigure}
The three folds give $\alpha$ in $[\RuleAlphaLo, \RuleAlphaHi]$, $\beta$ in $[\RuleBetaLo, \RuleBetaHi]$ and
$\gamma$ in $[\RuleGammaLo, \RuleGammaHi]$, and the full fit $(\RuleAlpha, \RuleBeta, \RuleGamma)$ draws the
boundary of Figure~\ref{fig:rule}(a). The held-out rule recovers the sign in \SignAgree{} cells, \SignClear{} of
them with intervals clear of zero (Figure~\ref{fig:rule}b), and $S = \SignScore{}$. \emph{The boundary runs
between state facts and lessons.} Locators and procedures are far from it (full-fit scores \ScoreG{} and
\ScoreP{}) and state facts and lessons close to it (\ScoreF{} and \ScoreL{}), and the scores order the four
components as their route gains do. \emph{The boundary also splits a component.} Split at the boundary by item
recurrence ($r = \HalfThrL{}$ for lessons, \HalfThrF{} for state facts), the half above goes to the weights and the
half below to the context: the route gains are \HalfHiL{} and \HalfLoL{} for lessons and \HalfHiF{} and
\HalfLoF{} for state facts, and all \HalfAgree{} halves of the three families fall on the side the rule gave them
before the runs (Figure~\ref{fig:rule}, small markers; Table~\ref{tab:app_halves}). \emph{The online
quantity agrees with the proxy.} The state-conditionality read from retrieval, $\kappa(c)$ of
Equation~\ref{eq:kappa}, agrees in rank with $\kappa_0(c)$ over the twelve MobileGym cells (Spearman
\KappaOnline{}) and puts procedures and state facts above lessons and locators on every backbone.

\subsection{Interventions}
\label{sec:rule2}

\begin{table}[t]
\centering
\caption{\textbf{Interventions.} Gain over the base agent (points) on MobileGym when one property of a component is changed: copying the training items raises the training dose of every key, as recurrence across tasks does; retrieving the notes by task text alone removes their state condition. Subscripts are the std over three seeds. $\Delta$ is the shift from the unmodified route above it (\up{rise} / \down{drop}), averaged over the backbones the arm has.}
\label{tab:interventions}
\footnotesize
\setlength{\tabcolsep}{4.5pt}
\renewcommand{\arraystretch}{1.15}
\begin{tabular}{l|ccc|c}
\toprule
& \multicolumn{3}{c|}{\cellcolor{SearchBg}\textbf{MobileGym, gain over the base agent}} & \\
\cmidrule(lr){2-4}
\textbf{Arm} & \textbf{GUI-Owl-8B} & \textbf{Qwen3-VL-8B} & \textbf{ScaleCUA-7B} & \textbf{$\Delta$} \\
\midrule
Procedures to the context & +5.3\std{0.7} & +5.2\std{0.2} & +5.1\std{0.0} & -- \\
\quad state condition removed & +1.4\std{0.8} & +1.8\std{0.1} & +0.8\std{0.4} & \down{3.9} \\
Procedures to the weights & +0.3\std{0.6} & +2.0\std{0.9} & +0.6\std{0.3} & -- \\
\quad items copied $\times 8$ & +3.3\std{0.6} & +6.1\std{0.3} & +3.9\std{0.5} & \up{3.5} \\
\midrule
State facts to the context & +4.6\std{0.4} & +5.4\std{0.6} & +4.9\std{0.3} & -- \\
\quad state condition removed & +1.1\std{0.7} & +3.1\std{0.5} & +0.6\std{0.9} & \down{3.4} \\
State facts to the weights & +0.1\std{0.1} & +2.2\std{0.8} & +0.2\std{0.3} & -- \\
\quad items copied $\times 8$ & +4.0\std{1.2} & +5.8\std{0.6} & +3.4\std{1.3} & \up{3.6} \\
\midrule
Random notes from other applications & $-$1.5\std{0.6} & $-$1.7\std{0.3} & $-$1.6\std{0.6} & -- \\
\bottomrule
\end{tabular}
\end{table}

\paragraph{One property at a time.}
Changing one property moves a component the way the rule says. Table~\ref{tab:interventions} reports the arms that
change one property of a component, with the shift from the unmodified route in its last column.
\emph{A larger training dose moves procedures toward the weights.} A key that recurs in more tasks contributes more
samples to the weights route, so recurrence reaches the weights as the number of updates a key receives. Copying
every procedure item eightfold before fine-tuning reproduces that dose and changes no item: it moves the route
gain of procedures from \DeltaP{} to \DupEight{}, across zero on \llmname{Qwen3-VL-8B} (\DupEightQwen{}). Read
as recurrence, eight copies place procedures at $r \approx \DupR{}$, short of the boundary at their
state-conditionality ($r \approx \DupBoundaryR{}$), and the route gain stays below zero on average. State facts
move by \DupShiftF{} and reach the context route on \llmname{Qwen3-VL-8B}, and with two and four copies the gain
grows with the dose (Table~\ref{tab:interventions_full}). At ten times the learning rate the weights route changes
by \LrTenGainP{} and \LrTenGainF{} points, and the route gains of procedures and state facts stay at \LrTenDeltaP{}
and \LrTenDeltaF{}.
\emph{Removing the state condition removes most of the context gain.} Serving the same notes retrieved by task text
alone, without their state condition, costs \DestateDrop{} points for state facts and \DestateDropP{} for
procedures, so what the context route buys for these two components is the condition, not the payload. Reading
the condition from the visible screen elements instead of the environment state keeps most of it: the notes lose
\ElemCostP{} points for procedures and \ElemCostF{} for state facts, and both stay in the context (route gains
\ElemDeltaP{} and \ElemDeltaF{}).
\emph{Shuffled items do
nothing.} Shuffling the keys of the notes or of the training samples leaves the base agent within \ShuffleMax{}
points (Table~\ref{tab:interventions_full}), and random notes from other applications change it by \RandomNotes{}; neither the prompt format nor the
fine-tuning itself produces the gains.

\subsection{Readout and Conflict}
\label{sec:rule3}

\begin{figure}[t]
  \centering
  \includegraphics[width=0.95\textwidth]{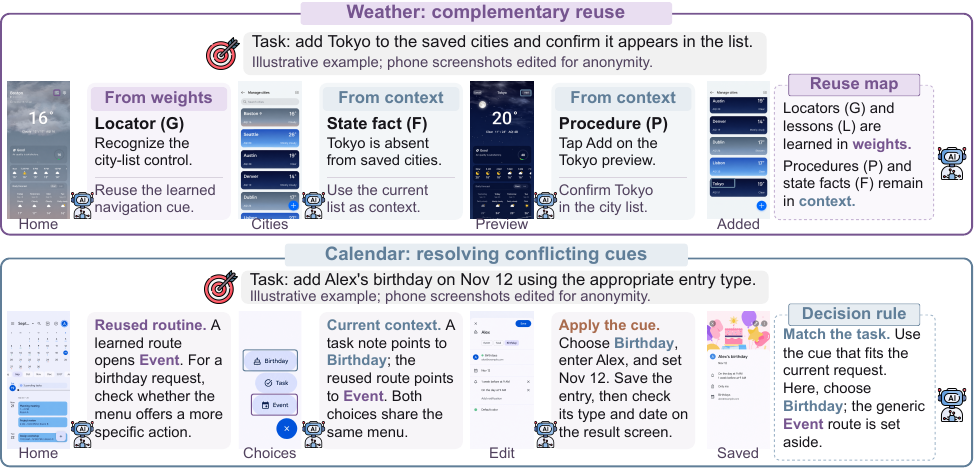}
  \caption{\textbf{Two illustrative episodes.} Top: the locator comes from the weights, the state fact and
  the procedure from the context. Bottom: a note and a learned routine disagree on the entry type, and the agent
  takes the action of the note. Screens are illustrative.}
  \label{fig:case}
\end{figure}
\paragraph{Stale weights, fresh notes.}
When the weights and the context disagree, the more an item recurs, the less often the agent follows the note and
the more often it takes neither action. To see what happens when the two disagree, we train the weights on version~1 of
the OpenApps applications \citep{Ullrich2026ICLR}, serve notes from version~2, and record, at the first step of a task where the two
versions call for different actions, whether the agent follows the note, the weights or neither
(Figure~\ref{fig:case}, bottom, shows one such step). On the new version, stale weights alone change the base agent by \OAStaleW{}
points and fresh notes alone by \OAFreshC{}; serving both scores between \OAConflictLo{} and \OAConflictHi{}
points below notes alone. \emph{Recurrence shifts the outcome.} At the divergence step the agent follows the note in
\ConflictCtxLow{} of the low-recurrence cases and \ConflictCtxHigh{} of the high-recurrence ones, the weights in
\ConflictWLow{} and \ConflictWHigh{}, and neither in \ConflictNeitherLow{} and \ConflictNeither{}. This is the
readout result from the other side: the items the weights hold best are the items a contradicting note corrects
least, and when the weights hold a stale version of such an item, the agent more often takes neither action.

\subsection{Routing against the Alternatives}
\label{sec:rule4}

\begin{table}[t]
\centering
\caption{\textbf{Routing against the alternatives.} Held-out success (percent) on the same held-out instances with \llmname{GUI-Owl-8B}, \llmname{Qwen3-VL-8B} and \llmname{ScaleCUA-7B}; subscripts are the std over three seeds. Self-retry uses \nRetry{} attempts per instance. The best value in each family is \underline{underlined}; the gain of routing over each row, averaged over the three MobileGym backbones, appears in the \colorbox{DeltaBg}{$\Delta$ column} (points; shade $\propto$ gain).}
\label{tab:main}
\footnotesize
\setlength{\tabcolsep}{2.8pt}
\renewcommand{\arraystretch}{1.12}
\begin{tabular}{l|ccc|ccc|c}
\toprule
& \multicolumn{3}{c|}{\cellcolor{SearchBg}\textbf{MobileGym}} & \multicolumn{3}{c|}{\cellcolor{ScienceBg}\textbf{AndroidWorld}} & \\
\cmidrule(lr){2-4}\cmidrule(lr){5-7}
\textbf{Arm} & \textbf{GUI-Owl} & \textbf{Qwen3-VL} & \textbf{ScaleCUA} & \textbf{GUI-Owl} & \textbf{Qwen3-VL} & \textbf{ScaleCUA} & \textbf{$\Delta$} \\
\midrule
Base agent & 21.6\std{0.9} & 16.7\std{2.0} & 19.2\std{2.2} & 23.6\std{1.9} & 17.8\std{2.7} & 24.0\std{3.0} & \mygreen{56}\gain{14.1} \\
\midrule
Self-retry, three attempts & \underline{26.4\std{0.7}} & \underline{20.6\std{1.6}} & \underline{24.6\std{1.3}} & \underline{26.1\std{0.7}} & \underline{22.7\std{3.9}} & \underline{27.2\std{4.0}} & \mygreen{37}\gain{9.3} \\
Success-filtered fine-tuning & 22.2\std{0.1} & 17.3\std{1.4} & 20.0\std{2.4} & 24.1\std{2.0} & 19.3\std{4.5} & 23.3\std{3.2} & \mygreen{53}\gain{13.4} \\
Retrieve one trajectory & 21.0\std{0.4} & 15.1\std{2.0} & 18.1\std{1.5} & 22.8\std{4.1} & 18.1\std{3.1} & 22.1\std{2.3} & \mygreen{61}\gain{15.2} \\
Retrieve ten step records & 22.9\std{1.1} & 17.9\std{2.1} & 20.8\std{2.2} & 24.3\std{2.4} & 19.5\std{3.6} & 25.3\std{3.2} & \mygreen{51}\gain{12.6} \\
Context then distill: procedures & 23.8\std{1.1} & 19.8\std{2.2} & 21.8\std{2.9} & 25.9\std{3.0} & 19.3\std{1.6} & 24.6\std{2.4} & \mygreen{46}\gain{11.4} \\
Context then distill: state facts & 23.7\std{0.3} & 18.4\std{1.9} & 21.8\std{3.1} & 25.1\std{3.2} & 22.1\std{3.9} & 25.3\std{1.6} & \mygreen{48}\gain{11.9} \\
\midrule
All to the context & \underline{29.6\std{0.6}} & 27.3\std{2.0} & \underline{28.7\std{2.1}} & \underline{33.6\std{4.6}} & 28.0\std{2.3} & 29.0\std{3.3} & \mygreen{19}\gain{4.7} \\
All to the weights & 28.6\std{0.6} & \underline{28.0\std{2.6}} & 27.3\std{2.0} & 29.0\std{1.9} & \underline{28.2\std{0.9}} & \underline{29.5\std{2.5}} & \mygreen{21}\gain{5.2} \\
\midrule
\rowcolor{oursrowcolor}
\textbf{Routed (Ours)} & \textbf{34.8\std{0.8}} & \textbf{30.5\std{2.3}} & \textbf{34.3\std{2.4}} & \textbf{33.3\std{1.8}} & \textbf{33.2\std{0.9}} & \textbf{32.6\std{1.8}} & -- \\
\midrule
Reverse assignment & 25.2\std{1.2} & 24.4\std{2.0} & 23.2\std{2.3} & 27.7\std{3.8} & 23.4\std{2.1} & 27.9\std{2.4} & \mygreen{36}\gain{8.9} \\
\bottomrule
\end{tabular}
\end{table}

\paragraph{Against the alternatives.}
Routing by the held-out rule beats every whole-trajectory baseline and, on average over backbones, the better
single destination. Table~\ref{tab:main} compares them on the same held-out instances, with the gain of routing
over each row on MobileGym in its last column (compute and tokens in Appendix~\ref{app:compute}). Routed sends
each component where the held-out rule says, so its assignment on a backbone is fixed before that backbone is
trained. Fitted on one half of the held-out templates instead, the rule gives the same assignment in \SplitSame{}
folds, and routing leads the better single destination on the other half by \SplitLeadMGLo{} to
\SplitLeadMGHi{} points on MobileGym and \SplitLeadAWLo{} to \SplitLeadAWHi{} on AndroidWorld
(Table~\ref{tab:app_tsplit}). On the full set it exceeds the better single destination on every MobileGym
backbone, by \RoutedGainMG{} points on average; on AndroidWorld it exceeds it by \RoutedGainAWQwen{} with
\llmname{Qwen3-VL-8B} and by \RoutedGainAWScale{} with \llmname{ScaleCUA-7B}, and matches it with
\llmname{GUI-Owl-8B}. \emph{The single
destinations fail on different components.} All to the context forgoes the weights gain of locators, all to the
weights the context gain of procedures and state facts; routing keeps both (Figure~\ref{fig:case}, top).
\emph{Components carry what whole trajectories do not.} The best whole-trajectory baseline, three sampled attempts
with self-retry, is \RoutedVsTraj{} points below routing, and fine-tuning on whole trajectories, retrieval of one
trajectory or of ten step records, and context-then-distill all stay below the single destinations. The lead comes
from what the components extract: localization samples for locators, preference pairs drawn from failed attempts
for lessons, and notes matched to the current state for procedures and state facts. Without the state match, the
notes of procedures and state facts gain \DestateGainP{} and \DestateGainF{}, as ten step records retrieved by task
text do (\AtomicGain{}); retrieved by the state signature instead, the same records gain \StateStepGain{}
(Table~\ref{tab:interventions_full}), still
\StateStepLeadP{} and \StateStepLeadF{} points below the notes of procedures and state facts. Fine-tuning on whole
trajectories gains \SFTGain{}, about as much as procedures in the weights (\GainWP{}). \emph{The assignment is what matters.} The reverse assignment loses \AntiGap{} points against
routing on MobileGym and \AntiGapAW{} on AndroidWorld, and the in-sample rule assigns the same routes, so routing
by the full fit differs by \RoutedSelfGap{}.

\section{Producer and Consumer Shifts}
\label{sec:shifts}

Observation~2 found that notes travel from a producer to a consumer and fine-tuning samples do not. This section
measures three gaps between the two models before any transfer, shows which gap each route pays for, and reads
off what survives a change of producer or of consumer.

\begin{figure}[t]
  \centering
  \includegraphics[width=\textwidth]{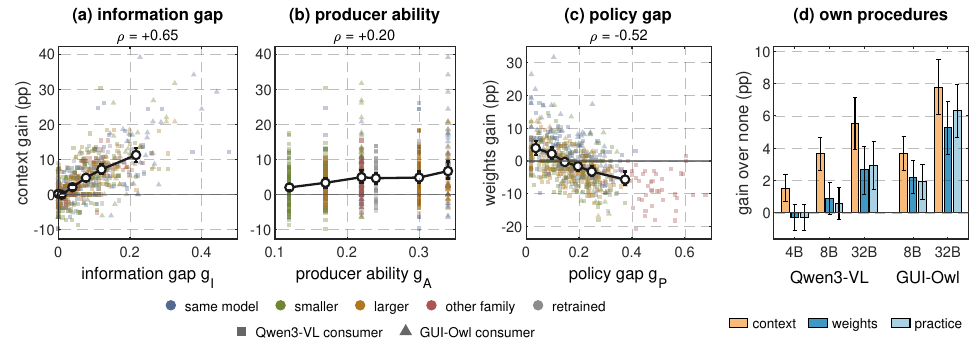}
  \caption{\textbf{The three gaps.} (a--c)~Family-level gain of the procedures route against the information gap,
  the producer's ability and the policy gap: \nFamilyCells{} cells, binned means with 95 percent intervals from a
  bootstrap over the \nProducerPairs{} producer--consumer pairs. (d)~Five consumers on their own procedures:
  context, weights, and weights from own successes only.}
  \label{fig:gaps}
\end{figure}

\subsection{Three Gaps}
\label{sec:gaps}

\paragraph{Gaps and gains.}
The context route pays for information and the weights route for policy agreement. For a producer $p$, a consumer $q$ and a task family $\varphi$, let $\sigma_\varphi(m)$ be the pass rate of
model $m$ on the family in its own pool, $X(p,q)$ the states the two pools share under exact matching, $a_m(x)$
the action of $m$ at state $x$, $U_\varphi$ the held-out success on the family and $U_0(m)$ that of model $m$
alone. The three gaps, the two gains and the practice term are
\begin{equation}
\begin{aligned}
g_I(p,q,\varphi) \;&\triangleq\; \sigma_\varphi(p)\,\bigl[1-\sigma_\varphi(q)\bigr], \qquad
g_P(p,q) \;\triangleq\; \frac{1}{|X(p,q)|}\sum_{x\in X(p,q)} \ind\bigl[\,a_q(x)\neq a_p(x)\,\bigr], \\
g_A(p) \;&\triangleq\; U_0(p), \qquad
\hlc{G_C(p,q,\varphi)} \;\triangleq\; U_\varphi(\ccol{C_P^{\,p}}) - U_\varphi(q), \qquad
\hlw{G_W(p,q,\varphi)} \;\triangleq\; U_\varphi(\wcol{W_P^{\,p}}) - U_\varphi(q), \\
\hlw{P(q)} \;&\triangleq\; U(\wcol{W_P^{\dagger}}) - U_0(q),
\qquad \eqcmt{$q$ trained on its own successes only, from families it already solves}
\end{aligned}
\label{eq:gaps}
\end{equation}
so that $g_I$ is the share of the family that the producer solves and the consumer does not, $g_P$ the rate at
which the consumer would act differently from the producer on the same state, and $g_A$ the producer's own
success. Figure~\ref{fig:gaps} places the \nFamilyCells{} cells of \nProducerPairs{} producer--consumer pairs on
\nFamilies{} task families against the three gaps; the information and policy gaps vary almost independently
across the cells (Spearman \GapCorr{}). \emph{Information carries the context gain.} The context gain rises with
the information gap from \CtxBinLo{} points in the lowest sixth of the cells to \CtxBinHi{} in the highest
(Spearman \CtxInfo{}); the producer's ability tracks it at \CtxCap{} on its own and at \CtxCapPartial{} once the
information gap is held fixed, and with the consumer's headroom on the family held fixed the information gap keeps
\CtxInfoHead{}. A note pays when it says something the consumer does not know, whoever wrote it.
\emph{Policy disagreement takes the weights gain away.} The weights gain falls with the policy gap from
\WBinLo{} points in the lowest sixth to \WBinHi{} in the highest (\PolSlope{}; \PolPartial{} given the
information gap) and rises with the information gap (\PolSlopeInfo{}): fine-tuning buys the same information as
the notes and pays for it on every state where the producer's action is one the consumer would not take.

\subsection{Producer Shifts}
\label{sec:producers}

\paragraph{Along the ladder.}
A larger producer widens the information gap, a foreign one opens a policy gap, and each route follows its
own. For the \llmname{Qwen3-VL-8B} consumer, the 4B, 8B and 32B producers of its family have information gaps of
\InfoSmall{}, \InfoSelf{} and \InfoLarge{} and the same policy gap (\PolSame{}); their notes gain more in that
order, and in the weights only the consumer's own samples gain (Figure~\ref{fig:producer}a), the practice term of
Equation~\ref{eq:gaps}. \emph{A foreign producer is a policy gap, not an information gap.} The
\llmname{GUI-Owl-8B} pool has the information gap of a same-family producer (\InfoForeign{}) and a policy gap of
\PolForeign{}: its notes help as much as the consumer's own, and its samples hurt (Figure~\ref{fig:producer}b).
Retraining that producer on the consumer's successes closes the policy gap (\PolRetrained{}), leaves the
information gap (\InfoRetrained{}), and brings its samples to what a same-family producer's give.

\subsection{Consumer Shifts}
\label{sec:consumers}

\paragraph{Consumer upgrade.}
After an upgrade the notes keep their value and the weights start over, because what the weights bought was
practice. When the consumer moves from the 8B to the 32B model of the same family, the notes of the 8B producer still give
the 32B consumer \UpC{} points and the same samples in its weights \UpW{}. \emph{Practice needs no producer.} On
all five consumers, fine-tuning on the consumer's own successes alone, from families it already solves, matches
its full procedures within \PracticeDiff{} points (Figure~\ref{fig:gaps}d). A note library therefore carries
over a change of consumer, and the weights route is rebuilt from the new consumer's own trajectories.

\section{Conclusion}
\label{sec:conclusion}
We presented component routing, which sends each component of a GUI agent's own experience to the context or to
the weights by two properties measured on the pool before any training, recurrence and state-conditionality. On
the same items and held-out tasks across three backbone families and two environments, one pool has two
destinations: locators and lessons win in the weights, procedures and state facts in the context. A rule in the
two properties recovers the destination of every component on a held-out family and of both halves of a component
split at its boundary, and routing by it beats every whole-trajectory baseline and, on average, the better single
destination of each backbone. Note readout falls once the same component is in the weights, most for the
components that recur most; context gains grow with the information gap and weights gains shrink with the policy
gap, so where experience belongs also depends on who produced it and who consumes it.

\bibliographystyle{style/tmlr}
\bibliography{refs}

\begin{thebibliography}{68}
\providecommand{\natexlab}[1]{#1}
\providecommand{\url}[1]{\texttt{#1}}
\expandafter\ifx\csname urlstyle\endcsname\relax
  \providecommand{\doi}[1]{doi: #1}\else
  \providecommand{\doi}{doi: \begingroup \urlstyle{rm}\Url}\fi

\bibitem[Advani(2026)]{Advani2026ARXIV}
Laksh Advani.
\newblock {From Confident Closing to Silent Failure: Characterizing False
  Success in LLM Agents}.
\newblock arXiv preprint arXiv:2606.09863, 2026.

\bibitem[Agrawal et~al.(2026)Agrawal, Tan, Soylu, Ziems, Khare, Opsahl-Ong,
  Singhvi, Shandilya, Ryan, Jiang, Potts, Sen, Dimakis, Stoica, Klein, Zaharia,
  and Khattab]{Agrawal2026ICLR}
Lakshya~A Agrawal, Shangyin Tan, Dilara Soylu, Noah Ziems, Rishi Khare, Krista
  Opsahl-Ong, Arnav Singhvi, Herumb Shandilya, Michael~J Ryan, Meng Jiang,
  Christopher Potts, Koushik Sen, Alexandros~G. Dimakis, Ion Stoica, Dan Klein,
  Matei Zaharia, and Omar Khattab.
\newblock {GEPA: Reflective Prompt Evolution Can Outperform Reinforcement
  Learning}.
\newblock In \emph{International Conference on Learning Representations}, 2026.

\bibitem[Alakuijala et~al.(2026)Alakuijala, Gao, Ananov, Kaski, Marttinen,
  Ilin, and Valpola]{Alakuijala2026TMLR}
Minttu Alakuijala, Ya~Gao, Georgy Ananov, Samuel Kaski, Pekka Marttinen,
  Alexander Ilin, and Harri Valpola.
\newblock {Memento No More: Coaching AI Agents to Master Multiple Tasks via
  Hints Internalization}.
\newblock \emph{Transactions on Machine Learning Research}, 2026.

\bibitem[Bai et~al.(2025)Bai, Cai, Chen, Chen, Chen, Cheng, Deng, Ding, Gao,
  Ge, Ge, Guo, Huang, Huang, Huang, Hui, Jiang, Li, Li, Li, Li, Lin, Lin, Liu,
  Liu, Liu, Liu, Liu, Liu, Lu, Luo, Lv, Men, Meng, Ren, Ren, Song, Sun, Tang,
  Tu, Wan, Wang, Wang, Wang, Wang, Xie, Xu, Xu, Xu, Yang, Yang, Yang, Yang, Yu,
  Zhang, Zhang, Zhang, Zheng, Zhong, Zhou, Zhou, Zhou, Zhu, and
  Zhu]{Bai2025ARXIV}
Shuai Bai, Yuxuan Cai, Ruizhe Chen, Keqin Chen, Xionghui Chen, Zesen Cheng,
  Lianghao Deng, Wei Ding, Chang Gao, Chunjiang Ge, Wenbin Ge, Zhifang Guo,
  Qidong Huang, Jie Huang, Fei Huang, Binyuan Hui, Shutong Jiang, Zhaohai Li,
  Mingsheng Li, Mei Li, Kaixin Li, Zicheng Lin, Junyang Lin, Xuejing Liu,
  Jiawei Liu, Chenglong Liu, Yang Liu, Dayiheng Liu, Shixuan Liu, Dunjie Lu,
  Ruilin Luo, Chenxu Lv, Rui Men, Lingchen Meng, Xuancheng Ren, Xingzhang Ren,
  Sibo Song, Yuchong Sun, Jun Tang, Jianhong Tu, Jianqiang Wan, Peng Wang,
  Pengfei Wang, Qiuyue Wang, Yuxuan Wang, Tianbao Xie, Yiheng Xu, Haiyang Xu,
  Jin Xu, Zhibo Yang, Mingkun Yang, Jianxin Yang, An~Yang, Bowen Yu, Fei Zhang,
  Hang Zhang, Xi~Zhang, Bo~Zheng, Humen Zhong, Jingren Zhou, Fan Zhou, Jing
  Zhou, Yuanzhi Zhu, and Ke~Zhu.
\newblock {Qwen3-VL Technical Report}.
\newblock arXiv preprint arXiv:2511.21631, 2025.

\bibitem[Chan et~al.(2025)Chan, Chen, György, and Schuurmans]{Chan2025ICLR}
Bryan Chan, Xinyi Chen, András György, and Dale Schuurmans.
\newblock {Toward Understanding In-context vs. In-weight Learning}.
\newblock In \emph{International Conference on Learning Representations}, 2025.

\bibitem[Chen et~al.(2026)Chen, Yang, Fan, Nie, Sun, Zheng, Hu, Pan, Zeng, and
  Lin]{Chen2026ARXIV}
Jingwen Chen, Wenkai Yang, Shengda Fan, Wenbo Nie, Chenxing Sun, Shaodong
  Zheng, Yangen Hu, Lu~Pan, Ke~Zeng, and Yankai Lin.
\newblock {Rethinking Continual Experience Internalization for Self-Evolving
  LLM Agents}.
\newblock arXiv preprint arXiv:2606.04703, 2026.

\bibitem[Ding \& Huang(2026)Ding and Huang]{Ding2026MNET}
Zihao Ding and Jun Huang.
\newblock {Toward Trustworthy Federated Unlearning for Mobile Autonomous
  Systems}.
\newblock \emph{IEEE Network}, 2026.
\newblock \doi{10.1109/MNET.2026.3704364}.

\bibitem[Ding et~al.(2025)Ding, Huang, Duan, Zhang, Zhao, and
  Gu]{Ding2025IPCCC}
Zihao Ding, Jun Huang, Qiang Duan, Cheng Zhang, Yanxiao Zhao, and Shuyang Gu.
\newblock {A Dual-Level Game-Theoretic Approach for Collaborative Learning in
  UAV-Assisted Heterogeneous Vehicle Networks}.
\newblock In \emph{Proceedings of the IEEE International Performance,
  Computing, and Communications Conference}, pp.\  1--8, 2025.
\newblock \doi{10.1109/IPCCC66453.2025.11304656}.

\bibitem[Ding et~al.(2026{\natexlab{a}})Ding, Huang, and Qi]{Ding2026ICNC}
Zihao Ding, Jun Huang, and Junjian Qi.
\newblock {Learning to Defend: A Multi-Agent Reinforcement Learning Framework
  for Stackelberg Security Game in Mobile Edge Computing}.
\newblock In \emph{Proceedings of the International Conference on Computing,
  Networking and Communications}, pp.\  769--774, 2026{\natexlab{a}}.
\newblock \doi{10.1109/ICNC68183.2026.11416846}.

\bibitem[Ding et~al.(2026{\natexlab{b}})Ding, Wu, and Huang]{Ding2026ARXIVEASE}
Zihao Ding, Beining Wu, and Jun Huang.
\newblock {EASE: Federated Multimodal Unlearning via Entanglement-Aware Anchor
  Closure}.
\newblock arXiv preprint arXiv:2605.00733, 2026{\natexlab{b}}.

\bibitem[Ding et~al.(2026{\natexlab{c}})Ding, Wu, and Huang]{Ding2026ICDCS}
Zihao Ding, Beining Wu, and Jun Huang.
\newblock {SCALE: Sensitivity-Aware Federated Unlearning with Information
  Freshness Optimization for Mobile Edge Computing}.
\newblock In \emph{Proceedings of the IEEE International Conference on
  Distributed Computing Systems}, pp.\  1476--1486, 2026{\natexlab{c}}.
\newblock \doi{10.1109/2575-8411.2026.00143}.

\bibitem[Ding et~al.(2026{\natexlab{d}})Ding, Wu, Huang, and Mao]{Ding2026JSAC}
Zihao Ding, Beining Wu, Jun Huang, and Shiwen Mao.
\newblock {Application-Aware Twin-in-the-Loop Planning for Federated Split
  Learning over Wireless Edge Networks}.
\newblock \emph{IEEE Journal on Selected Areas in Communications},
  2026{\natexlab{d}}.
\newblock in press.

\bibitem[Duan et~al.(2023)Duan, Huang, Hu, Deng, Lu, and Yu]{Duan2023COMST}
Qiang Duan, Jun Huang, Shijing Hu, Ruijun Deng, Zhihui Lu, and Shui Yu.
\newblock {Combining Federated Learning and Edge Computing Toward Ubiquitous
  Intelligence in 6G Network: Challenges, Recent Advances, and Future
  Directions}.
\newblock \emph{IEEE Communications Surveys \& Tutorials}, 25\penalty0
  (4):\penalty0 2892--2950, 2023.
\newblock \doi{10.1109/COMST.2023.3316615}.

\bibitem[Fan et~al.(2026)Fan, Yu, Shen, Yan, Peng, Peng, Zhang, Zhang, Tan, Ge,
  Zhang, and Jiang]{Fan2026ARXIVa}
Zhimin Fan, Hongwei Yu, Yeqing Shen, Haolong Yan, Guozhen Peng, Tianhao Peng,
  Yudong Zhang, Xiaowen Zhang, Kaijun Tan, Zheng Ge, Xiangyu Zhang, and Daxin
  Jiang.
\newblock {Skill-Guided Continuation Distillation for GUI Agents}.
\newblock arXiv preprint arXiv:2606.18890, 2026.

\bibitem[Fang et~al.(2025{\natexlab{a}})Fang, Guo, Zhang, An, Ding, and
  Fang]{Fang2025ARXIV}
Zhengru Fang, Yu~Guo, Yuang Zhang, Haonan An, Wenbo Ding, and Yuguang Fang.
\newblock {Shared Spatial Memory Through Predictive Coding}.
\newblock arXiv preprint arXiv:2511.04235, 2025{\natexlab{a}}.

\bibitem[Fang et~al.(2025{\natexlab{b}})Fang, Hu, Wang, Deng, Chen, and
  Fang]{Fang2025TON}
Zhengru Fang, Senkang Hu, Jingjing Wang, Yiqin Deng, Xianhao Chen, and Yuguang
  Fang.
\newblock {Prioritized Information Bottleneck Theoretic Framework With
  Distributed Online Learning for Edge Video Analytics}.
\newblock \emph{IEEE/ACM Transactions on Networking}, 33\penalty0 (3):\penalty0
  1203--1219, 2025{\natexlab{b}}.
\newblock \doi{10.1109/TON.2025.3526148}.

\bibitem[Fang et~al.(2025{\natexlab{c}})Fang, Wang, Ma, Tao, Deng, Chen, and
  Fang]{Fang2025JSAC}
Zhengru Fang, Jingjing Wang, Yanan Ma, Yihang Tao, Yiqin Deng, Xianhao Chen,
  and Yuguang Fang.
\newblock {R-ACP: Real-Time Adaptive Collaborative Perception Leveraging Robust
  Task-Oriented Communications}.
\newblock \emph{IEEE Journal on Selected Areas in Communications}, 43\penalty0
  (12):\penalty0 4215--4230, 2025{\natexlab{c}}.
\newblock \doi{10.1109/JSAC.2025.3623179}.

\bibitem[Gou et~al.(2026)Gou, Tu, Fang, Cai, and Rezatofighi]{Gou2026ARXIV}
Chenhui Gou, Haoqin Tu, Yunhao Fang, Jianfei Cai, and Hamid Rezatofighi.
\newblock {Sample-Efficient Learning from Agent Experience}.
\newblock arXiv preprint arXiv:2607.21051, 2026.

\bibitem[Gu et~al.(2026)Gu, Li, Yu, Sun, Yang, Wang, Jia, Zhang, and
  Ye]{Gu2026ICML}
Weizheng Gu, Chengze Li, Zhuohao Yu, Mengyuan Sun, Zhibang Yang, Wei Wang,
  Hongrui Jia, Shikun Zhang, and Wei Ye.
\newblock {What Do Agents Learn from Trajectory-SFT: Semantics or Interfaces?}
\newblock In \emph{Proceedings of the International Conference on Machine
  Learning}, 2026.

\bibitem[Hajimiri et~al.(2026)Hajimiri, Aminbeidokhti, Dolz, Ayed, Laradji,
  Gella, and Gontier]{Hajimiri2026ARXIV}
Sina Hajimiri, Masih Aminbeidokhti, Jose Dolz, Ismail~Ben Ayed, Issam~H.
  Laradji, Spandana Gella, and Nicolas Gontier.
\newblock {Are Online Skill and Memory Modules Always Worth Their Tokens? A
  Budget-Constrained Study of Web Agents}.
\newblock arXiv preprint arXiv:2606.15017, 2026.

\bibitem[He et~al.(2026)He, Zhang, Ding, and Liu]{He2026ARXIV}
Bowei He, Xiaokun Zhang, Meng Ding, and Xue Liu.
\newblock {SCAFFOLD: Self-Improving Web Agents via Recursive Parametric Skill
  Abstraction}.
\newblock arXiv preprint arXiv:2609.05511, 2026.

\bibitem[Hu et~al.(2022)Hu, Shen, Wallis, Allen-Zhu, Li, Wang, Wang, and
  Chen]{Hu2022ICLR}
Edward~J. Hu, Yelong Shen, Phillip Wallis, Zeyuan Allen-Zhu, Yuanzhi Li, Shean
  Wang, Lu~Wang, and Weizhu Chen.
\newblock {LoRA: Low-Rank Adaptation of Large Language Models}.
\newblock In \emph{International Conference on Learning Representations}, 2022.

\bibitem[Hu et~al.(2026)Hu, Liang, and Zhu]{Hu2026ARXIV}
Longtao Hu, Xiao Liang, and Linchao Zhu.
\newblock {From Interaction Traces to Persistent Skills: Online Evolution for
  Computer-Use Agents}.
\newblock arXiv preprint arXiv:2609.04869, 2026.

\bibitem[Huang et~al.(2025{\natexlab{a}})Huang, Wu, Ding, and
  Ostigaard]{Huang2025RACS}
Jun Huang, Beining Wu, Zihao Ding, and Leo Ostigaard.
\newblock {Reinforcement Learning-Based Energy-Aware Coverage Path Planning for
  Precision Agriculture}.
\newblock In \emph{Proceedings of the International Conference on Research in
  Adaptive and Convergent Systems}, pp.\  1--8, 2025{\natexlab{a}}.
\newblock \doi{10.1145/3769002.3769971}.

\bibitem[Huang et~al.(2025{\natexlab{b}})Huang, Wu, Duan, Dong, and
  Yu]{Huang2025TMC}
Jun Huang, Beining Wu, Qiang Duan, Liang Dong, and Shui Yu.
\newblock {A Fast UAV Trajectory Planning Framework in RIS-Assisted
  Communication Systems With Accelerated Learning via Multithreading and
  Federating}.
\newblock \emph{IEEE Transactions on Mobile Computing}, 24\penalty0
  (8):\penalty0 6870--6885, 2025{\natexlab{b}}.
\newblock \doi{10.1109/TMC.2025.3544903}.

\bibitem[Ji et~al.(2026)Ji, Huang, Liu, Liu, Su, and Pei]{Ji2026ARXIV}
Tianyun Ji, Zhenya Huang, Jiayu Liu, Zirui Liu, Yu~Su, and Hongbin Pei.
\newblock {Learning What to Remember and What to Internalize in LLM
  Self-Evolution via Adaptive Memory-Parameter Coordination}.
\newblock arXiv preprint arXiv:2608.01234, 2026.

\bibitem[Kim et~al.(2026{\natexlab{a}})Kim, Lee, Lee, Finn, and
  Lee]{Kim2026ARXIVc}
Haechan Kim, Yoonho Lee, Gisang Lee, Chelsea Finn, and Kangwook Lee.
\newblock {WHALE: A Simple Recipe for Joint Harness-Weight Optimization}.
\newblock arXiv preprint arXiv:2609.00196, 2026{\natexlab{a}}.

\bibitem[Kim et~al.(2026{\natexlab{b}})Kim, Kim, and Hwang]{Kim2026ARXIVa}
Suji Kim, Kangsan Kim, and Sung~Ju Hwang.
\newblock {Learn from Weaknesses: Automated Domain Specialization for Small
  Computer-Use Agents}.
\newblock arXiv preprint arXiv:2605.28775, 2026{\natexlab{b}}.

\bibitem[Kim et~al.(2026{\natexlab{c}})Kim, He, Jain, Agrawal, Arabzadeh, and
  Diaz]{Kim2026ARXIVb}
To~Eun Kim, Xuhong He, Dishank Jain, Ambuj Agrawal, Negar Arabzadeh, and
  Fernando Diaz.
\newblock {Multi-Agent Transactive Memory}.
\newblock arXiv preprint arXiv:2606.19911, 2026{\natexlab{c}}.

\bibitem[Lampinen et~al.(2025)Lampinen, Chaudhry, Chan, Wild, Wan, Ku,
  Bornschein, Pascanu, Shanahan, and McClelland]{Lampinen2025ARXIV}
Andrew~K. Lampinen, Arslan Chaudhry, Stephanie C.~Y. Chan, Cody Wild, Diane
  Wan, Alex Ku, Jörg Bornschein, Razvan Pascanu, Murray Shanahan, and James~L.
  McClelland.
\newblock {On the generalization of language models from in-context learning
  and finetuning: a controlled study}.
\newblock arXiv preprint arXiv:2505.00661, 2025.

\bibitem[Li et~al.(2026)Li, Lin, Deng, Zhang, He, Ji, Cao, and
  Hooi]{Li2026ICML}
Yibo Li, Zijie Lin, Ailin Deng, Xuan Zhang, Yufei He, Shuo Ji, Tri Cao, and
  Bryan Hooi.
\newblock {Just-In-Time Reinforcement Learning: Continual Learning in LLM
  Agents Without Gradient Updates}.
\newblock In \emph{Proceedings of the International Conference on Machine
  Learning}, 2026.

\bibitem[Lin et~al.(2026)Lin, Wu, Wang, Shi, Sang, He, Liu, Wei, Wu, Zhang,
  Wang, Zhang, Dumoulin, Xie, Zhou, Wang, and Lu]{Lin2026ARXIV}
Minhua Lin, Juncheng Wu, Zijun Wang, Zhan Shi, Yisi Sang, Bing He, Zewen Liu,
  Tianxin Wei, Zongyu Wu, Zhiwei Zhang, Dakuo Wang, Xiang Zhang, Benoit
  Dumoulin, Cihang Xie, Yuyin Zhou, Suhang Wang, and Hanqing Lu.
\newblock {Harness Updating Is Not Harness Benefit: Disentangling Evolution
  Capabilities in Self-Evolving LLM Agents}.
\newblock arXiv preprint arXiv:2605.30621, 2026.

\bibitem[Liu et~al.(2026{\natexlab{a}})Liu, Xie, Ding, Li, Yang, Wu, Wang, Sun,
  Liu, Wang, Ye, Li, Dong, Yu, Lu, Mo, Yan, Tian, Zhang, Huang, Liu, Su, Luo,
  Yue, Qi, Chen, Zhou, Qiao, Chen, and Wang]{Liu2026ICLR}
Zhaoyang Liu, Jingjing Xie, Zichen Ding, Zehao Li, Bowen Yang, Zhenyu Wu,
  Xuehui Wang, Qiushi Sun, Shi Liu, Weiyun Wang, Shenglong Ye, Qingyun Li, Xuan
  Dong, Yue Yu, Chenyu Lu, YunXiang Mo, Yao Yan, Zeyue Tian, Xiao Zhang, Yuan
  Huang, Yiqian Liu, Weijie Su, Gen Luo, Xiangyu Yue, Biqing Qi, Kai Chen,
  Bowen Zhou, Yu~Qiao, Qifeng Chen, and Wenhai Wang.
\newblock {ScaleCUA: Scaling Open-Source Computer Use Agents with
  Cross-Platform Data}.
\newblock In \emph{International Conference on Learning Representations},
  2026{\natexlab{a}}.

\bibitem[Liu et~al.(2026{\natexlab{b}})Liu, Huang, Su, Pan, Han, Chen, Song,
  and Zhang]{Liu2026ARXIV}
Zhou Liu, Ligang Huang, Zeli Su, Zewei Pan, Zhaoyang Han, Xing Chen, Yuanfeng
  Song, and Wentao Zhang.
\newblock {SkillLens: Visual Skill Cards for Retrieval-Augmented GUI Action
  Prediction and On-Policy Distillation}.
\newblock arXiv preprint arXiv:2608.10775, 2026{\natexlab{b}}.

\bibitem[Lou et~al.(2026)Lou, Peng, Yao, Wu, Cheng, Nath, Yin, and
  Gao]{Lou2026COLM}
Renze Lou, Baolin Peng, Wenlin Yao, Qianhui Wu, Hao Cheng, Suman Nath, Wenpeng
  Yin, and Jianfeng Gao.
\newblock {The Tool Illusion: Rethinking Tool Use in Web Agents}.
\newblock In \emph{Conference on Language Modeling}, 2026.

\bibitem[Men et~al.(2026)Men, Jin, Cao, Chen, Liu, and Zhao]{Men2026ACL}
Tianyi Men, Zhuoran Jin, Pengfei Cao, Yubo Chen, Kang Liu, and Jun Zhao.
\newblock {Empowering GUI Agents via Autonomous Experience Exploration and
  Hindsight Experience Utilization for Task Planning}.
\newblock In \emph{Proceedings of the Annual Meeting of the Association for
  Computational Linguistics}, 2026.

\bibitem[Pan et~al.(2023)Pan, Wu, Sun, and Xu]{Pan2023SCIS}
Dong Pan, Bei-Ning Wu, Yi-Liu Sun, and Yi-Peng Xu.
\newblock {A Fault-Tolerant and Energy-Efficient Design of a Network Switch
  Based on a Quantum-Based Nano-Communication Technique}.
\newblock \emph{Sustainable Computing: Informatics and Systems}, 37:\penalty0
  100827, 2023.
\newblock \doi{10.1016/j.suscom.2022.100827}.

\bibitem[Rafailov et~al.(2023)Rafailov, Sharma, Mitchell, Ermon, Manning, and
  Finn]{Rafailov2023NEURIPS}
Rafael Rafailov, Archit Sharma, Eric Mitchell, Stefano Ermon, Christopher~D.
  Manning, and Chelsea Finn.
\newblock {Direct Preference Optimization: Your Language Model is Secretly a
  Reward Model}.
\newblock In \emph{Advances in Neural Information Processing Systems}, 2023.

\bibitem[Rawles et~al.(2025)Rawles, Clinckemaillie, Chang, Waltz, Lau, Fair,
  Li, Bishop, Li, Campbell-Ajala, Toyama, Berry, Tyamagundlu, Lillicrap, and
  Riva]{Rawles2025ICLR}
Christopher Rawles, Sarah Clinckemaillie, Yifan Chang, Jonathan Waltz,
  Gabrielle Lau, Marybeth Fair, Alice Li, William Bishop, Wei Li, Folawiyo
  Campbell-Ajala, Daniel Toyama, Robert Berry, Divya Tyamagundlu, Timothy
  Lillicrap, and Oriana Riva.
\newblock {AndroidWorld: A Dynamic Benchmarking Environment for Autonomous
  Agents}.
\newblock In \emph{International Conference on Learning Representations}, 2025.

\bibitem[Sui et~al.(2026)Sui, Zhao, Tang, Zhao, Wu, Tu, and Qin]{Sui2026ARXIV}
Xingyu Sui, Weixiang Zhao, Yongxin Tang, Yanyan Zhao, Yang Wu, Dandan Tu, and
  Bing Qin.
\newblock {Learning to Learn from Multimodal Experience}.
\newblock arXiv preprint arXiv:2605.16857, 2026.

\bibitem[Sun et~al.(2026)Sun, Zhang, Wang, and Wei]{Sun2026ACL}
Libo Sun, Jiwen Zhang, Siyuan Wang, and Zhongyu Wei.
\newblock {MAGNET: Towards Adaptive GUI Agents with Memory-Driven Knowledge
  Evolution}.
\newblock In \emph{Proceedings of the Annual Meeting of the Association for
  Computational Linguistics}, 2026.

\bibitem[Tchalla et~al.(2026)Tchalla, Wu, Huang, Gu, and
  Duan]{Tchalla2026ARXIV}
Dhe~Yeong Tchalla, Beining Wu, Jun Huang, Shuyang Gu, and Qiang Duan.
\newblock {FedSceneX: Time-to-Target Orchestration for Same-Scene Multimodal
  Federated Edge Learning}.
\newblock arXiv preprint arXiv:2608.07730, 2026.

\bibitem[Ullrich et~al.(2026)Ullrich, Su, Shi, Subramonian, Bar, Evtimov,
  Tsilivis, Balestriero, Kempe, and Ibrahim]{Ullrich2026ICLR}
Karen Ullrich, Jingtong Su, Claudia Shi, Arjun Subramonian, Amir Bar, Ivan
  Evtimov, Nikolaos Tsilivis, Randall Balestriero, Julia Kempe, and Mark
  Ibrahim.
\newblock {OpenApps: Simulating Environment Variations to Measure UI-Agent
  Reliability}.
\newblock In \emph{International Conference on Learning Representations}, 2026.

\bibitem[Wang et~al.(2026{\natexlab{a}})Wang, Lu, Wang, Bai, Liu, Zhang, Wang,
  Hu, Xie, Bai, Liu, Shen, Lin, and Yu]{Wang2026ARXIVb}
Bowen Wang, Dunjie Lu, Junli Wang, Tianyi Bai, Shixuan Liu, Zhipeng Zhang,
  Haiquan Wang, Hao Hu, Tianbao Xie, Shuai Bai, Dayiheng Liu, Que Shen, Junyang
  Lin, and Tao Yu.
\newblock {CUA-Gym: Scaling Verifiable Training Environments and Tasks for
  Computer-Use Agents}.
\newblock arXiv preprint arXiv:2605.25624, 2026{\natexlab{a}}.

\bibitem[Wang et~al.(2025)Wang, Wang, Lu, Yang, Xie, Wang, Deng, Guo, Xu, Wu,
  Shen, Li, Li, Li, Chen, Zheng, Li, Lei, Cao, Fu, Shin, Shin, Hu, Wang, Chen,
  Ye, Zhang, Du, Hu, Chen, Zhou, Yao, Chen, Gu, Wang, Wang, Yang, Zhong, Sung,
  Charles, Yang, and Yu]{Wang2025NEURIPS}
Xinyuan Wang, Bowen Wang, Dunjie Lu, Junlin Yang, Tianbao Xie, Junli Wang,
  Jiaqi Deng, Xiaole Guo, Yiheng Xu, Chen~Henry Wu, Zhennan Shen, Zhuokai Li,
  Ryan Li, Xiaochuan Li, Junda Chen, Boyuan Zheng, Peihang Li, Fangyu Lei,
  Ruisheng Cao, Yeqiao Fu, Dongchan Shin, Martin Shin, Jiarui Hu, Yuyan Wang,
  Jixuan Chen, Yuxiao Ye, Danyang Zhang, Dikang Du, Hao Hu, Huarong Chen, Zaida
  Zhou, Haotian Yao, Ziwei Chen, Qizheng Gu, Yipu Wang, Heng Wang, Diyi Yang,
  Victor Zhong, Flood Sung, Y.~Charles, Zhilin Yang, and Tao Yu.
\newblock {OpenCUA: Open Foundations for Computer-Use Agents}.
\newblock In \emph{Advances in Neural Information Processing Systems}, 2025.

\bibitem[Wang et~al.(2026{\natexlab{b}})Wang, Gupta, Dong, and
  MacLellan]{Wang2026ARXIVa}
Zekun Wang, Anant Gupta, Zihan Dong, and Christopher~J. MacLellan.
\newblock {Self-Consolidating Language Models: Continual Knowledge
  Incorporation from Context}.
\newblock arXiv preprint arXiv:2605.07076, 2026{\natexlab{b}}.

\bibitem[Wei et~al.(2026)Wei, Shi, Lin, He, Liu, Sang, Bei, Ning, Zou, Li, Lin,
  Zhao, Wang, Dumoulin, Wang, He, and Lu]{Wei2026ARXIV}
Tianxin Wei, Zhan Shi, Minhua Lin, Bing He, Zewen Liu, Yisi Sang, Yuanchen Bei,
  Xuying Ning, Jiaru Zou, Ting-Wei Li, Xiao Lin, Yanjun Zhao, Chi Wang, Benoit
  Dumoulin, Dakuo Wang, Jingrui He, and Hanqing Lu.
\newblock {Evo-Harness: Context-to-Harness Skill Compilation for Self-Evolving
  Agents}.
\newblock arXiv preprint arXiv:2608.15071, 2026.

\bibitem[Wu \& Huang(2026{\natexlab{a}})Wu and Huang]{Wu2026ARXIVCrystalMem}
Beining Wu and Jun Huang.
\newblock {CrystalMem: Elastic Memory for Self-Evolving LLM Agents via
  Knowledge Crystallization}.
\newblock arXiv preprint arXiv:2608.00303, 2026{\natexlab{a}}.

\bibitem[Wu \& Huang(2026{\natexlab{b}})Wu and Huang]{Wu2026ARXIVLifecycle}
Beining Wu and Jun Huang.
\newblock {Lifecycle-Aware Federated Continual Learning in Mobile Autonomous
  Systems}.
\newblock arXiv preprint arXiv:2604.20745, 2026{\natexlab{b}}.

\bibitem[Wu \& Wu(2023)Wu and Wu]{Wu2023MPE}
Beining Wu and Wei Wu.
\newblock {Model-Free Cooperative Optimal Output Regulation for Linear
  Discrete-Time Multi-Agent Systems Using Reinforcement Learning}.
\newblock \emph{Mathematical Problems in Engineering}, 2023\penalty0
  (1):\penalty0 6350647, 2023.
\newblock \doi{10.1155/2023/6350647}.

\bibitem[Wu et~al.(2023)Wu, Cai, Wu, and Yin]{Wu2023ACCESS}
Beining Wu, Zhengkun Cai, Wei Wu, and Xiaobin Yin.
\newblock {AoI-Aware Resource Management for Smart Health via Deep
  Reinforcement Learning}.
\newblock \emph{IEEE Access}, 11:\penalty0 81180--81195, 2023.
\newblock \doi{10.1109/ACCESS.2023.3299340}.

\bibitem[Wu et~al.(2025)Wu, Huang, and Duan]{Wu2025WASA}
Beining Wu, Jun Huang, and Qiang Duan.
\newblock {FedTD3: An Accelerated Learning Approach for UAV Trajectory
  Planning}.
\newblock In \emph{Proceedings of the International Conference on Wireless
  Artificial Intelligent Computing Systems and Applications}, pp.\  13--24,
  2025.
\newblock \doi{10.1007/978-981-96-8725-1_2}.

\bibitem[Wu et~al.(2026{\natexlab{a}})Wu, Ding, and Huang]{Wu2026ARXIVERRAND}
Beining Wu, Zihao Ding, and Jun Huang.
\newblock {ERRAND: Budgeted Maintenance of Agent Memory}.
\newblock arXiv preprint arXiv:2609.29545, 2026{\natexlab{a}}.

\bibitem[Wu et~al.(2026{\natexlab{b}})Wu, Ding, and Huang]{Wu2026IoTJ}
Beining Wu, Zihao Ding, and Jun Huang.
\newblock {RELIEF: Turning Missing Modalities into Training Acceleration for
  Federated Learning on Heterogeneous IoT Edge}.
\newblock \emph{IEEE Internet of Things Journal}, 2026{\natexlab{b}}.
\newblock \doi{10.1109/JIOT.2026.3725593}.

\bibitem[Wu et~al.(2026{\natexlab{c}})Wu, Ding, and Huang]{Wu2026TMM}
Beining Wu, Zihao Ding, and Jun Huang.
\newblock {Exposing and Resolving Spurious Isolation in Federated Multimodal
  Continual Learning}.
\newblock \emph{IEEE Transactions on Multimedia}, 2026{\natexlab{c}}.
\newblock \doi{10.1109/TMM.2026.3736516}.

\bibitem[Wu et~al.(2026{\natexlab{d}})Wu, Ding, and Huang]{Wu2026TNSE}
Beining Wu, Zihao Ding, and Jun Huang.
\newblock {A Review of Continual Learning in Edge AI}.
\newblock \emph{IEEE Transactions on Network Science and Engineering},
  13:\penalty0 6571--6588, 2026{\natexlab{d}}.
\newblock \doi{10.1109/TNSE.2026.3657652}.

\bibitem[Wu et~al.(2026{\natexlab{e}})Wu, Ding, Huang, and
  Zhao]{Wu2026ARXIVForget}
Beining Wu, Zihao Ding, Jun Huang, and Yanxiao Zhao.
\newblock {Forget to Improve: On-Device LLM-Agent Continual Learning via
  Budget-Curated Memory}.
\newblock arXiv preprint arXiv:2606.25115, 2026{\natexlab{e}}.

\bibitem[Wu et~al.(2026{\natexlab{f}})Wu, Huang, and Duan]{Wu2026MNET}
Beining Wu, Jun Huang, and Qiang Duan.
\newblock {Real-Time Intelligent Healthcare Enabled by Federated Digital Twins
  With AoI Optimization}.
\newblock \emph{IEEE Network}, 40\penalty0 (2):\penalty0 184--191,
  2026{\natexlab{f}}.
\newblock \doi{10.1109/MNET.2025.3565977}.

\bibitem[Wu et~al.(2026{\natexlab{g}})Wu, Huang, Duan, Dong, and
  Cai]{Wu2026TON}
Beining Wu, Jun Huang, Qiang Duan, Liang Dong, and Zhipeng Cai.
\newblock {Enhancing Vehicular Platooning With Wireless Federated Learning: A
  Resource-Aware Control Framework}.
\newblock \emph{IEEE/ACM Transactions on Networking}, 34:\penalty0 1479--1494,
  2026{\natexlab{g}}.
\newblock \doi{10.1109/TON.2025.3625084}.

\bibitem[Wu et~al.(2026{\natexlab{h}})Wu, Huang, and Yu]{Wu2026COMST}
Beining Wu, Jun Huang, and Shui Yu.
\newblock {``X of Information'' Continuum: A Survey on AI-Driven
  Multi-Dimensional Metrics for Next-Generation Networked Systems}.
\newblock \emph{IEEE Communications Surveys \& Tutorials}, 28:\penalty0
  5307--5344, 2026{\natexlab{h}}.
\newblock \doi{10.1109/COMST.2026.3670279}.

\bibitem[Wu et~al.(2026{\natexlab{i}})Wu, Huang, and Zhao]{Wu2026ICDCS}
Beining Wu, Jun Huang, and Yanxiao Zhao.
\newblock {From Alpha to Omega: Lifecycle-Aware Forgetting Defense in Federated
  Continual Learning for Planetary Exploration}.
\newblock In \emph{Proceedings of the IEEE International Conference on
  Distributed Computing Systems}, pp.\  1--11, 2026{\natexlab{i}}.
\newblock \doi{10.1109/2575-8411.2026.00061}.

\bibitem[Wu et~al.(2026{\natexlab{j}})Wu, Hao, Wang, Wu, Xiao, Li, Zhou, Ju,
  Liu, Fan, and Zhang]{Wu2026ARXIVa}
Dingbang Wu, Rui Hao, Haiyang Wang, Shuzhe Wu, Han Xiao, Zhenghong Li, Bojiang
  Zhou, Zheng Ju, Zichen Liu, Lue Fan, and Zhaoxiang Zhang.
\newblock {MobileGym: A Verifiable and Highly Parallel Simulation Platform for
  Mobile GUI Agent Research}.
\newblock arXiv preprint arXiv:2605.26114, 2026{\natexlab{j}}.

\bibitem[Wu et~al.(2026{\natexlab{k}})Wu, Yang, Lu, Zhang, Shen, Feng, Luo,
  Lian, Zhang, Wen, and Tao]{Wu2026ARXIVb}
Jinyang Wu, Shuo Yang, Zhengxi Lu, Fan Zhang, Yuhao Shen, Lang Feng, Haoran
  Luo, Zheng Lian, Shuai Zhang, Zhengqi Wen, and Jianhua Tao.
\newblock {SEED: Self-Evolving On-Policy Distillation for Agentic Reinforcement
  Learning}.
\newblock arXiv preprint arXiv:2607.14777, 2026{\natexlab{k}}.

\bibitem[Xing et~al.(2025)Xing, Ding, and Huang]{Xing2025ACR}
Cong-Cong Xing, Zihao Ding, and Jun Huang.
\newblock {A Stochastic Geometry-Based Analysis of SWIPT-Assisted Underlaid
  Device-to-Device Energy Harvesting}.
\newblock \emph{ACM SIGAPP Applied Computing Review}, 25\penalty0 (4):\penalty0
  18--34, 2025.
\newblock \doi{10.1145/3787594.3787596}.

\bibitem[Xiong et~al.(2026)Xiong, Lin, Xie, He, Liu, Tang, Lakkaraju, and
  Xiang]{Xiong2026ACL}
Zidi Xiong, Yuping Lin, Wenya Xie, Pengfei He, Zirui Liu, Jiliang Tang,
  Himabindu Lakkaraju, and Zhen Xiang.
\newblock {How Memory Management Impacts LLM Agents: An Empirical Study of
  Experience-Following Behavior}.
\newblock In \emph{Proceedings of the Annual Meeting of the Association for
  Computational Linguistics}, 2026.

\bibitem[Xu et~al.(2026)Xu, Zhang, Liu, Wang, Zhu, Zhou, Hu, Gao, Cao, Wang,
  Chen, Liao, Zheng, Zeng, Xu, Bai, Lin, Zhou, and Yan]{Xu2026ARXIV}
Haiyang Xu, Xi~Zhang, Haowei Liu, Junyang Wang, Zhaozai Zhu, Shengjie Zhou,
  Xuhao Hu, Feiyu Gao, Junjie Cao, Zihua Wang, Zhiyuan Chen, Jitong Liao,
  Qi~Zheng, Jiahui Zeng, Ze~Xu, Shuai Bai, Junyang Lin, Jingren Zhou, and Ming
  Yan.
\newblock {Mobile-Agent-v3.5: Multi-platform Fundamental GUI Agents}.
\newblock arXiv preprint arXiv:2602.16855, 2026.

\bibitem[Ye et~al.(2026)Ye, Li, Pruksachatkun, Zhang, and Wu]{Ye2026ARXIVa}
Qinyuan Ye, Yu~Li, Yada Pruksachatkun, Jiaxin Zhang, and Chien-Sheng Wu.
\newblock {On the Fragility of Self-Improving Agents: Variance, Task Order, and
  Underspecification}.
\newblock arXiv preprint arXiv:2608.18066, 2026.

\bibitem[Zhang et~al.(2026)Zhang, Chen, Liu, Xue, Liao, Liu, Wang, Ning, Chen,
  Fu, Xie, Sun, Gou, Qi, Meng, Yang, Zhang, Li, Shah, Huynh, Li, Yang, Cao,
  Jang, Zhou, Zhu, Sun, Weston, Su, and Wu]{Zhang2026ICML}
Kai Zhang, Xiangchao Chen, Bo~Liu, Tianci Xue, Zeyi Liao, Zhihan Liu, Xiyao
  Wang, Yuting Ning, Zhaorun Chen, Xiaohan Fu, Jian Xie, Yuxuan Sun, Boyu Gou,
  Qi~Qi, Zihang Meng, Jianwei Yang, Ning Zhang, Xian Li, Ashish Shah, Dat
  Huynh, Hengduo Li, Zi~Yang, Sara Cao, Lawrence Jang, Shuyan Zhou, Jiacheng
  Zhu, Huan Sun, Jason Weston, Yu~Su, and Yifan Wu.
\newblock {Agent Learning via Early Experience}.
\newblock In \emph{Proceedings of the International Conference on Machine
  Learning}, 2026.

\end{thebibliography}

\appendix
\clearpage

\makeatletter
\setlength{\@fptop}{0pt}
\makeatother

\section*{Table of Contents}
\setcounter{tocdepth}{2}
\renewcommand{\contentsname}{Appendix Contents}
\startcontents[appendix]
\printcontents[appendix]{}{1}{}

\counterwithin{table}{section}
\counterwithin{figure}{section}
\counterwithin{equation}{section}
\renewcommand{\theHtable}{app.\thesection.\arabic{table}}
\renewcommand{\theHfigure}{app.\thesection.\arabic{figure}}
\renewcommand{\theHequation}{app.\thesection.\arabic{equation}}

\providecommand{\Comment}[1]{\hfill$\triangleright$ \textit{#1}}

\newtcolorbox{compcardbox}[1]{breakable, enhanced, colback=toolcardbox,
  colframe=toolcardborder, coltitle=white, fonttitle=\bfseries\small,
  title={Component Card: #1}, rounded corners, boxrule=1pt,
  left=5pt, right=5pt, top=4pt, bottom=4pt, fontupper=\small}

\clearpage
\section{Components and Items}
\label{app:components}

\begin{figure}[!htbp]
  \centering
  \includegraphics[width=\textwidth]{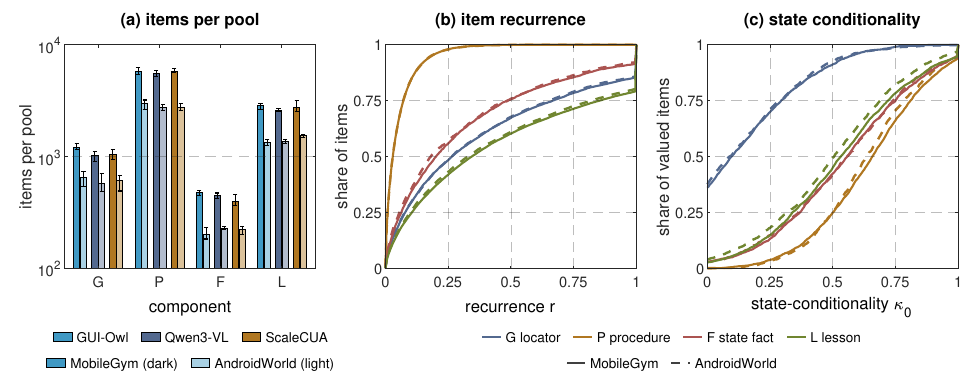}
  \caption{\textbf{Pool statistics.} (a)~Items per pool by component and backbone on MobileGym (dark) and
  AndroidWorld (light), mean over seeds with the seed range. (b)~Distribution of item recurrence and (c)~of
  state-conditionality over the items that have one, per component; MobileGym solid, AndroidWorld dashed.}
  \label{fig:app_pools}
\end{figure}

An item is one unit of experience taken from the pool: a key that identifies it across episodes, a payload
that the route carries, and a state condition under which the payload applies. The four cards below give the
three fields of each component together with the note the context route serves and the training sample the
weights route trains on, as defined in Section~\ref{sec:setup}. The example on each card is one item of a
\llmname{GUI-Owl-8B} pool as its note, in the template of Appendix~\ref{app:notesblock}, with an English
rendering. Table~\ref{tab:app_items}
counts the items per component and environment, and Figure~\ref{fig:app_pools} shows the item-level
distributions behind it.

\subsection{Component Cards}
\label{app:cards}

\begin{compcardbox}{Locators (\textit{G})}
\setlength{\parindent}{0pt}
\setlength{\parskip}{4pt}
\textbf{Description:} where a named element sits on a screen.

\textbf{Key:} (application, role, element name).

\textbf{Payload:} the position of the element on that screen.

\textbf{State condition:} the screen on which the element was found.

\textbf{As a note:} one line that names the element and its position, served when the current screen matches
the key.

\textbf{As a training sample:} the screenshot paired with a click on the element.

{\raggedright
\textbf{Example item:} \begin{CJK}{UTF8}{gbsn}在 weather 的该页面上，元素「城市管理」（div）的点击坐标是 (770, 84)。\end{CJK}
\emph{On this page of weather, the element ``City management'' (div) is tapped at (770, 84).} Key (weather, div,
City management); payload (770, 84).\par}
\end{compcardbox}

\begin{compcardbox}{Procedures (\textit{P})}
\setlength{\parindent}{0pt}
\setlength{\parskip}{4pt}
\textbf{Description:} a short run of action targets that recurs inside a task family.

\textbf{Key:} the run of three consecutive targets.

\textbf{Payload:} the subgoal sequence of the run.

\textbf{State condition:} the task family and the screen at which the run starts.

\textbf{As a note:} a hint that lists the subgoals in order, served when the task and the screen match the key.

\textbf{As a training sample:} task, screenshot and history mapped to the next action, with the subgoals as
the thought.

{\raggedright
\textbf{Example item:} \begin{CJK}{UTF8}{gbsn}在 weather 中，做完「城市管理」「深圳市」后，下一步通常是「如：一般」。\end{CJK}
\emph{In weather, after ``City management'' and ``Shenzhen'', the next step is usually ``e.g.\ Fair''.} Key
(City management, Shenzhen, e.g.\ Fair); payload: the third target.\par}
\end{compcardbox}

\begin{compcardbox}{State facts (\textit{F})}
\setlength{\parindent}{0pt}
\setlength{\parskip}{4pt}
\textbf{Description:} a precondition that must hold before an action succeeds, taken from a failed and a
successful attempt at the same target.

\textbf{Key:} (application, target, predicate).

\textbf{Payload:} the predicate.

\textbf{State condition:} the screen that shows the target.

\textbf{As a note:} one line that states the precondition, served when the target is on the current screen.

\textbf{As a training sample:} the source step mapped to the action, with the predicate as the thought.

{\raggedright
\textbf{Example item:} \begin{CJK}{UTF8}{gbsn}在 answer\_sheet 中执行「提交答案」之前，需要满足：apps.answer\_sheet.answers.0 须为任务要求的值。\end{CJK}
\emph{Before doing ``Submit answer'' in answer\_sheet, the following must hold: apps.answer\_sheet.answers.0
must be the value the task requires.} Key (answer\_sheet, Submit answer, answers.0 set to the required value);
payload: the predicate.\par}
\end{compcardbox}

\begin{compcardbox}{Lessons (\textit{L})}
\setlength{\parindent}{0pt}
\setlength{\parskip}{4pt}
\textbf{Description:} an action that failed in a state, paired with the action that later succeeded there.

\textbf{Key:} (application, state, failed target).

\textbf{Payload:} the action that succeeded.

\textbf{State condition:} the state in which the failed action was taken.

\textbf{As a note:} one line that names the failed action and the one that succeeded, served when the state
matches the key.

\textbf{As a training sample:} the preference pair at the source state, the succeeded action chosen and the
failed action rejected.

{\raggedright
\textbf{Example item:} \begin{CJK}{UTF8}{gbsn}在 answer\_sheet 的该状态下不要执行「如：一般」，应执行「提交答案」。\end{CJK}
\emph{In this state of answer\_sheet, do not do ``e.g.\ Fair''; do ``Submit answer'' instead.} Key
(answer\_sheet, the source state, e.g.\ Fair); payload: Submit answer.\par}
\end{compcardbox}

\subsection{Extraction}
\label{app:extraction}

Two rule-based extractors read the accessibility state and the action of every step in a pool and emit the
items of the four components: a strict one keeps element names verbatim, and a lenient one normalizes them
before keys are compared. Every pool is built with the lenient extractor. Items are rendered as notes and as
training samples in the fixed templates of Appendix~\ref{app:notesblock}. Table~\ref{tab:app_items} gives the
pool coordinates at the item level: the mean recurrence and the median state-conditionality that place each
component in Figure~\ref{fig:rule}(a) are means and medians over these items.

\begin{table}[!htbp]
\centering
\caption{\textbf{Items per component.} Number of items, mean recurrence $r$, median state-conditionality $\kappa_0$ over the items that have one, and the share of items that have one, summed or averaged over the nine pools of each environment (three backbones, \nSeeds{} seeds).}
\label{tab:app_items}
\footnotesize
\setlength{\tabcolsep}{4.5pt}
\renewcommand{\arraystretch}{1.15}
\begin{tabularx}{\textwidth}{l|>{\raggedleft\arraybackslash}X>{\raggedleft\arraybackslash}X>{\raggedleft\arraybackslash}X>{\raggedleft\arraybackslash}X|>{\raggedleft\arraybackslash}X>{\raggedleft\arraybackslash}X>{\raggedleft\arraybackslash}X>{\raggedleft\arraybackslash}X}
\toprule
& \multicolumn{4}{c|}{\cellcolor{SearchBg}\textbf{MobileGym}} & \multicolumn{4}{c}{\cellcolor{MathBg}\textbf{AndroidWorld}} \\ \cmidrule(lr){2-5} \cmidrule(lr){6-9}
\textbf{Component} & \textbf{Items} & $\boldsymbol{r}$ & $\boldsymbol{\kappa_0}$ & \textbf{Valued} & \textbf{Items} & $\boldsymbol{r}$ & $\boldsymbol{\kappa_0}$ & \textbf{Valued} \\
\midrule
Locators & 9907 & 0.39 & 0.10 & 0.21 & 5544 & 0.38 & 0.09 & 0.20 \\
Procedures & 51052 & 0.05 & 0.66 & 0.06 & 25519 & 0.05 & 0.63 & 0.05 \\
State facts & 3980 & 0.32 & 0.56 & 0.53 & 1972 & 0.31 & 0.56 & 0.56 \\
Lessons & 24651 & 0.45 & 0.53 & 0.15 & 12775 & 0.44 & 0.50 & 0.15 \\
\bottomrule
\end{tabularx}
\end{table}

\clearpage
\section{Arms and Protocol}
\label{app:arms}

This appendix gives the design constants behind Section~\ref{sec:setup}: the two environments and their splits,
the pools, every arm, and the interaction budget. All of them were fixed before the runs; measured outcomes are
in the main text and in Appendix~\ref{app:results}.

\subsection{Environments and Splits}
\label{app:splits}

\paragraph{MobileGym.}
MobileGym has \nApps{} applications and \nTemplates{} task templates in \nSuites{} suites, split by the benchmark
into \nTrainTemplates{} training and \nTestTemplates{} test templates. A template instantiates a task with sampled
parameters, and the environment judges an episode by comparing the final application state with the target
state. Applications are ranked by their number of training templates, and every fourth application in that
ranking is held out as unseen; this gives \nUnseenApps{} unseen applications. Training templates whose
applications are all seen form the experience pool (\nExpTemplates{} templates). Test templates whose
applications are all seen form the seen split (\nSeenTemplates{}), test templates whose applications are all
unseen form the unseen split (\nUnseenTemplates{}), and the \nMixedTemplates{} test templates that touch both are
reported separately and enter no main-text number. Each held-out template is instantiated up to \kInstances{}
times per seed (\nTemplatesEight{} templates have \kInstances{} instances and \nTemplatesOne{} have one), with
parameters fixed by (template, instance, seed) and stored, so every arm runs the same \nPairsPerSeed{} instances
per seed.

\paragraph{AndroidWorld.}
AndroidWorld has \nAWTasks{} tasks over \nAWApps{} applications, judged by the official evaluator. The unseen tier
takes \nAWUnseenApps{} of the \nAWApps{} applications (identifiers \texttt{audiorec}, \texttt{draw},
\texttt{joplin}, \texttt{notes}, \texttt{osmand}, \texttt{retro}, \texttt{simplecal} and \texttt{vlc}). The \nAWHeldoutTasks{} held-out tasks are instantiated \kAWInstances{}
times per seed, \nAWPairsPerSeed{} instances, and the remaining tasks form the pool. The state signature is the
set of visible accessibility nodes (class, resource id, normalized text), capped at \nAWNodes{} nodes, together
with the package and activity. Locator keys are (application, class, resource id or text); procedure keys are
three consecutive action-target signatures; state-fact keys are the node signatures present in the successful
state and absent in the failed one, at most \nFPred{}; lesson keys are as on MobileGym. Actions follow the
official action space through each backbone's own adapter.

\paragraph{Pools.}
On MobileGym the base agent of each family runs every experience template \nEpPerTemplate{} times at
temperature~1 with a fresh instance each time, once per seed. Every step records the screenshot, the environment
state, the element list, the action and its target. On AndroidWorld the pool runs each pool task \nAWEpPerTask{}
times. The four component extractors run on the pool once, and the recurrence and state-conditionality of every
item are computed from the pool alone. The coordinates of a cell are the mean recurrence and the median
state-conditionality over the items of its component, merged over the three seeds. For the producer ladder of
Section~\ref{sec:shifts} the pools are repeated with the 4B and 32B members of the Qwen3-VL family and the 32B
member of the GUI-Owl family as producers.

\paragraph{Producer--consumer batch.}
The measurements of Sections~\ref{sec:obs2} and~\ref{sec:shifts} come from one batch of their own. Every
procedures adapter of this batch trains on \nBudgetB{} samples drawn from the producer's pool (\nBudgetBL{} for
the 32B consumers), so every producer enters the weights at the same training budget, and its own-procedure arms
are separate runs from those of Table~\ref{tab:interventions}. The batch groups the held-out templates into
\nFamilies{} task families: \nFamSingle{} single-application suites, \nFamCross{} cross-application suites and
\nFamShared{} kinds of task shared across applications (search and filter, compose and send, settings toggles,
media control and multi-step forms).

\paragraph{Version change.}
The conflict measurement of Section~\ref{sec:rule3} uses OpenApps \citep{Ullrich2026ICLR}, whose \nOATasks{}
tasks over \nOAApps{} applications exist in two application versions. Each version is sampled \nOAEp{} times per
task. The weights route is trained on version~1 items, the notes are served from version~2, and every arm runs on
all three families and \nSeeds{} seeds, on \nOAInstances{} version-2 instances per seed (ten goal variants per
task). At the first state where the two versions' successful procedures diverge, the arm's action is compared with
the version-2 note and with the version-1 weights without notes, under the same-action rule of
Appendix~\ref{app:budget}; the shares of Section~\ref{sec:rule3} count \nOASteps{} such steps, \nOAStepsOwl{} on
\llmname{GUI-Owl-8B} and \nOAStepsOther{} on each other family.

\paragraph{Evaluation.}
All arms use greedy decoding, the official step limit, the native screen resolution, and the same stored
instances. An episode is a success when the environment judge accepts the final state. Every reported
difference between two arms is paired over (template, instance, seed) and carries a percentile interval from
\nBoot{} bootstrap resamples of the paired differences. Cells pool the \nSeeds{} seeds; per-seed values give the
subscripts of the tables.

\subsection{Arms}
\label{app:armlist}

The arm names below are the ones of the delivered result tables. Unless stated, an arm runs on all three
families, both environments and all seeds.

\textbf{Base agent and single-component routes:}
\begin{itemize}[leftmargin=1em, nosep]
\item \textbf{none}: the base agent with the evaluation prompt and no experience; the zero of every gain.
\item \textbf{C\_G, C\_P, C\_F, C\_L}: the context route of one component. At every step the items of the
same application are ranked by the Jaccard similarity of their source state signature to the current one.
Exact matches (at least \jExact{}) come first and relaxed matches (at least \jRelaxed{}) fill up to \kNotes{}
notes, appended to the prompt as a notes block. Procedures retrieve the skill file of the most similar task
text first and the step-level items within it second.
\item \textbf{C\_c\_exact, C\_c\_relaxed}: the context route restricted to the exact or to the relaxed tier;
their difference against \textbf{none} gives the online state-conditionality of Equation~\ref{eq:kappa}.
\item \textbf{W\_G, W\_P, W\_F}: the weights route of one component, a LoRA adapter fine-tuned on the
component's items rendered as samples (Appendix~\ref{app:training}). \textbf{W\_L} trains the adapter by
preference optimization, with the successful action as chosen and the failed action as rejected.
\item \textbf{CW\_G, CW\_P, CW\_F, CW\_L}: the weights-route adapter served with the notes of the same
component. Every note-bearing step also runs once without the note, which gives the after-training readout of
Section~\ref{sec:obs3}.
\end{itemize}

\textbf{Aggregate arms:}
\begin{itemize}[leftmargin=1em, nosep]
\item \textbf{all\_C}: the four libraries served together, at most \nAllCPer{} notes per component and
\nAllCTotal{} in total, exact matches first.
\item \textbf{all\_W}: one adapter trained on the locator, procedure and state-fact samples together, then
preference-optimized on the lessons from that adapter.
\item \textbf{routed}: each component sent where the rule fitted with the family held out says
(Appendix~\ref{app:rule}). The components sent to the weights are trained as in \textbf{all\_W} restricted to
those components; the components sent to the context are served with the \textbf{all\_C} quota. The assignment
is written to disk before the adapter is trained.
\item \textbf{anti\_routed}: the same with every destination reversed.
\item \textbf{routed\_self}: the destinations read from the family's own observed signs,
used as an in-sample routing reference; \llmname{GUI-Owl-8B} only.
\end{itemize}

\textbf{Whole-trajectory baselines:}
\begin{itemize}[leftmargin=1em, nosep]
\item \textbf{B1\_bon}: \nRetry{} samples of the same instance at temperature~1. The main-text number keeps the
first attempt that the agent declares complete, and the last attempt otherwise (self-retry).
\item \textbf{B2\_wholetraj\_sft}: supervised fine-tuning on every step of every successful pool episode.
\item \textbf{B3\_wholetraj\_k1}: the action sequence of the single successful pool episode with the most
similar task text, pasted into the prompt.
\item \textbf{B4\_atomic\_k10}: \kAtomic{} step-level records (state change, action) retrieved by task-text
similarity, without state matching.
\item \textbf{B5\_icl\_distill\_P, B5\_icl\_distill\_F}: context-then-distill. The context route of the
component runs on the experience templates as a teacher (\nDistillEp{} episodes per template), and its
successful trajectories are fine-tuned into the weights without a notes block.
\end{itemize}

\textbf{Interventions:}
\begin{itemize}[leftmargin=1em, nosep]
\item \textbf{W\_P\_dup$m$, W\_F\_dup$m$}: the source episodes of the component are copied $m$ times into the
pool before extraction, so the training set of the weights route, and with it the training dose of every key,
grows $m$-fold while the notes do not change; $m = 8$ on all three families, $m \in \{2, 4\}$ on
\llmname{GUI-Owl-8B} only.
\item \textbf{C\_P\_destate, C\_F\_destate}: the same notes retrieved by task-text similarity only, without
the state signature.
\item \textbf{C\_P\_elem, C\_F\_elem}: the same notes retrieved with a signature read from the screen, the set of
visible elements as (role, normalized name) pairs, in place of the environment state; thresholds and ranking
unchanged.
\item \textbf{W\_P\_lr$\eta$, W\_F\_lr$\eta$}: the samples of \textbf{W\_P} and \textbf{W\_F}, trained at learning
rate $\eta \in \{3{\times}10^{-5}, 10^{-4}\}$ instead of $10^{-5}$.
\item \textbf{W\_c\_rhi, W\_c\_rlo, C\_c\_rhi, C\_c\_rlo}, $c \in \{L, F\}$: the items of lessons or of state facts
of each pool split by item recurrence at the full-fit boundary for the mean $\kappa_0$ of the component
($r = \HalfThrL{}$ for lessons, \HalfThrF{} for state facts), each half sent to the weights and to the context as
the whole component is.
\item \textbf{B4\_atomic\_state\_k10}: the step-level records of \textbf{B4\_atomic\_k10} retrieved by the state
signature, exact matches first, instead of by task text.
\end{itemize}
The last four groups run on MobileGym with all three families and three seeds.

\textbf{Controls and reference arms:}
\begin{itemize}[leftmargin=1em, nosep]
\item \textbf{random\_cards}: \kNotes{} items drawn at random from other applications, in the same format.
\item \textbf{shuffle\_c\_C}: the payloads of a component's items permuted across its keys;
\textbf{shuffle\_c\_W}: the labels of the training samples permuted across samples (coordinates for locators,
actions for procedures and state facts, chosen and rejected swapped for lessons); \llmname{GUI-Owl-8B} only.
\item \textbf{Sensitivity variants} of the context and weights routes (Appendix~\ref{app:sensitivity}): Jaccard
thresholds of \jSensHi{} and \jSensLo{}, \kSensLo{} or \kSensHi{} notes, adapter ranks \rankSensLo{} and
\rankSensHi{}, pools of \poolSensLo{} or \poolSensHi{} episodes per template, library caps of \capSensLo{} and
\capSensHi{} items, a flat notes block of \nAllCTotalFlat{} items without
components, and routing by the online state-conditionality; \llmname{GUI-Owl-8B}, seed~0.
\end{itemize}

\subsection{Interaction Budget}
\label{app:budget}

All arms use the same held-out instances and the official per-episode step limit. Each arm runs one episode
per instance, except self-retry, which runs \nRetry{} attempts. The two routes consume the same items: the notes
and the training samples of a component are rendered from the same extracted items of the same pool. Two quantities the budget
does not equalize are reported in Appendix~\ref{app:compute}: the notes block that the context route adds to
every prompt, and the adapter training run that the weights route adds. Two actions count as the same when they
are of the same type and, for taps, hit the same element or land within \sameDiag{} of the screen diagonal of
each other; for text entry, carry the same stripped text; for scrolls, go in the same direction. This rule
decides the readout of Section~\ref{sec:obs3}, the policy gap of Section~\ref{sec:shifts} and the conflict
outcome of Section~\ref{sec:rule3}.

\clearpage
\section{Training Setup}
\label{app:training}

Every weights-route arm is a LoRA adapter trained on the base agent of its family and merged before serving.
The vision tower stays frozen, and a weights-route arm is evaluated with the prompt of the base agent, without a
notes block. Every adapter of every family uses the same configuration, fixed before the runs: rank
\loraRank{}, $\alpha$ \loraAlpha{}, \loraEpochs{} epochs, learning rate \loraLR{}. Table~\ref{tab:app_training}
lists the samples of every adapter and its token accuracy on a held-in tenth of them.

\subsection{Weights Route}
\label{app:weightsroute}

\paragraph{Samples.}
A locator item becomes one sample per source step: the screenshot and the instruction to tap the named element,
with the tap coordinates as the target. A procedure item becomes the step-by-step samples of the successful
trajectories of its task family, with the sub-goal sequence as the thought. A state-fact item becomes its source
step, task, screenshot and history, with the predicate as the thought and the action as the target. A lesson
item becomes a preference pair on its source state, the successful action as chosen and the failed action as
rejected, trained by preference optimization from the same adapter configuration. Success-filtered fine-tuning
takes every step of every successful pool episode; context-then-distill takes the successful teacher episodes of
the context route, at the sample count of the weights route of the same component; the copy interventions repeat the source episodes of a component before
extraction, so their sample counts are the multiples listed in the table.

\paragraph{Aggregate adapters.}
All to the weights trains one adapter on the locator, procedure and state-fact samples together and then runs
the lesson preference stage from that adapter. The routed and reverse-assignment adapters follow the same two
stages restricted to the components each assignment sends to the weights; their assignment is written to disk
before the adapter is trained.

\begin{table}[!htbp]
\centering
\caption{\textbf{Training runs.} Training samples and development token accuracy of every adapter, mean over the seeds the arm has. All adapters use LoRA rank \loraRank{}, $\alpha$ \loraAlpha{}, \loraEpochs{} epochs and learning rate \loraLR{} with the vision tower frozen, except the two learning-rate rows, which train the samples of the default adapter; lessons and the second stage of \textbf{all\_W} / \textbf{routed} use preference optimization with the same adapter configuration. The two halves of a component together train the samples of its full adapter.}
\label{tab:app_training}
\footnotesize
\setlength{\tabcolsep}{3.5pt}
\renewcommand{\arraystretch}{1.15}
\begin{tabularx}{\textwidth}{p{0.29\textwidth}|>{\raggedleft\arraybackslash}X>{\raggedleft\arraybackslash}X>{\raggedleft\arraybackslash}X|>{\centering\arraybackslash}X>{\centering\arraybackslash}X>{\centering\arraybackslash}X}
\toprule
\textbf{MobileGym} & \multicolumn{3}{c|}{\cellcolor{SearchBg}\textbf{Training samples}} & \multicolumn{3}{c}{\cellcolor{MathBg}\textbf{Dev.\ accuracy}} \\ \cmidrule(lr){2-4}\cmidrule(lr){5-7}
\textbf{Arm} & \textbf{GUI-Owl} & \textbf{Qwen3-VL} & \textbf{ScaleCUA} & \textbf{GUI-Owl} & \textbf{Qwen3-VL} & \textbf{ScaleCUA} \\
\midrule
Locators to the weights & 1211 & 1035 & 1056 & 0.952 & 0.949 & 0.959 \\
Procedures to the weights & 5721 & 5498 & 5798 & 0.932 & 0.946 & 0.946 \\
State facts to the weights & 476 & 450 & 401 & 0.954 & 0.954 & 0.961 \\
Lessons to the weights & 2854 & 2600 & 2763 & 0.931 & 0.934 & 0.940 \\
\midrule
All to the weights & 10263 & 9583 & 10018 & 0.942 & 0.930 & 0.941 \\
Routed (weights part) & 4065 & 3635 & 3819 & 0.941 & 0.922 & 0.932 \\
Reverse assignment & 6197 & -- & -- & 0.952 & -- & -- \\
\midrule
Success-filtered fine-tuning & 8210 & 7666 & 8014 & 0.942 & 0.937 & 0.929 \\
Context then distill: procedures & 5721 & 5498 & 5798 & 0.933 & 0.944 & 0.935 \\
Context then distill: state facts & 476 & 450 & 401 & 0.950 & 0.964 & 0.946 \\
\midrule
Procedures, items copied $\times 8$ & 45771 & -- & -- & 0.933 & -- & -- \\
State facts, items copied $\times 8$ & 3808 & -- & -- & 0.932 & -- & -- \\
\midrule
Procedures, learning rate $3\times10^{-5}$ & 5721 & 5498 & 5798 & 0.950 & 0.946 & 0.939 \\
Procedures, learning rate $10^{-4}$ & 5721 & 5498 & 5798 & 0.940 & 0.927 & 0.932 \\
State facts, learning rate $3\times10^{-5}$ & 476 & 450 & 401 & 0.957 & 0.946 & 0.954 \\
State facts, learning rate $10^{-4}$ & 476 & 450 & 401 & 0.959 & 0.955 & 0.951 \\
\midrule
Lessons, high-recurrence half & 1363 & 1175 & 1364 & 0.948 & 0.956 & 0.949 \\
Lessons, low-recurrence half & 1491 & 1425 & 1399 & 0.957 & 0.951 & 0.950 \\
State facts, high-recurrence half & 142 & 147 & 122 & 0.943 & 0.948 & 0.960 \\
State facts, low-recurrence half & 334 & 303 & 279 & 0.957 & 0.936 & 0.959 \\
\bottomrule
\end{tabularx}
\par\vspace{0.8em}
\begin{tabularx}{\textwidth}{p{0.29\textwidth}|>{\raggedleft\arraybackslash}X>{\raggedleft\arraybackslash}X>{\raggedleft\arraybackslash}X|>{\centering\arraybackslash}X>{\centering\arraybackslash}X>{\centering\arraybackslash}X}
\toprule
\textbf{AndroidWorld} & \multicolumn{3}{c|}{\cellcolor{SearchBg}\textbf{Training samples}} & \multicolumn{3}{c}{\cellcolor{MathBg}\textbf{Dev.\ accuracy}} \\ \cmidrule(lr){2-4}\cmidrule(lr){5-7}
\textbf{Arm} & \textbf{GUI-Owl} & \textbf{Qwen3-VL} & \textbf{ScaleCUA} & \textbf{GUI-Owl} & \textbf{Qwen3-VL} & \textbf{ScaleCUA} \\
\midrule
Locators to the weights & 658 & 578 & 611 & 0.943 & 0.956 & 0.958 \\
Procedures to the weights & 2997 & 2751 & 2759 & 0.947 & 0.943 & 0.944 \\
State facts to the weights & 203 & 230 & 224 & 0.952 & 0.944 & 0.946 \\
Lessons to the weights & 1349 & 1369 & 1540 & 0.947 & 0.956 & 0.962 \\
\midrule
All to the weights & 5208 & 4928 & 5134 & 0.932 & 0.944 & 0.930 \\
Routed (weights part) & 2007 & 1948 & 2151 & 0.949 & 0.949 & 0.950 \\
\bottomrule
\end{tabularx}
\end{table}

\subsection{Context Route}
\label{app:contextroute}

The notes of a component are its items rendered as one line each in a fixed template. A locator note names the
element, its role and its position; a procedure note gives the sub-goal sequence of its skill file and the next
sub-goal; a state-fact note states the condition that must hold before its action; a lesson note names the
action to avoid in its state and the action to take instead. At every step the library of the current
application is ranked by the Jaccard similarity between each item's source state signature and the current
signature, exact matches (at least \jExact{}) before relaxed matches (at least \jRelaxed{}). The top \kNotes{}
notes are appended to the prompt as a notes block, and the rest of the prompt is the base agent's. Procedures
are retrieved in two stages: the skill file of the most similar task text, then the step-level items within it.
The aggregate context arms serve at most \nAllCPer{} notes per component and \nAllCTotal{} in total, exact
matches first. Prompt tokens are recorded with and without the notes block on every step and enter
Appendix~\ref{app:compute}.

\clearpage
\section{The Rule Fit}
\label{app:rule}

The rule of Section~\ref{sec:rule1} is fitted by a procedure fixed before the runs. This appendix gives its
notation, the objective as implemented, the routine as run, and the per-cell held-out table behind
Figure~\ref{fig:rule}.

\subsection{Notation and Objective}
\label{app:notation}

{\small
\setlength{\tabcolsep}{5pt}
\renewcommand{\arraystretch}{1.18}
\noindent
\begin{tabularx}{\textwidth}{l|X}
\toprule
\textbf{Symbol} & \textbf{Description} \\
\midrule
$\mathcal{C}$ & the twenty-four cells, one per (component, backbone family, environment) \\
$\mathcal{C}_f$ & the eight cells of family $f$; the fold that holds $f$ out fits on $\mathcal{C} \setminus \mathcal{C}_f$ \\
$r(c),\ \kappa_0(c)$ & the pool coordinates of cell $c$: mean recurrence and median state-conditionality of its items \\
$\Delta(c),\ I(c)$ & the route gain of $c$ and its 95 percent paired-bootstrap interval \\
$y(c)$ & the sign of $\Delta(c)$, $+1$ toward the weights and $-1$ toward the context \\
$w(c)$ & the weight of the cell, $\tfrac{1}{2}$ when $0 \in I(c)$ and $1$ otherwise \\
$z(c)$ & the standardised coordinates, each of $r$ and $\kappa_0$ centred and scaled by the mean and standard deviation over the cells in the fit \\
$u,\ b$ & the two slopes on $z(c)$ and the unpenalised intercept \\
$\lambda$ & the ridge strength, one over the number of cells in the fit \\
$s_\theta(c)$ & the score on the raw scale, $\alpha\, r(c) - \beta\, \kappa_0(c) - \gamma$, obtained from $(u, b)$ by undoing the standardisation \\
$S$ & the held-out score, $\sum_{c} w(c)\, \mathds{1}[\operatorname{sgn} s_f(c) = y(c)]$, at most 24 \\
\bottomrule
\end{tabularx}\par}

\medskip
\noindent
The fit minimises
\begin{equation}
\frac{1}{\sum_{c} w(c)} \sum_{c \,\in\, \mathcal{C} \setminus \mathcal{C}_f} w(c)\,
\log\Bigl(1 + \exp\bigl[-y(c)\,(u^{\top} z(c) + b)\bigr]\Bigr) \;+\; \frac{\lambda}{2}\,\lVert u \rVert^2
\label{eq:app_objective}
\end{equation}
by full-batch gradient descent from zero with step size $0.1$ for twenty thousand iterations, as fixed before the
runs. The intercept is not penalised, and the result does not change under ten times as many
iterations. A cell of the held-out family goes to the weights when $s_f(c) > 0$ and to the context otherwise.
The point count of Section~\ref{sec:rule1} is the number of held-out cells whose predicted sign equals $y(c)$;
$S$ weights the cells whose interval crosses zero by one half.

\subsection{The Routine}
\label{app:routine}
Algorithm~\ref{alg:routing} is the complete procedure behind Sections~\ref{sec:setup}
and~\ref{sec:rule}: it measures the pool, fits the rule with one backbone family held out, routes every component
of the held-out family by the fitted sign, trains or serves accordingly, and reads the result on the held-out
instances. Notation follows Section~\ref{sec:setup} and Appendix~\ref{app:notation}.

\begin{algorithm}[!htbp]
\caption{Component routing, one held-out family
(\colorbox{bandblue}{Measure}, \colorbox{bandgreen}{Fit}, \colorbox{bandpurple}{Route to the weights},
\colorbox{bandorange}{Route to the context}, and \colorbox{bandred}{Evaluate})}
\label{alg:routing}
\textbf{Input:} the base agent $\pi_0$ of family $f$, its pool $\mathcal{D}$ of trajectories on the training
templates, the held-out instances $\mathcal{H}$, the cells $\mathcal{C}$ of the other families with their signs,
the extractor $\phi$, ridge strength $\lambda$
\begin{algorithmic}[1]
\STATE \textit{Phase I: Measure} \Comment{before any training; pool only}
\colorbox{bandblue}{
\parbox{0.9\linewidth}{
\STATE extract from every trajectory in $\mathcal{D}$ the items of the four components with $\phi$: a key, a
payload and a state condition each
\STATE for each component $c$, set $r(c)$ to the mean over its items of the share of unrelated tasks whose
episodes contain the key, and $\kappa_0(c)$ to the median state-conditionality over its valued items
}}

\STATE \textit{Phase II: Fit} \Comment{family $f$ is not in the fit}
\colorbox{bandgreen}{
\parbox{0.9\linewidth}{
\STATE standardise $r$ and $\kappa_0$ on the sixteen held-in cells $\mathcal{C} \setminus \mathcal{C}_f$
\STATE $(u, b) \gets$ the minimiser of Equation~\ref{eq:app_objective} by gradient descent from zero
\STATE for each component $c$ of family $f$: $s_f(c) \gets \alpha_f\, r(c) - \beta_f\, \kappa_0(c) - \gamma_f$
}}

\STATE \textit{Phase III: Route} \Comment{one destination per component, fixed before training}
\FORALL{components $c$ with $s_f(c) > 0$}
\colorbox{bandpurple}{
\parbox{0.88\linewidth}{
\STATE render every item of $c$ as a training sample; fine-tune an adapter on $\pi_0$ in the frozen
configuration of Appendix~\ref{app:training} (lessons by preference pairs, the other components by supervised steps)
}}
\ENDFOR
\FORALL{components $c$ with $s_f(c) \le 0$}
\colorbox{bandorange}{
\parbox{0.88\linewidth}{
\STATE render every item of $c$ as a note with its state condition; at each step retrieve at most \nAllCPer{}
notes of $c$, the quota of all to the context, whose key matches the current screen and whose condition holds, and
place them in the prompt
}}
\ENDFOR

\STATE \textit{Phase IV: Evaluate} \Comment{same held-out instances and per-episode step limit}
\colorbox{bandred}{
\parbox{0.9\linewidth}{
\STATE run the routed agent on $\mathcal{H}$ with the adapter of Phase III and the notes of Phase III
\STATE pair each episode with the base agent and the single-destination arms on the same instance and seed;
report the mean difference with a paired bootstrap of \nBoot{} resamples
}}
\RETURN the assignment $\{c \mapsto \wcol{W} \text{ or } \ccol{C}\}$ and the held-out success
\end{algorithmic}
\end{algorithm}

\subsection{Held-out Table}
\label{app:heldout}

\begin{figure}[!htbp]
  \centering
  \includegraphics[width=\textwidth]{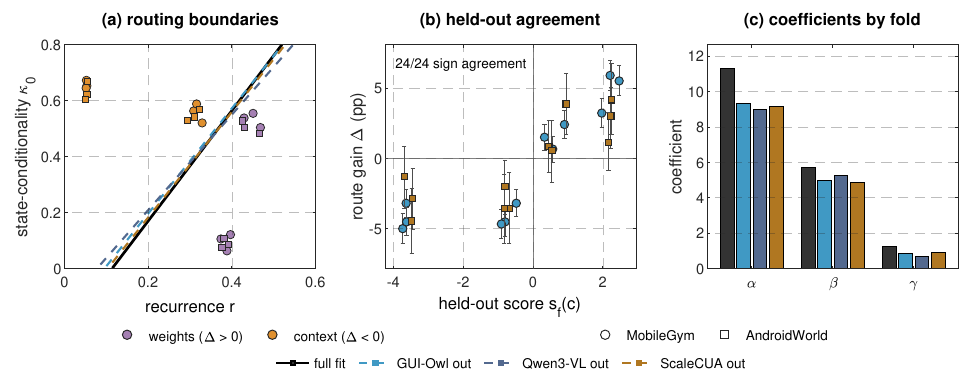}
  \caption{\textbf{The fit by fold.} (a)~The twenty-four cells, filled by the sign of their route gain, with the
  boundary of the full fit (solid) and of each fold (dashed). (b)~Held-out score against the observed route gain
  with its interval. (c)~Coefficients of the full fit and of the three folds.}
  \label{fig:app_fit}
\end{figure}

Table~\ref{tab:app_rulecoef} gives the coefficients of the full fit and of the three folds;
Table~\ref{tab:app_heldout} gives the twenty-four cells with their pool coordinates, route gains, weights and
held-out scores. The three folds agree in sign with the observed route gain in \SignAgree{} cells, \SignClear{}
of them with intervals clear of zero, for $S = \SignScore{}$. The full fit draws the boundary of
Figure~\ref{fig:rule}(a) and agrees with all twenty-four cells in sample. Under every fold the held-out family
receives the same assignment, locators and lessons to the weights and procedures and state facts to the
context, and this is the assignment the routed arm was trained with. Figure~\ref{fig:app_fit} draws the four
boundaries and the held-out scores against the observed gains. Table~\ref{tab:app_tsplit} repeats the fit on one
half of the held-out templates and reads routing on the other half.

\paragraph{Halves of a component.} Lessons and state facts lie closest to the boundary, so each of them is split
by item recurrence at the full-fit boundary for its mean $\kappa_0$, and both halves are sent to both routes
(Appendix~\ref{app:arms}). The halves differ in recurrence and hardly in state-conditionality, and no fit used
them: their destinations were read off the full fit before the runs, and the fold that holds a family out gives
the same side. Table~\ref{tab:app_halves} lists the twelve halves; every half above the boundary is better off in
the weights and every half below it in the context, with intervals clear of zero, by \HalfDiffL{} points between
the two halves of lessons and \HalfDiffF{} between those of state facts.

\begin{table}[!htbp]
\centering
\caption{\textbf{The rule fit.} Coefficients of $s_\theta(c) = \alpha\, r(c) - \beta\, \kappa_0(c) - \gamma$ for the full fit and for each fold that holds out one backbone family; the last column counts held-out cells whose predicted sign matches the observed sign (in-sample for the full fit).}
\label{tab:app_rulecoef}
\footnotesize
\setlength{\tabcolsep}{4.5pt}
\renewcommand{\arraystretch}{1.15}
\begin{tabularx}{\textwidth}{l>{\raggedleft\arraybackslash}X|>{\raggedleft\arraybackslash}X>{\raggedleft\arraybackslash}X>{\raggedleft\arraybackslash}X|>{\centering\arraybackslash}X}
\toprule
\multicolumn{2}{c|}{} & \multicolumn{3}{c|}{\cellcolor{MathBg}\textbf{Routing coefficients}} & \multicolumn{1}{c}{\cellcolor{DeltaBg}\textbf{Agreement}} \\ \cmidrule(lr){3-5}\cmidrule(lr){6-6}
\textbf{Fit} & \textbf{Cells} & $\boldsymbol{\alpha}$ & $\boldsymbol{\beta}$ & $\boldsymbol{\gamma}$ & \textbf{Sign} \\
\midrule
All 24 cells & 24 & 11.29 & 5.73 & 1.28 & 24/24 \\
\midrule
Held out: GUI-Owl-8B & 16 & 9.32 & 4.96 & 0.89 & 8/8 \\
Held out: Qwen3-VL-8B & 16 & 9.01 & 5.28 & 0.70 & 8/8 \\
Held out: ScaleCUA-7B & 16 & 9.16 & 4.86 & 0.94 & 8/8 \\
\bottomrule
\end{tabularx}
\end{table}

\begin{table}[!htbp]
\centering
\caption{\textbf{The twenty-four cells.} Pool coordinates, route gain with its 95 percent paired-bootstrap interval (points), weight $w$ (one half when the interval crosses zero), held-out score, and predicted / observed destination. The score of a cell comes from the fold that excludes its family.}
\label{tab:app_heldout}
\footnotesize
\setlength{\tabcolsep}{3pt}
\renewcommand{\arraystretch}{1.15}
\begin{tabularx}{\textwidth}{p{55pt}p{75pt}|>{\centering\arraybackslash}p{30pt}>{\centering\arraybackslash}p{30pt}|>{\centering\arraybackslash}X|>{\centering\arraybackslash}p{16pt}>{\raggedleft\arraybackslash}p{36pt}>{\centering\arraybackslash}p{63pt}}
\toprule
\multicolumn{2}{l|}{\textbf{MobileGym}} & \multicolumn{2}{c|}{\cellcolor{SearchBg}\textbf{Pool coordinates}} & \multicolumn{1}{c|}{\cellcolor{DeltaBg}\textbf{Route gain}} & \multicolumn{3}{c}{\cellcolor{MathBg}\textbf{Held-out rule}} \\ \cmidrule(lr){3-4}\cmidrule(lr){5-5}\cmidrule(lr){6-8}
\textbf{Component} & \textbf{Family} & $\boldsymbol{r}$ & $\boldsymbol{\kappa_0}$ & $\boldsymbol{\Delta}$ \textbf{[95\%]} & $\boldsymbol{w}$ & $\boldsymbol{s_f}$ & \textbf{Pred.\ / obs.} \\
\midrule
\multirow{3}{*}{Locators} & GUI-Owl-8B & 0.397 & 0.121 & +5.9 [+4.9, +7.0] & 1 & +2.21 & \wcol{W} / \wcol{W} \\
 & Qwen3-VL-8B & 0.388 & 0.064 & +5.5 [+4.5, +6.6] & 1 & +2.47 & \wcol{W} / \wcol{W} \\
 & ScaleCUA-7B & 0.374 & 0.106 & +3.2 [+2.2, +4.3] & 1 & +1.97 & \wcol{W} / \wcol{W} \\
\midrule
\multirow{3}{*}{Procedures} & GUI-Owl-8B & 0.052 & 0.673 & $-$5.0 [$-$6.0, $-$4.0] & 1 & -3.74 & \ccol{C} / \ccol{C} \\
 & Qwen3-VL-8B & 0.051 & 0.645 & $-$3.2 [$-$4.2, $-$2.2] & 1 & -3.64 & \ccol{C} / \ccol{C} \\
 & ScaleCUA-7B & 0.053 & 0.653 & $-$4.5 [$-$5.5, $-$3.5] & 1 & -3.63 & \ccol{C} / \ccol{C} \\
\midrule
\multirow{3}{*}{State facts} & GUI-Owl-8B & 0.308 & 0.564 & $-$4.5 [$-$5.6, $-$3.4] & 1 & -0.81 & \ccol{C} / \ccol{C} \\
 & Qwen3-VL-8B & 0.329 & 0.521 & $-$3.2 [$-$4.2, $-$2.2] & 1 & -0.48 & \ccol{C} / \ccol{C} \\
 & ScaleCUA-7B & 0.316 & 0.589 & $-$4.7 [$-$5.7, $-$3.6] & 1 & -0.91 & \ccol{C} / \ccol{C} \\
\midrule
\multirow{3}{*}{Lessons} & GUI-Owl-8B & 0.451 & 0.555 & +0.7 [$-$0.3, +1.6] & 0.5 & +0.56 & \wcol{W} / \wcol{W} \\
 & Qwen3-VL-8B & 0.429 & 0.538 & +1.5 [+0.5, +2.4] & 1 & +0.33 & \wcol{W} / \wcol{W} \\
 & ScaleCUA-7B & 0.468 & 0.504 & +2.4 [+1.4, +3.4] & 1 & +0.90 & \wcol{W} / \wcol{W} \\
\bottomrule
\end{tabularx}
\par\vspace{0.8em}
\begin{tabularx}{\textwidth}{p{55pt}p{75pt}|>{\centering\arraybackslash}p{30pt}>{\centering\arraybackslash}p{30pt}|>{\centering\arraybackslash}X|>{\centering\arraybackslash}p{16pt}>{\raggedleft\arraybackslash}p{36pt}>{\centering\arraybackslash}p{63pt}}
\toprule
\multicolumn{2}{l|}{\textbf{AndroidWorld}} & \multicolumn{2}{c|}{\cellcolor{SearchBg}\textbf{Pool coordinates}} & \multicolumn{1}{c|}{\cellcolor{DeltaBg}\textbf{Route gain}} & \multicolumn{3}{c}{\cellcolor{MathBg}\textbf{Held-out rule}} \\ \cmidrule(lr){3-4}\cmidrule(lr){5-5}\cmidrule(lr){6-8}
\textbf{Component} & \textbf{Family} & $\boldsymbol{r}$ & $\boldsymbol{\kappa_0}$ & $\boldsymbol{\Delta}$ \textbf{[95\%]} & $\boldsymbol{w}$ & $\boldsymbol{s_f}$ & \textbf{Pred.\ / obs.} \\
\midrule
\multirow{3}{*}{Locators} & GUI-Owl-8B & 0.376 & 0.075 & +4.2 [+1.7, +6.8] & 1 & +2.25 & \wcol{W} / \wcol{W} \\
 & Qwen3-VL-8B & 0.381 & 0.109 & +1.1 [$-$0.9, +3.3] & 0.5 & +2.16 & \wcol{W} / \wcol{W} \\
 & ScaleCUA-7B & 0.392 & 0.087 & +3.0 [+0.7, +5.0] & 1 & +2.23 & \wcol{W} / \wcol{W} \\
\midrule
\multirow{3}{*}{Procedures} & GUI-Owl-8B & 0.053 & 0.623 & $-$4.5 [$-$6.8, $-$2.2] & 1 & -3.48 & \ccol{C} / \ccol{C} \\
 & Qwen3-VL-8B & 0.050 & 0.606 & $-$2.9 [$-$5.0, $-$0.7] & 1 & -3.45 & \ccol{C} / \ccol{C} \\
 & ScaleCUA-7B & 0.054 & 0.669 & $-$1.3 [$-$3.6, +0.9] & 0.5 & -3.70 & \ccol{C} / \ccol{C} \\
\midrule
\multirow{3}{*}{State facts} & GUI-Owl-8B & 0.310 & 0.542 & $-$3.6 [$-$6.0, $-$1.0] & 1 & -0.68 & \ccol{C} / \ccol{C} \\
 & Qwen3-VL-8B & 0.322 & 0.570 & $-$2.0 [$-$4.0, $-$0.1] & 1 & -0.81 & \ccol{C} / \ccol{C} \\
 & ScaleCUA-7B & 0.293 & 0.529 & $-$3.6 [$-$6.0, $-$1.1] & 1 & -0.83 & \ccol{C} / \ccol{C} \\
\midrule
\multirow{3}{*}{Lessons} & GUI-Owl-8B & 0.424 & 0.528 & +0.9 [$-$1.0, +2.7] & 0.5 & +0.45 & \wcol{W} / \wcol{W} \\
 & Qwen3-VL-8B & 0.466 & 0.484 & +3.9 [+1.7, +6.0] & 1 & +0.95 & \wcol{W} / \wcol{W} \\
 & ScaleCUA-7B & 0.429 & 0.504 & +0.6 [$-$1.7, +2.7] & 0.5 & +0.54 & \wcol{W} / \wcol{W} \\
\bottomrule
\end{tabularx}
\end{table}

\begin{table}[!htbp]
\centering
\caption{\textbf{Split of the held-out templates.} The held-out templates of each environment are sorted by identifier and split alternately within each tier into two halves (113 and 111 on MobileGym, 29 and 29 on AndroidWorld). The rule is fitted with one family held out on the route gains of one half and scored on the other half. Assignments give the destinations of locators, procedures, state facts and lessons (MobileGym / AndroidWorld where the two differ); the gains on the scoring half are routing minus the better single destination of each family (Single) and minus the reverse assignment (Reverse), points, mean over the three families.}
\label{tab:app_tsplit}
\footnotesize
\setlength{\tabcolsep}{3.5pt}
\renewcommand{\arraystretch}{1.15}
\begin{tabularx}{\textwidth}{l|>{\centering\arraybackslash}X>{\centering\arraybackslash}X>{\centering\arraybackslash}X|>{\raggedleft\arraybackslash}X>{\raggedleft\arraybackslash}X|>{\raggedleft\arraybackslash}X>{\raggedleft\arraybackslash}X}
\toprule
& \multicolumn{3}{c|}{\cellcolor{MathBg}\textbf{Assignment of the held-out family}} & \multicolumn{2}{c|}{\cellcolor{DeltaBg}\textbf{MobileGym}} & \multicolumn{2}{c}{\cellcolor{ScienceBg}\textbf{AndroidWorld}} \\ \cmidrule(lr){2-4}\cmidrule(lr){5-6}\cmidrule(lr){7-8}
\textbf{Fit $\rightarrow$ score} & \textbf{GUI-Owl} & \textbf{Qwen3-VL} & \textbf{ScaleCUA} & \textbf{Single} & \textbf{Reverse} & \textbf{Single} & \textbf{Reverse} \\
\midrule
Half 1 $\rightarrow$ half 2 & \wcol{W}\ccol{C}\ccol{C}\wcol{W} & \wcol{W}\ccol{C}\ccol{C}\ccol{C} / \wcol{W}\ccol{C}\ccol{C}\wcol{W} & \wcol{W}\ccol{C}\ccol{C}\wcol{W} & +3.7 & +8.5 & +2.4 & +6.8 \\
Half 2 $\rightarrow$ half 1 & \wcol{W}\ccol{C}\ccol{C}\wcol{W} & \wcol{W}\ccol{C}\ccol{C}\wcol{W} & \wcol{W}\ccol{C}\ccol{C}\wcol{W} & +5.1 & +9.3 & +2.6 & +6.6 \\
\bottomrule
\end{tabularx}
\end{table}

\begin{table}[!htbp]
\centering
\caption{\textbf{The twelve halves.} Lessons and state facts of every MobileGym pool split by item recurrence at the full-fit boundary for the mean $\kappa_0$ of the component ($r = 0.377$ for lessons, $0.394$ for state facts); each half goes to the context and to the weights on the same held-out instances. Items over the three seeds, pool coordinates, route gain with its 95 percent paired-bootstrap interval (points), score of the fold that holds the family out, and predicted / observed destination. No fit used the halves, and their predicted destinations were fixed before the runs.}
\label{tab:app_halves}
\footnotesize
\setlength{\tabcolsep}{3pt}
\renewcommand{\arraystretch}{1.15}
\begin{tabularx}{\textwidth}{p{78pt}p{56pt}|>{\raggedleft\arraybackslash}p{26pt}>{\centering\arraybackslash}p{28pt}>{\centering\arraybackslash}p{28pt}|>{\centering\arraybackslash}X|>{\raggedleft\arraybackslash}p{28pt}>{\centering\arraybackslash}p{56pt}}
\toprule
\multicolumn{2}{l|}{\textbf{MobileGym}} & \multicolumn{3}{c|}{\cellcolor{SearchBg}\textbf{Half of the pool}} & \multicolumn{1}{c|}{\cellcolor{DeltaBg}\textbf{Route gain}} & \multicolumn{2}{c}{\cellcolor{MathBg}\textbf{Held-out rule}} \\ \cmidrule(lr){3-5}\cmidrule(lr){6-6}\cmidrule(lr){7-8}
\textbf{Half} & \textbf{Family} & \textbf{Items} & $\boldsymbol{r}$ & $\boldsymbol{\kappa_0}$ & $\boldsymbol{\Delta}$ \textbf{[95\%]} & $\boldsymbol{s_f}$ & \textbf{Pred.\ / obs.} \\
\midrule
\multirow{3}{*}{Lessons, high $r$} & GUI-Owl-8B & 4090 & 0.796 & 0.557 & +2.4 [+1.4, +3.5] & +3.77 & \wcol{W} / \wcol{W} \\
 & Qwen3-VL-8B & 3525 & 0.782 & 0.535 & +2.2 [+1.2, +3.2] & +3.52 & \wcol{W} / \wcol{W} \\
 & ScaleCUA-7B & 4092 & 0.806 & 0.506 & +2.8 [+1.7, +3.8] & +3.99 & \wcol{W} / \wcol{W} \\
\midrule
\multirow{3}{*}{Lessons, low $r$} & GUI-Owl-8B & 4472 & 0.134 & 0.549 & $-$1.6 [$-$2.6, $-$0.7] & -2.36 & \ccol{C} / \ccol{C} \\
 & Qwen3-VL-8B & 4276 & 0.138 & 0.540 & $-$2.1 [$-$3.0, $-$1.1] & -2.30 & \ccol{C} / \ccol{C} \\
 & ScaleCUA-7B & 4196 & 0.138 & 0.501 & $-$1.8 [$-$2.7, $-$0.8] & -2.11 & \ccol{C} / \ccol{C} \\
\midrule
\multirow{3}{*}{State facts, high $r$} & GUI-Owl-8B & 427 & 0.727 & 0.518 & +2.5 [+1.5, +3.5] & +3.32 & \wcol{W} / \wcol{W} \\
 & Qwen3-VL-8B & 441 & 0.727 & 0.528 & +3.2 [+2.2, +4.2] & +3.07 & \wcol{W} / \wcol{W} \\
 & ScaleCUA-7B & 367 & 0.730 & 0.615 & +2.9 [+1.9, +3.9] & +2.76 & \wcol{W} / \wcol{W} \\
\midrule
\multirow{3}{*}{State facts, low $r$} & GUI-Owl-8B & 1001 & 0.130 & 0.573 & $-$1.8 [$-$2.7, $-$0.9] & -2.52 & \ccol{C} / \ccol{C} \\
 & Qwen3-VL-8B & 908 & 0.135 & 0.515 & $-$2.2 [$-$3.2, $-$1.3] & -2.19 & \ccol{C} / \ccol{C} \\
 & ScaleCUA-7B & 836 & 0.134 & 0.586 & $-$2.2 [$-$3.1, $-$1.3] & -2.56 & \ccol{C} / \ccol{C} \\
\bottomrule
\end{tabularx}
\end{table}

\clearpage
\section{Full Results}
\label{app:results}

\begin{figure}[!htbp]
  \centering
  \includegraphics[width=\textwidth]{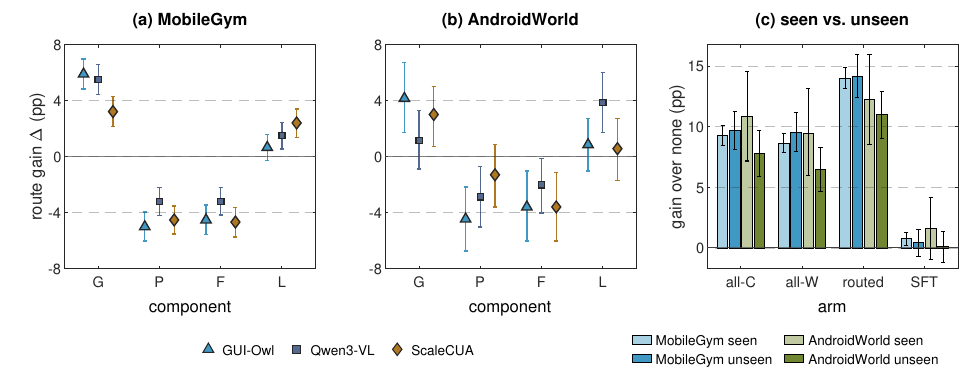}
  \caption{\textbf{Per-cell route gains and the split.} (a)~MobileGym and (b)~AndroidWorld: the route gain of
  every backbone and component with its 95 percent interval. (c)~Gain over the base agent on seen and unseen
  applications, pooled over backbones and seeds.}
  \label{fig:app_cells}
\end{figure}

The tables of this appendix expand the main-text measurements into their per-family form. Cells are means
over the \nSeeds{} seeds with the seed standard deviation as subscript where an arm has three seeds; every
difference is paired over (template, instance, seed). The twenty-four cells of the rule are in
Table~\ref{tab:app_heldout}; Figure~\ref{fig:app_cells} draws them with their intervals and adds the gains on
seen and unseen applications.

\subsection{All Intervention Arms}
\label{app:interventions}
\begin{figure}[!htbp]
  \centering
  \includegraphics[width=\textwidth]{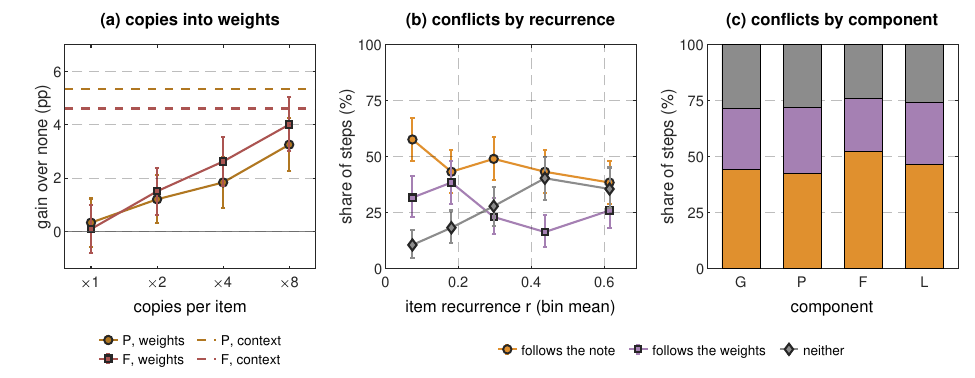}
  \caption{\textbf{Copies and conflicts.} (a)~Gain of procedures and state facts written into the weights with
  each item copied one to eight times on \llmname{GUI-Owl-8B}; dashed lines are the same components served as
  notes. (b)~At the divergence steps of OpenApps, the share of steps that follow the note, the weights or
  neither, by recurrence bin, and (c)~by component.}
  \label{fig:app_dose}
\end{figure}

Table~\ref{tab:interventions_full} lists every intervention arm of Section~\ref{sec:rule2}, including the copy
arms with two and four copies and the shuffled-key controls, which were run on \llmname{GUI-Owl-8B} only. It adds
three checks run on all three families: notes retrieved by the screen elements instead of the environment state
cost the context route \ElemCostP{} points for procedures and \ElemCostF{} for state facts, and $\kappa_0$ read with
that signature ranks the items as before (Spearman \ElemKappaRho{}); a learning rate ten times the default raises
the weights route by \LrTenGainP{} points for procedures and \LrTenGainF{} for state facts and leaves their route
gains at \LrTenDeltaP{} and \LrTenDeltaF{} (\LrThreeDeltaP{} and \LrThreeDeltaF{} at three times the default); and
ten step records retrieved by the state signature gain \StateStepGain{}, \StateStepShift{} over the same records
retrieved by task text.
Figure~\ref{fig:app_dose} draws the copy arms as a dose and adds the divergence steps of Section~\ref{sec:rule3}
by recurrence and by component.

\begin{table}[!htbp]
\centering
\caption{\textbf{All intervention arms.} Gain over the base agent (points) on MobileGym when one property of a component is changed: copying the training items raises the training dose of every key, as recurrence across tasks does; retrieving the notes by task text alone removes their state condition. The screen-element signature replaces the environment state in retrieval, the two learning rates train the same samples as the default $10^{-5}$, and the last block retrieves ten step records by task text or by the state signature. Subscripts are the std over three seeds; a dash marks an arm run on \llmname{GUI-Owl-8B} only. $\Delta$ is the shift from the unmodified route above it (\up{rise} / \down{drop}), averaged over the backbones the arm has.}
\label{tab:interventions_full}
\footnotesize
\setlength{\tabcolsep}{4.5pt}
\renewcommand{\arraystretch}{1.15}
\begin{tabularx}{\textwidth}{l|*{3}{>{\centering\arraybackslash}X}|c}
\toprule
& \multicolumn{3}{c|}{\cellcolor{SearchBg}\textbf{MobileGym, gain over the base agent}} & \\
\cmidrule(lr){2-4}
\textbf{Arm} & \textbf{GUI-Owl-8B} & \textbf{Qwen3-VL-8B} & \textbf{ScaleCUA-7B} & \textbf{$\Delta$} \\
\midrule
Procedures to the context & +5.3\std{0.7} & +5.2\std{0.2} & +5.1\std{0.0} & -- \\
\quad state condition removed & +1.4\std{0.8} & +1.8\std{0.1} & +0.8\std{0.4} & \down{3.9} \\
\quad signature read from the screen elements & +4.0\std{0.4} & +5.5\std{0.2} & +3.9\std{0.3} & \down{0.7} \\
Procedures to the weights & +0.3\std{0.6} & +2.0\std{0.9} & +0.6\std{0.3} & -- \\
\quad items copied $\times 2$ & +1.2\std{1.8} & -- & -- & \up{0.9} \\
\quad items copied $\times 4$ & +1.8\std{1.4} & -- & -- & \up{1.5} \\
\quad items copied $\times 8$ & +3.3\std{0.6} & +6.1\std{0.3} & +3.9\std{0.5} & \up{3.5} \\
\quad learning rate $3\times10^{-5}$ & +0.9\std{0.6} & +3.1\std{0.0} & +1.4\std{0.5} & \up{0.8} \\
\quad learning rate $10^{-4}$ & +1.1\std{1.0} & +3.2\std{0.8} & +1.5\std{0.8} & \up{1.0} \\
\midrule
State facts to the context & +4.6\std{0.4} & +5.4\std{0.6} & +4.9\std{0.3} & -- \\
\quad state condition removed & +1.1\std{0.7} & +3.1\std{0.5} & +0.6\std{0.9} & \down{3.4} \\
\quad signature read from the screen elements & +4.0\std{0.3} & +4.9\std{1.1} & +3.9\std{0.9} & \down{0.7} \\
State facts to the weights & +0.1\std{0.1} & +2.2\std{0.8} & +0.2\std{0.3} & -- \\
\quad items copied $\times 2$ & +1.5\std{1.1} & -- & -- & \up{1.4} \\
\quad items copied $\times 4$ & +2.6\std{0.6} & -- & -- & \up{2.5} \\
\quad items copied $\times 8$ & +4.0\std{1.2} & +5.8\std{0.6} & +3.4\std{1.3} & \up{3.6} \\
\quad learning rate $3\times10^{-5}$ & +0.7\std{0.8} & +3.6\std{0.2} & +0.5\std{0.7} & \up{0.7} \\
\quad learning rate $10^{-4}$ & +1.4\std{1.3} & +3.6\std{0.6} & +0.3\std{0.7} & \up{1.0} \\
\midrule
Ten step records retrieved by task text & +1.4\std{1.0} & +1.2\std{0.2} & +1.7\std{1.4} & -- \\
\quad retrieved by the state signature & +2.1\std{0.9} & +1.1\std{0.2} & +1.5\std{0.7} & \up{0.2} \\
\midrule
Shuffled keys, worst of eight arms & $-$1.1 & -- & -- & -- \\
Random notes from other applications & $-$1.5\std{0.6} & $-$1.7\std{0.3} & $-$1.6\std{0.6} & -- \\
\bottomrule
\end{tabularx}
\end{table}

\subsection{AndroidWorld Contrasts}
\label{app:awcontrasts}
Table~\ref{tab:app_awcontrasts} gives, per family, the AndroidWorld contrasts behind Section~\ref{sec:rule4}:
routing against each single destination and against the reverse assignment, with paired-bootstrap intervals.

\begin{table}[!htbp]
\centering
\caption{\textbf{AndroidWorld contrasts.} Held-out success of routing minus each aggregate arm (points) with 95 percent paired-bootstrap intervals over the \nAWPairsPerSeed{} instances and \nSeeds{} seeds of each family.}
\label{tab:app_awcontrasts}
\footnotesize
\setlength{\tabcolsep}{4.5pt}
\renewcommand{\arraystretch}{1.15}
\begin{tabularx}{\textwidth}{l|>{\centering\arraybackslash}X>{\centering\arraybackslash}X>{\centering\arraybackslash}X}
\toprule
& \multicolumn{3}{c}{\cellcolor{ScienceBg}\textbf{AndroidWorld: routing gain (points)}} \\ \cmidrule(lr){2-4}
\textbf{Family} & \textbf{vs.\ all to the context} & \textbf{vs.\ all to the weights} & \textbf{vs.\ reverse assignment} \\
\midrule
GUI-Owl-8B & $-$0.3 [$-$2.6, +2.0] & +4.3 [+1.7, +6.8] & +5.6 [+2.9, +8.2] \\
Qwen3-VL-8B & +5.2 [+2.7, +7.6] & +5.0 [+2.4, +7.6] & +9.8 [+6.8, +12.6] \\
ScaleCUA-7B & +3.6 [+1.3, +5.9] & +3.2 [+0.7, +5.6] & +4.7 [+2.3, +7.3] \\
\bottomrule
\end{tabularx}
\end{table}

\clearpage
\subsection{Producers and Consumers}
\label{app:ladderfull}
Table~\ref{tab:app_ladder} lists every producer and consumer pair of Section~\ref{sec:shifts} with the gain of
the procedures route by notes, by weights, by an equal mixture of the two producers' samples, and by practice
on the consumer's own successful families.

\begin{table}[!htbp]
\centering
\caption{\textbf{Producers and consumers.} Gain over the base agent (points) of the procedures route for every producer and consumer pair of Section~\ref{sec:shifts}: notes from the producer, weights trained on the producer, an equal mixture, and the practice arm (the consumer trained on its own successful families only). Every adapter of this batch trains on \nBudgetB{} samples (\nBudgetBL{} for the 32B consumers); subscripts are the std over seeds.}
\label{tab:app_ladder}
\footnotesize
\setlength{\tabcolsep}{4.5pt}
\renewcommand{\arraystretch}{1.15}
\begin{tabularx}{\textwidth}{ll|>{\centering\arraybackslash}X>{\centering\arraybackslash}X>{\centering\arraybackslash}X|>{\centering\arraybackslash}X}
\toprule
\multicolumn{2}{c|}{} & \multicolumn{3}{c|}{\cellcolor{SearchBg}\textbf{Producer experience}} & \multicolumn{1}{c}{\cellcolor{ScienceBg}\textbf{Own experience}} \\ \cmidrule(lr){3-5}\cmidrule(lr){6-6}
\textbf{Consumer} & \textbf{Producer} & \textbf{Notes} & \textbf{Weights} & \textbf{Mixed} & \textbf{Practice} \\
\midrule
\multirow{3}{*}{Qwen3-VL-4B} & Qwen3-VL-4B & +1.5\std{0.7} & $-$0.3\std{0.7} & -- & $-$0.3\std{1.0} \\
 & Qwen3-VL-8B & +1.9\std{0.5} & $-$2.7\std{0.5} & $-$0.5\std{0.9} & -- \\
 & Qwen3-VL-32B & +3.1\std{0.5} & $-$1.9\std{0.7} & +0.0\std{0.4} & -- \\
\midrule
\multirow{5}{*}{Qwen3-VL-8B} & Qwen3-VL-4B & +2.2\std{0.5} & $-$2.3\std{1.0} & $-$0.8\std{0.6} & -- \\
 & Qwen3-VL-8B & +3.6\std{0.5} & +0.9\std{1.5} & -- & +0.6\std{1.5} \\
 & Qwen3-VL-32B & +4.8\std{1.2} & $-$1.4\std{0.7} & +0.5\std{0.1} & -- \\
 & GUI-Owl-8B & +3.6\std{0.2} & $-$7.5\std{0.4} & $-$3.2\std{0.5} & -- \\
 & GUI-Owl-8B, retrained & +4.4\std{0.5} & $-$2.0\std{1.1} & +0.6\std{0.8} & -- \\
\midrule
\multirow{3}{*}{Qwen3-VL-32B} & Qwen3-VL-4B & +1.8\std{0.4} & $-$2.5\std{0.8} & +0.1\std{0.5} & -- \\
 & Qwen3-VL-8B & +3.2\std{0.9} & $-$1.4\std{0.6} & +1.0\std{0.6} & -- \\
 & Qwen3-VL-32B & +5.5\std{0.6} & +2.7\std{0.9} & -- & +2.9\std{0.4} \\
\midrule
\multirow{2}{*}{GUI-Owl-8B} & GUI-Owl-8B & +3.6\std{1.3} & +2.2\std{1.0} & -- & +1.9\std{1.6} \\
 & GUI-Owl-32B & +5.1\std{0.2} & $-$0.3\std{0.6} & +2.5\std{0.6} & -- \\
\midrule
\multirow{2}{*}{GUI-Owl-32B} & GUI-Owl-8B & +6.9\std{0.3} & $-$0.4\std{0.3} & +2.9\std{0.5} & -- \\
 & GUI-Owl-32B & +7.8\std{0.6} & +5.3\std{0.8} & -- & +6.3\std{0.2} \\
\bottomrule
\end{tabularx}
\end{table}

\subsection{Readout}
\label{app:readoutfull}
Table~\ref{tab:app_readout} gives the readout rates of Section~\ref{sec:obs3} per consumer and component,
before and after the component is in the weights, and by thirds of item recurrence within each component.

\begin{table}[!htbp]
\centering
\caption{\textbf{Readout rates.} Share of note-bearing steps (percent) whose action changes when the note is removed, before training (the base agent with the notes) and after (the adapter of the same component with the same notes), per consumer and component, pooled over seeds. The lower panel splits the items of each component into thirds of recurrence, pooled over consumers and seeds, and adds the after-to-before ratio.}
\label{tab:app_readout}
\footnotesize
\setlength{\tabcolsep}{4.5pt}
\renewcommand{\arraystretch}{1.15}
\begin{tabularx}{\textwidth}{l|>{\centering\arraybackslash}X>{\centering\arraybackslash}X|>{\centering\arraybackslash}X>{\centering\arraybackslash}X|>{\centering\arraybackslash}X>{\centering\arraybackslash}X|>{\centering\arraybackslash}X>{\centering\arraybackslash}X}
\toprule
& \multicolumn{2}{c|}{\cellcolor{SearchBg}\textbf{Locators}} & \multicolumn{2}{c|}{\cellcolor{MathBg}\textbf{Procedures}} & \multicolumn{2}{c|}{\cellcolor{ScienceBg}\textbf{State facts}} & \multicolumn{2}{c}{\cellcolor{AgenticBg}\textbf{Lessons}} \\ \cmidrule(lr){2-3}\cmidrule(lr){4-5}\cmidrule(lr){6-7}\cmidrule(lr){8-9}
\textbf{Consumer} & \textbf{Before} & \textbf{After} & \textbf{Before} & \textbf{After} & \textbf{Before} & \textbf{After} & \textbf{Before} & \textbf{After} \\
\midrule
GUI-Owl-8B & 45 & 17 & 33 & 23 & 43 & 17 & 50 & 15 \\
Qwen3-VL-8B & 45 & 17 & 32 & 22 & 41 & 19 & 46 & 14 \\
ScaleCUA-7B & 46 & 16 & 35 & 22 & 42 & 19 & 50 & 15 \\
\bottomrule
\end{tabularx}
\par\vspace{0.8em}
\begin{tabularx}{\textwidth}{l|>{\centering\arraybackslash}X>{\centering\arraybackslash}X>{\centering\arraybackslash}X|>{\centering\arraybackslash}X>{\centering\arraybackslash}X>{\centering\arraybackslash}X|>{\centering\arraybackslash}X>{\centering\arraybackslash}X>{\centering\arraybackslash}X}
\toprule
& \multicolumn{3}{c|}{\cellcolor{SearchBg}\textbf{Lowest third of $r$}} & \multicolumn{3}{c|}{\cellcolor{MathBg}\textbf{Middle third}} & \multicolumn{3}{c}{\cellcolor{ScienceBg}\textbf{Highest third}} \\ \cmidrule(lr){2-4}\cmidrule(lr){5-7}\cmidrule(lr){8-10}
\textbf{Component} & \textbf{Before} & \textbf{After} & \textbf{Ratio} & \textbf{Before} & \textbf{After} & \textbf{Ratio} & \textbf{Before} & \textbf{After} & \textbf{Ratio} \\
\midrule
Locators & 30 & 22 & 0.73 & 40 & 19 & 0.48 & 65 & 7 & 0.11 \\
Procedures & 29 & 22 & 0.77 & 33 & 23 & 0.69 & 38 & 22 & 0.59 \\
State facts & 30 & 22 & 0.72 & 38 & 21 & 0.55 & 57 & 12 & 0.21 \\
Lessons & 32 & 20 & 0.63 & 44 & 16 & 0.37 & 69 & 8 & 0.11 \\
\bottomrule
\end{tabularx}
\end{table}

\clearpage
\section{Sensitivity}
\label{app:sensitivity}

Each variant of Table~\ref{tab:app_sensitivity} changes one setting of a default arm on \llmname{GUI-Owl-8B},
seed~0, and leaves everything else as in Appendix~\ref{app:arms}. The table reports the gain of the default arm
over the base agent on the same instances, the gain of the variant, and their difference.

\paragraph{Matching thresholds.}
The exact and relaxed tiers of the context route move from \jExact{} and \jRelaxed{} to \jSensHi{} and
\jSensLo{}, for all four components.

\paragraph{Notes per step.}
The context route of locators and state facts serves \kSensLo{} or \kSensHi{} notes instead of \kNotes{}.

\paragraph{Adapter rank.}
The weights route of locators and state facts is trained at rank \rankSensLo{} and \rankSensHi{} instead of
\loraRank{}, at the same number of epochs.

\paragraph{Pool size.}
Both routes of locators and state facts are built from nested subsets of \poolSensLo{} and \poolSensHi{}
episodes per template instead of \nEpPerTemplate{}.

\paragraph{Library cap.}
The locator library is capped at \capSensLo{} and \capSensHi{} items.

\paragraph{Notes block and rule input.}
All to the context serves \nAllCTotalFlat{} notes without the per-component quota. Routing reads the online
state-conditionality of Equation~\ref{eq:kappa} instead of $\kappa_0$ when the rule assigns the destinations.

\begin{table}[!htbp]
\centering
\caption{\textbf{Sensitivity.} Gain over the base agent (points) on \llmname{GUI-Owl-8B}, seed 0, when one setting of the default arm is changed: the default arm, the variant, and their difference.}
\label{tab:app_sensitivity}
\footnotesize
\setlength{\tabcolsep}{4pt}
\renewcommand{\arraystretch}{1.12}
\begin{tabularx}{\textwidth}{lll|>{\raggedleft\arraybackslash}X>{\raggedleft\arraybackslash}X|>{\raggedleft\arraybackslash}X}
\toprule
\multicolumn{3}{c|}{} & \multicolumn{2}{c|}{\cellcolor{SearchBg}\textbf{Gain over base (points)}} & \multicolumn{1}{c}{\cellcolor{DeltaBg}\textbf{Difference}} \\ \cmidrule(lr){4-5}\cmidrule(lr){6-6}
\textbf{Setting} & \textbf{Variant} & \textbf{Arm} & \textbf{Default} & \textbf{Variant} & \textbf{Shift} \\
\midrule
\multirow{8}{*}{Jaccard thresholds} & \multirow{4}{*}{exact 0.8} & C\_F & +5.2 & +4.7 & $-$0.5 \\
 &  & C\_G & +1.9 & +1.5 & $-$0.4 \\
 &  & C\_L & +4.0 & +3.4 & $-$0.5 \\
 &  & C\_P & +5.6 & +5.3 & $-$0.3 \\
 & \multirow{4}{*}{relaxed 0.6} & C\_F & +5.2 & +4.9 & $-$0.3 \\
 &  & C\_G & +1.9 & +1.2 & $-$0.7 \\
 &  & C\_L & +4.0 & +3.6 & $-$0.4 \\
 &  & C\_P & +5.6 & +4.9 & $-$0.7 \\
\midrule
\multirow{4}{*}{Notes per step} & \multirow{2}{*}{three notes} & C\_F & +5.2 & +5.1 & $-$0.1 \\
 &  & C\_G & +1.9 & +1.1 & $-$0.8 \\
 & \multirow{2}{*}{ten notes} & C\_F & +5.2 & +4.8 & $-$0.4 \\
 &  & C\_G & +1.9 & +2.7 & +0.8 \\
\midrule
\multirow{4}{*}{Adapter rank} & \multirow{2}{*}{rank 64} & W\_F & +0.0 & +0.1 & +0.1 \\
 &  & W\_G & +8.0 & +8.3 & +0.4 \\
 & \multirow{2}{*}{rank 128} & W\_F & +0.0 & +0.1 & +0.1 \\
 &  & W\_G & +8.0 & +8.1 & +0.2 \\
\midrule
\multirow{8}{*}{Pool size} & \multirow{4}{*}{five episodes per template} & C\_F & +5.2 & +3.8 & $-$1.4 \\
 &  & C\_G & +1.9 & +1.2 & $-$0.7 \\
 &  & W\_F & +0.0 & +0.0 & +0.0 \\
 &  & W\_G & +8.0 & +6.0 & $-$2.0 \\
 & \multirow{4}{*}{ten episodes per template} & C\_F & +5.2 & +5.0 & $-$0.2 \\
 &  & C\_G & +1.9 & +1.9 & +0.0 \\
 &  & W\_F & +0.0 & $-$0.2 & $-$0.2 \\
 &  & W\_G & +8.0 & +7.2 & $-$0.7 \\
\midrule
\multirow{2}{*}{Library cap} & 100 items & C\_G & +1.9 & +1.6 & $-$0.3 \\
 & 300 items & C\_G & +1.9 & +1.7 & $-$0.2 \\
\midrule
Notes block & five notes, no components & all\_C & +8.8 & +7.9 & $-$0.9 \\
\midrule
Rule input & online $\kappa$ in the rule & routed & +14.0 & +14.2 & +0.2 \\
\bottomrule
\end{tabularx}
\end{table}

\clearpage
\section{Cases}
\label{app:cases}

These two illustrative cases expand the episodes of Figure~\ref{fig:case} step by step, with the source of
every action marked as the weights or the context.

\subsection{Case 1: One Task, Two Sources}
\label{app:case1}

The locator comes from the weights and the state fact and the procedure from the context; with everything in
the weights the agent reaches the preview and leaves it without Add.

\begin{custombox}[title=Example from Weather]{toolcardborder}
\textbf{Task:} Add Tokyo to the saved cities and confirm that it appears in the list.

\vspace{0.5em}

\textbf{Screens:} Home, Cities, Preview, Added.
\end{custombox}

\begin{small}
\begin{casebox}{successbg}{successframe}{Routed}
\textbf{Step 1:}
\begin{itemize}[nosep, leftmargin=*]
    \item \textbf{Screen:} Home
    \item \textbf{Source:} weights, locator (Weather, control, city list)
    \item \textbf{Action:} open the city list
    \item \textbf{Result:} Cities
\end{itemize}
\vspace{0.5em}
\textbf{Step 2:}
\begin{itemize}[nosep, leftmargin=*]
    \item \textbf{Screen:} Cities
    \item \textbf{Source:} context, state fact \emph{``Tokyo is absent from the saved cities.''}
    \item \textbf{Action:} search Tokyo and open its preview
    \item \textbf{Result:} Preview
\end{itemize}
\vspace{0.5em}
\textbf{Step 3:}
\begin{itemize}[nosep, leftmargin=*]
    \item \textbf{Screen:} Preview
    \item \textbf{Source:} context, procedure \emph{``Tap Add on the preview, then confirm the city in the list.''}
    \item \textbf{Action:} tap Add
    \item \textbf{Result:} Added, Tokyo in the list
\end{itemize}
\vspace{0.5em}
\textbf{Outcome:} \textcolor{successframe}{Tokyo in the saved cities, task complete.}
\end{casebox}

\begin{casebox}{failbg}{failframe}{All to the weights}
\textbf{Step 1:}
\begin{itemize}[nosep, leftmargin=*]
    \item \textbf{Screen:} Home
    \item \textbf{Source:} weights, locator (Weather, control, city list)
    \item \textbf{Action:} open the city list
    \item \textbf{Result:} Cities
\end{itemize}
\vspace{0.5em}
\textbf{Step 2:}
\begin{itemize}[nosep, leftmargin=*]
    \item \textbf{Screen:} Cities
    \item \textbf{Source:} weights
    \item \textbf{Action:} search Tokyo and open its preview
    \item \textbf{Result:} Preview
\end{itemize}
\vspace{0.5em}
\textbf{Step 3:}
\begin{itemize}[nosep, leftmargin=*]
    \item \textbf{Screen:} Preview
    \item \textbf{Source:} weights; no note is served
    \item \textbf{Action:} Back
    \item \textbf{Result:} Cities, Tokyo not in the list
\end{itemize}
\vspace{0.5em}
\textbf{Outcome:} \textcolor{failframe}{Tokyo not saved; the episode ends at the list.}
\end{casebox}
\end{small}

\subsection{Case 2: A Note and a Learned Route Disagree}
\label{app:case2}

When a learned route and a note call for different entries on the same menu, the agent takes the action of
the note.

\begin{custombox}[title=Example from Calendar]{toolcardborder}
\textbf{Task:} Add the birthday of Alex on November 12 with the appropriate entry type.

\vspace{0.5em}

\textbf{Screens:} Home, Choices, Edit, Saved.
\end{custombox}

\begin{small}
\begin{casebox}{successbg}{successframe}{Routed}
\textbf{Step 1:}
\begin{itemize}[nosep, leftmargin=*]
    \item \textbf{Screen:} Home
    \item \textbf{Source:} weights, learned route to a new entry
    \item \textbf{Action:} open the new-entry menu
    \item \textbf{Result:} Choices, with Event and Birthday
\end{itemize}
\vspace{0.5em}
\textbf{Step 2:}
\begin{itemize}[nosep, leftmargin=*]
    \item \textbf{Screen:} Choices
    \item \textbf{Source:} context, note \emph{``For a birthday, choose Birthday; Event is the generic entry.''}
    The learned route points to Event.
    \item \textbf{Action:} choose Birthday
    \item \textbf{Result:} Edit
\end{itemize}
\vspace{0.5em}
\textbf{Step 3:}
\begin{itemize}[nosep, leftmargin=*]
    \item \textbf{Screen:} Edit
    \item \textbf{Source:} weights
    \item \textbf{Action:} enter Alex, set November 12, save
    \item \textbf{Result:} Saved, entry type Birthday, date November 12
\end{itemize}
\vspace{0.5em}
\textbf{Outcome:} \textcolor{successframe}{Birthday entry on November 12, task complete.}
\end{casebox}

\begin{casebox}{failbg}{failframe}{All to the weights}
\textbf{Step 1:}
\begin{itemize}[nosep, leftmargin=*]
    \item \textbf{Screen:} Home
    \item \textbf{Source:} weights, learned route to a new entry
    \item \textbf{Action:} open the new-entry menu
    \item \textbf{Result:} Choices, with Event and Birthday
\end{itemize}
\vspace{0.5em}
\textbf{Step 2:}
\begin{itemize}[nosep, leftmargin=*]
    \item \textbf{Screen:} Choices
    \item \textbf{Source:} weights, the learned route; no note is served
    \item \textbf{Action:} choose Event
    \item \textbf{Result:} Edit
\end{itemize}
\vspace{0.5em}
\textbf{Step 3:}
\begin{itemize}[nosep, leftmargin=*]
    \item \textbf{Screen:} Edit
    \item \textbf{Source:} weights
    \item \textbf{Action:} enter Alex, set November 12, save
    \item \textbf{Result:} Saved, entry type Event, date November 12
\end{itemize}
\vspace{0.5em}
\textbf{Outcome:} \textcolor{failframe}{A generic event on November 12; the entry type is wrong.}
\end{casebox}
\end{small}

\clearpage
\section{Compute and Tokens}
\label{app:compute}

Every arm is evaluated on the same held-out instances (Appendix~\ref{app:budget}). Two
accounts differ between arms: the training compute of the adapter an arm trains, and the prompt tokens it
spends per episode. Table~\ref{tab:app_compute} lists both for every arm of Table~\ref{tab:main} on MobileGym
with \llmname{GUI-Owl-8B}, as means over seeds; the other two families train the same arms at the same
configuration on pools of the same size. Adapter training is booked per training sample at the rates of one
H100, and pool sampling at a fixed per-episode rate that every arm reading the pool shares. Prompt tokens
are counted on every step with the notes block present, so the ratio to the base agent is the token cost of
the context route at the interaction budget of the paper.

\begin{table}[!htbp]
\centering
\caption{\textbf{Compute and tokens.} MobileGym, \llmname{GUI-Owl-8B}, mean over seeds: training samples and GPU-hours of the adapter an arm trains, and prompt tokens per episode (thousands) with their ratio to the base agent. Training is booked per sample on one H100, 2.45\,s for a supervised adapter and 5.40\,s for an adapter with a preference stage; sampling one pool of 2440 episodes is booked at 3.7 GPU-hours at a fixed per-episode rate; arms without an adapter train nothing.}
\label{tab:app_compute}
\footnotesize
\setlength{\tabcolsep}{4.5pt}
\renewcommand{\arraystretch}{1.15}
\begin{tabularx}{\textwidth}{l|>{\raggedleft\arraybackslash}X>{\raggedleft\arraybackslash}X|>{\raggedleft\arraybackslash}X>{\raggedleft\arraybackslash}X}
\toprule
& \multicolumn{2}{c|}{\cellcolor{ScienceBg}\textbf{Adapter training}} & \multicolumn{2}{c}{\cellcolor{SearchBg}\textbf{Inference}} \\ \cmidrule(lr){2-3}\cmidrule(lr){4-5}
\textbf{Arm} & \textbf{Samples} & \textbf{GPU-h} & \textbf{Tokens (k)} & \textbf{Ratio} \\
\midrule
Base agent & -- & -- & 13.2 & 1.00 \\
\midrule
Locators to the context & -- & -- & 14.2 & 1.07 \\
Procedures to the context & -- & -- & 14.2 & 1.07 \\
State facts to the context & -- & -- & 14.2 & 1.07 \\
Lessons to the context & -- & -- & 14.2 & 1.07 \\
\midrule
Locators to the weights & 1211 & 0.8 & 13.3 & 1.00 \\
Procedures to the weights & 5721 & 3.9 & 13.2 & 1.00 \\
State facts to the weights & 476 & 0.3 & 13.3 & 1.00 \\
Lessons to the weights & 2854 & 4.3 & 13.2 & 1.00 \\
\midrule
All to the context & -- & -- & 14.2 & 1.07 \\
All to the weights & 10263 & 15.4 & 13.2 & 1.00 \\
Routed & 4065 & 6.1 & 14.2 & 1.07 \\
Reverse assignment & 6197 & 9.3 & 14.2 & 1.07 \\
\midrule
Self-retry, three attempts & -- & -- & 13.2 & 1.00 \\
Success-filtered fine-tuning & 8210 & 5.6 & 13.2 & 1.00 \\
Retrieve one trajectory & -- & -- & 13.2 & 1.00 \\
Retrieve ten step records & -- & -- & 13.2 & 1.00 \\
Context then distill: procedures & 5721 & 3.9 & 13.3 & 1.00 \\
Context then distill: state facts & 476 & 0.3 & 13.3 & 1.00 \\
\bottomrule
\end{tabularx}
\end{table}

\clearpage
\section{Prompts}
\label{app:prompts}

Three kinds of text reach the model: the system prompt of each backbone, the user turn of each step, and the
notes block that the context route appends to the user turn. The MobileGym applications are Chinese, and so are
the Qwen3-VL prompt, the notes block and every note; they are reproduced verbatim below, each followed by an
English rendering. Line breaks marked by a grey arrow are for display only.

\subsection{Agent Prompts}
\label{app:agentprompt}

\begin{custombox}[title={System prompt, GUI-Owl-1.5-8B (MobileGym adapter)}]{toolcardborder}
\begin{alltt}\scriptsize
# Tools

You may call one or more functions to assist with the user query.

You are provided with function signatures within <tools></tools> XML tags:
<tools>
\{"type": "function", "function": \{"name_for_human": "mobile_use", "name": "mobile_use", "description
\textcolor{gray}{\(\hookrightarrow\)} ": "Use a touchscreen to interact with a mobile device, and take screenshots.
* This is an interface to a mobile device with touchscreen. You can perform actions like clicking, t
\textcolor{gray}{\(\hookrightarrow\)} yping, swiping, etc.
* Some applications may take time to start or process actions, so you may need to wait and take succ
\textcolor{gray}{\(\hookrightarrow\)} essive screenshots to see the results of your actions.
* The screen's resolution is 1000x1000.
* Make sure to click any buttons, links, icons, etc with the cursor tip in the center of the element
\textcolor{gray}{\(\hookrightarrow\)} . Don't click boxes on their edges unless asked.", "parameters": \{"properties": \{"action": \{"descrip
\textcolor{gray}{\(\hookrightarrow\)} tion": "The action to perform. The available actions are:
* `key`: Perform a key event on the mobile device.
    - This supports adb's `keyevent` syntax.
    - Examples: "volume_up", "volume_down", "power", "camera", "clear".
* `click`: Click the point on the screen with coordinate (x, y).
* `long_press`: Press the point on the screen with coordinate (x, y) for specified seconds.
* `swipe`: Swipe from the starting point with coordinate (x, y) to the end point with coordinates2 (
\textcolor{gray}{\(\hookrightarrow\)} x2, y2).
* `type`: Input the specified text into the activated input box.
* `system_button`: Press the system button.
* `open`: Open an app on the device.
* `wait`: Wait specified seconds for the change to happen.
* `answer`: Terminate the current task and output the answer.
* `terminate`: Terminate the current task and report its completion status.", "enum": ["key", "click
\textcolor{gray}{\(\hookrightarrow\)} ", "long_press", "swipe", "type", "system_button", "open", "wait", "answer", "terminate"], "type": "
\textcolor{gray}{\(\hookrightarrow\)} string"\}, "coordinate": \{"description": "(x, y): The x (pixels from the left edge) and y (pixels fro
\textcolor{gray}{\(\hookrightarrow\)} m the top edge) coordinates to move the mouse to. Required only by `action=click`, `action=long_pres
\textcolor{gray}{\(\hookrightarrow\)} s`, and `action=swipe`.", "type": "array"\}, "coordinate2": \{"description": "(x, y): The x (pixels fr
\textcolor{gray}{\(\hookrightarrow\)} om the left edge) and y (pixels from the top edge) coordinates to move the mouse to. Required only b
\textcolor{gray}{\(\hookrightarrow\)} y `action=swipe`.", "type": "array"\}, "text": \{"description": "Required only by `action=key`, `actio
\textcolor{gray}{\(\hookrightarrow\)} n=type`, `action=open`, `action=answer`.", "type": "string"\}, "time": \{"description": "The seconds t
\textcolor{gray}{\(\hookrightarrow\)} o wait. Required only by `action=long_press` and `action=wait`.", "type": "number"\}, "button": \{"des
\textcolor{gray}{\(\hookrightarrow\)} cription": "Back means returning to the previous interface, Home means returning to the desktop, Men
\textcolor{gray}{\(\hookrightarrow\)} u means opening the application background menu, and Enter means pressing the enter. Required only b
\textcolor{gray}{\(\hookrightarrow\)} y `action=system_button`", "enum": ["Back", "Home", "Menu", "Enter"], "type": "string"\}, "status": \{
\textcolor{gray}{\(\hookrightarrow\)} "description": "The status of the task. Required only by `action=terminate`.", "type": "string", "en
\textcolor{gray}{\(\hookrightarrow\)} um": ["success", "failure"]\}\}, "required": ["action"], "type": "object"\}, "args_format": "Format the
\textcolor{gray}{\(\hookrightarrow\)}  arguments as a JSON object."\}\}
</tools>

For each function call, return a json object with function name and arguments within <tool_call></to
\textcolor{gray}{\(\hookrightarrow\)} ol_call> XML tags:
<tool_call>
\{"name": <function-name>, "arguments": <args-json-object>\}
</tool_call>

# Response format

Response format for every step:
1) Action: a short imperative describing what to do in the UI.
2) A single <tool_call>...</tool_call> block containing only the JSON: \{"name": <function-name>, "ar
\textcolor{gray}{\(\hookrightarrow\)} guments": <args-json-object>\}.

Rules:
- Output exactly in the order: Action, <tool_call>.
- Be brief: one for Action.
- Do not output anything else outside those two parts.
- If finishing, use action=terminate in the tool call.
\end{alltt}
\end{custombox}

\begin{custombox}[title={System prompt, ScaleCUA-7B}]{toolcardborder}
\begin{alltt}\scriptsize
You are an autonomous GUI agent operating on the **Android** platform. Your primary function is to a
\textcolor{gray}{\(\hookrightarrow\)} nalyze screen captures and perform appropriate UI actions to complete assigned tasks.

## Action Space
def click(
    x: float | None = None,
    y: float | None = None,
    clicks: int = 1,
    button: str = "left",
) -> None:
    """Clicks on the screen at the specified coordinates. The `x` and `y` parameter specify where th
\textcolor{gray}{\(\hookrightarrow\)} e mouse event occurs. If not provided, the current mouse position is used. The `clicks` parameter sp
\textcolor{gray}{\(\hookrightarrow\)} ecifies how many times to click, and the `button` parameter specifies which mouse button to use ('le
\textcolor{gray}{\(\hookrightarrow\)} ft', 'right', or 'middle')."""
    pass


def swipe(
    from_coord: tuple[float, float] | None = None,
    to_coord: tuple[float, float] | None = None,
    direction: str = "up",
    amount: float = 0.5,
) -> None:
    """Performs a swipe action on the screen. The `from_coord` and `to_coord` specify the starting a
\textcolor{gray}{\(\hookrightarrow\)} nd ending coordinates of the swipe. If `to_coord` is not provided, the `direction` and `amount` para
\textcolor{gray}{\(\hookrightarrow\)} meters are used to determine the swipe direction and distance. The `direction` can be 'up', 'down', 
\textcolor{gray}{\(\hookrightarrow\)} 'left', or 'right', and the `amount` specifies how far to swipe relative to the screen size (0 to 1)
\textcolor{gray}{\(\hookrightarrow\)} ."""
    pass


def long_press(x: float, y: float, duration: int = 1) -> None:
    """Long press on the screen at the specified coordinates. The `duration` specifies how long to h
\textcolor{gray}{\(\hookrightarrow\)} old the press in seconds."""
    pass


def open_app(app_name: str) -> None:
    """Open an app on the device."""
    pass


def navigate_home() -> None:
    """Press the home button."""
    pass


def navigate_back() -> None:
    """Press the back button."""
    pass


def write(message: str) -> None:
    """Write the specified text."""
    pass


def call_user() -> None:
    """Call the user."""
    pass


def wait(seconds: int = 3) -> None:
    """Wait for the change to happen."""
    pass


def response(answer: str) -> None:
    """Answer a question or provide a response to an user query."""
    pass


def terminate(status: str = "success", info: str | None = None) -> None:
    """Terminate the current task with a status. The `status` specifies the termination status ('suc
\textcolor{gray}{\(\hookrightarrow\)} cess', 'failure'), and the `info` can provide additional information about the termination."""
    pass


## Input Specification
- Screenshot of the current screen + task description + your past interaction history with UI to fin
\textcolor{gray}{\(\hookrightarrow\)} ish assigned tasks.

## Output Format
```
<think>
[Your reasoning process here]
</think>
<operation>
[Next intended operation description]
</operation>
<action>
[A set of executable action command]
</action>
```

## Note
- Avoid actions that would lead to invalid states.
- The generated action(s) must exist within the defined action space.
- The reasoning process, operation and action(s) in your response should be enclosed within <think><
\textcolor{gray}{\(\hookrightarrow\)} /think>, <operation></operation> and <action></action> tags, respectively.
\end{alltt}
\end{custombox}

\begin{custombox}[title={System prompt, Qwen3-VL-8B (Chinese, verbatim)}]{toolcardborder}
\begin{CJK}{UTF8}{gbsn}
\begin{alltt}\scriptsize
你是一个手机 GUI-Agent 操作专家。你需要根据用户下发的任务、手机屏幕截图以及历史操作记录，分析当前界
\textcolor{gray}{\(\hookrightarrow\)} 面并输出一个动作来与手机交互，从而完成任务。

坐标系：左上角为原点，x 向右，y 向下，取值范围均为 0-1000（归一化坐标）。

可用动作（JSON 格式）：

1. 点击：\{"action": "CLICK", "point": [x, y]\}
2. 双击：\{"action": "DOUBLE_TAP", "point": [x, y]\}
3. 长按：\{"action": "LONGPRESS", "point": [x, y]\}
4. 输入：\{"action": "TYPE", "value": "文本内容"\}  // 可选 "point": [x, y] 指定输入位置；可选 "clear"
\textcolor{gray}{\(\hookrightarrow\)} : true 先清空输入框再输入（默认追加到已有文本后面）
5. 滑动：\{"action": "SWIPE", "point1": [x1, y1], "point2": [x2, y2]\}
6. 拖拽：\{"action": "DRAG", "point1": [x1, y1], "point2": [x2, y2]\}  // 按住起点拖动到终点
7. 返回：\{"action": "BACK"\}
8. 回到桌面：\{"action": "HOME"\}
9. 打开最近任务：\{"action": "RECENT"\}
10. 输入回车：\{"action": "ENTER"\}
11. 等待：\{"action": "WAIT", "value": 秒数\}
12. 打开应用：\{"action": "AWAKE", "value": "应用名称"\}
13. 提交答案：\{"action": "ANSWER", "value": "纯答案文本"\}
14. 任务完成：\{"action": "COMPLETE", "return": "完成说明"\} // 所有任务完成后使用，给出简短的说明
15. 中止任务：\{"action": "ABORT", "value": "中止原因"\}  // 任务无法完成时使用，需要说明原因


你必须按以下格式输出：

<THINK>
在这里描述你对当前屏幕的理解、分析和决策过程。
包括：
1. 当前屏幕显示的内容是什么
2. 为了完成任务，下一步应该做什么
3. 具体要点击/操作哪个元素
</THINK>
<ANSWER>
\{
  "action": "动作类型",
  // 根据动作类型填写相应参数
\}
</ANSWER>


要求：
- 坐标必须为数字，范围 0-1000
- JSON 必须是有效格式
- 仔细观察屏幕截图，根据视觉信息做出判断
- 需要回答问题时，必须使用 ANSWER 提交答案
- COMPLETE 只用于结束任务，需要在执行完任务后使用
\end{alltt}
\end{CJK}
\end{custombox}

\noindent\emph{English rendering of the Qwen3-VL prompt.} You are an expert operator of a mobile GUI agent. Given
the task, the current screenshot and the history of operations, analyse the screen and output one action to
interact with the phone and complete the task. Coordinates have their origin at the top left, $x$ to the right
and $y$ downward, both in the range 0 to 1000. The fifteen actions, in JSON, are click, double tap, long press,
type (with an optional point and an optional clear flag), swipe, drag, back, home, recent apps, enter, wait,
open app, submit an answer, complete the task with a short note, and abort with a reason. The output is a
\texttt{<THINK>} block that states what the screen shows, what to do next and which element to operate, followed
by an \texttt{<ANSWER>} block with one JSON action. Coordinates must be numbers in 0 to 1000, the JSON must be
valid, decisions follow the screenshot, questions are answered with the answer action, and the complete action
is used only after the task has been carried out.

\subsection{User Turn}
\label{app:userturn}

The user turn of a step carries the current screenshot and, on the first step of the history window, the
instruction text below with the task and the summary of previous actions filled in. \llmname{GUI-Owl-1.5-8B}
keeps a window of the last four steps, each as its screenshot followed by the full model output of that step, and
lists older steps as one line each; \llmname{ScaleCUA-7B} sends the current screenshot with a one-line history;
\llmname{Qwen3-VL-8B} keeps its previous outputs as text and sends the screenshot of the current step only. A
weights-route arm and the base agent receive exactly these messages. A context-route arm receives the same
messages plus one text part appended after the current screenshot, the notes block of
Appendix~\ref{app:notesblock}, and nothing is appended on a step without retrieved items. For the readout probe
of Section~\ref{sec:obs3}, every step that received a notes block is also run once without it; both outputs are
logged, and the action with the block is the one executed.

\begin{custombox}[title={Instruction text of the first user turn, GUI-Owl-1.5-8B}]{toolcardborder}
\begin{alltt}\scriptsize
Please generate the next move according to the UI screenshot, instruction and previous actions.

Instruction: \{task\}

Previous actions:
\{previous_actions\}
\end{alltt}
\end{custombox}
\begin{custombox}[title={Instruction text of the user turn, ScaleCUA-7B}]{toolcardborder}
\begin{alltt}\scriptsize
Please generate the next move according to the UI screenshot, the task and previous operations.

Task:
\{task\}

Previous operations:
\{history\}
\end{alltt}
\end{custombox}
\begin{custombox}[title={First user turn, Qwen3-VL-8B}]{toolcardborder}
\begin{CJK}{UTF8}{gbsn}
\begin{alltt}\scriptsize
[任务]
\{task\}
\end{alltt}
\end{CJK}
\end{custombox}

\noindent\emph{English rendering of the Qwen3-VL user turn.} The first turn is the line ``[Task]'' followed by the
task; later turns are ``[Step $i$]'' followed by the current screenshot.

\subsection{Notes Block}
\label{app:notesblock}

The block is one text part with a fixed header and one line per retrieved item, in retrieval order; the
aggregate arms list the components in the order locators, procedures, state facts, lessons. Two headers exist,
one per coordinate frame.

\begin{custombox}[title={Notes block, GUI-Owl-1.5-8B and Qwen3-VL-8B (0--1000 frame)}]{toolcardborder}
\begin{CJK}{UTF8}{gbsn}
\begin{alltt}\scriptsize
[经验笔记]（来自过去在同一应用上的使用经验，仅供参考；坐标与截图同为 0–1000 尺度；以当前截图和任务为
\textcolor{gray}{\(\hookrightarrow\)} 准）
\{notes\}
\end{alltt}
\end{CJK}
\end{custombox}
\begin{custombox}[title={Notes block, ScaleCUA-7B (952 x 2156 frame)}]{toolcardborder}
\begin{CJK}{UTF8}{gbsn}
\begin{alltt}\scriptsize
[经验笔记]（来自过去在同一应用上的使用经验，仅供参考；坐标为模型看到的截图像素坐标（952 × 2156）；以
\textcolor{gray}{\(\hookrightarrow\)} 当前截图和任务为准）
\{notes\}
\end{alltt}
\end{CJK}
\end{custombox}

\noindent\emph{English rendering of the header.} ``[Experience notes] (from past use of the same application, for
reference only; coordinates are on the same 0 to 1000 scale as the screenshot; the current screenshot and task
take precedence)''. The ScaleCUA header states instead that coordinates are in the pixel frame the model sees,
952 by 2156.

\begin{custombox}[title={Item templates, one line per item}]{toolcardborder}
\begin{CJK}{UTF8}{gbsn}
\begin{alltt}\scriptsize
G  在 \{app\} 的该页面上，元素「\{name\}」（\{role\}）的点击坐标是 (\{x\}, \{y\})。
P  在 \{app\} 中，做完「\{name1\}」「\{name2\}」后，下一步通常是「\{name3\}」。
F  在 \{app\} 中执行「\{name\}」之前，需要满足：\{pred1\}；\{pred2\}；…。
L  在 \{app\} 的该状态下不要执行「\{name_fail\}」，应执行「\{name_success\}」。
\end{alltt}
\end{CJK}
\end{custombox}

\noindent\emph{English rendering of the templates.} A locator note reads ``On this page of \{app\}, the element
`\{name\}' (\{role\}) is tapped at (\{x\}, \{y\})''. A procedure note reads ``In \{app\}, after `\{name1\}' and
`\{name2\}', the next step is usually `\{name3\}'{}''. A state-fact note reads ``Before doing `\{name\}' in \{app\},
the following must hold: \{pred1\}; \{pred2\}; \ldots'', with at most five predicates. A lesson note reads ``In
this state of \{app\}, do not do `\{name\_fail\}'; do `\{name\_success\}' instead''. A predicate is one of three
forms over a path of the application state, with list identifiers replaced by a placeholder: ``\{path\} exists'',
``\{path\} is set (non-empty)'', or ``\{path\} must be the value the task requires''; keys that change on every
run, such as timestamps and identifiers, never enter a predicate. Retrieval and the per-step quota are those of
Appendix~\ref{app:contextroute}.

\stopcontents[appendix]

\end{document}